\documentclass[preprint,11pt,authoryear]{elsarticle}

\usepackage{amsmath,amssymb,mathrsfs}
\usepackage{amsfonts}
\usepackage{algorithm}
\usepackage{algorithmic}
\usepackage{indentfirst}
\usepackage{subfigure}
\usepackage{caption}
\usepackage{hyperref}
\usepackage{booktabs}
\usepackage{multirow}
\usepackage{threeparttable}
\usepackage{verbatim}
\usepackage{xcolor}
\usepackage{setspace}
\usepackage{geometry}
\usepackage{etoolbox}
\usepackage{amsthm}

\newtheorem{definition}{Definition}
\newtheorem{theorem}{Theorem}
\newtheorem{lemma}{Lemma}

\newtheorem{corollary}{Corollary}
\newtheorem{assumption}{Assumption}
\newtheorem{remark}{Remark}

\begin{document}

\begin{frontmatter}



\title{Efficient Generalized Low-Rank Tensor Contextual Bandits}


\author[a]{Qianxin Yi}\ead{YiQianXin01@163.com}
\author[a]{Yiyang Yang}\ead{yyyang817@gmail.com}
\author[b]{Shaojie Tang}\ead{Shaojie.Tang@utdallas.edu}
\author[a]{Jiapeng Liu}\ead{jiapengliu@mail.xjtu.edu.cn}
\author[a]{Yao Wang\corref{correspondingauthor}}
\cortext[correspondingauthor]{Corresponding author at: School of Management, Xi'an Jiaotong University, No.28, West Xianning Road, Xi'an, Shaanxi, 710049, P.R. China.}\ead{yao.s.wang@gmail.com}
\address[a]{Center for Intelligent Decision-Making and Machine Learning, School of Management, Xi'an Jiaotong University, 
	Xi'an,
	China}
\address[b]{Naveen Jindal School of Management, The University of Texas at Dallas, 
	Richardson,
	Texas,
	USA}

\begin{abstract}
In this paper, we aim to build a novel bandits algorithm that is capable of fully harnessing the power of multi-dimensional data and the inherent non-linearity of reward functions to provide high-usable and accountable decision-making services. To this end, we introduce a generalized low-rank tensor contextual bandits model in which an action is formed from three feature vectors, and thus can be represented by a tensor. In this formulation, the reward is determined through a generalized linear function applied to the inner product of the action's feature tensor and a fixed but unknown parameter tensor with a low tubal rank. To effectively achieve the trade-off between exploration and exploitation, we introduce a novel algorithm called ``Generalized Low-Rank Tensor Exploration Subspace then Refine" (G-LowTESTR). This algorithm first collects raw data to explore the intrinsic low-rank tensor subspace information embedded in the decision-making scenario, and then converts the original problem into an almost lower-dimensional generalized linear contextual bandits problem. Rigorous theoretical analysis shows that the regret bound of G-LowTESTR is superior to those in vectorization and matricization cases. We conduct a series of simulations and real data experiments to further highlight the effectiveness of G-LowTESTR, leveraging its ability to capitalize on the low-rank tensor structure for enhanced learning.
\end{abstract}
\begin{keyword}
Artificial intelligence, Generalized linear tensor learning \sep Low-rank contextual bandits \sep Transformed t-SVD \sep Sequential decision-making


\end{keyword}

\end{frontmatter}


\section{Introduction}
The bandits problem represents a form of sequential decision-making challenge, where a learner must repeatedly make choices among options with unknown outcomes, aiming to maximize cumulative rewards over time \citep{thompson1933likelihood}. Central to all bandits problems is the intricate balance between exploration and exploitation. Various practical domains can be modeled using bandits frameworks, including precision medicine \citep{bastani2020online, lu2021bandit}, recommendation systems \citep{bastani2022learning, gur2022adaptive}, and online advertising \citep{schwartz2017customer, aramayo2023multiarmed}. To illustrate, in precision medicine, the challenge is to find an equilibrium between exploring novel treatment avenues (exploration) and leveraging established methods (exploitation), promptly selecting the optimal treatment plan to enhance therapeutic effects. In recommendation systems, the goal is to strike a harmony between exploring new strategies (exploration) and evaluating existing ones (exploitation), thus rapidly crafting the finest recommendation approach to maximize user satisfaction. In online advertising, companies must juggle the continuous exploration of fresh advertising strategies (exploration) with the utilization of proven methods (exploitation), swiftly pinpointing the optimal strategy for maximizing advertising revenue. In various contexts, the classical model for this scenario is the multi-armed bandits (MABs) model \citep{lai1985asymptotically, auer2002finite, bubeck2012regret}. The fundamental goal of MABs is that, at each time point, the decision-maker must select an action from a discrete action space to execute and then receive feedback from the environment.

Contextual bandits are designed to harness feature information during the decision-making process and have gained significant attention. Specifically, decision-makers possess key features (e.g., user profiles) linked to actions prior to choices. Integrating this feature information with environmental feedback is crucial for optimizing rewards within the given context. One intuitive modeling approach involves establishing a linear relationship between the action features and the expected rewards, known as linear contextual bandits \citep{li2010contextual, abbasi2011improved, bastani2020online}. However, real-world scenarios often reveal a generalized linear relationship between actions and rewards \citep{grant2020adaptive, grant2021filtered, agrawal2023tractable}. The generalized linear model extends linear regression by linking the linear model with the response variable via a link function, encompassing techniques such as linear regression, logistic regression, and Poisson regression. For example, in online advertising, user clicks are typically treated as rewards, yielding a binary reward variable (0 for no click, 1 for a click), which can be effectively modeled using a binary logistic model. Linear contextual bandits models fall short in such and other similar cases. To tackle this challenge, works \citep{filippi2010parametric, li2012unbiased} introduce the link function to extend linear contextual bandits models into generalized linear contextual bandits models. These models are suitable for decision scenarios where action features are represented as vectors.

However, in many real-world application scenarios, available actions may involve choices among two or three categories of objects. For instance, on a travel website, these categories could be [\textit{flight, hotel}], while on a clothing website, they might include [\textit{top, pant, shoe}]. Additionally, each category of objects has its corresponding feature information. In such cases, representing action feature data as vectors becomes impractical, rendering the aforementioned generalized linear contextual bandits framework unsuitable for these intricate decision scenarios. In reality, for action choice problems involving two or three categories of objects, a natural approach is to utilize matrices or tensors to represent their action features. Considering that various options within the same category often exhibit a high degree of similarity (for example, in the case of hotel selection, those with the same star rating often share common attributes like price and services), the feature matrix or tensor inherently possesses a low-rank structure. Derived from this insight, the generalized low-rank matrix contextual bandits and low-rank tensor contextual bandits models emerge \citep{jun2019bilinear, zhou2020stochastic, lu2021low, kang2022efficient, ide2022targeted, shi2023high}, with a specific focus on addressing contextual bandits problems characterized by low-rank structural features.

As stated above, in both matrix and tensor scenarios, researchers have incorporated the low-rank information arising from feature interactions into linear contextual bandits. However, the non-linearity, particularly the generalized non-linearity, of rewards in tensor contextual bandits has not been thoroughly explored. Yet, in many real-world applications, feature information often presents itself in tensor form, and rewards display non-linear patterns. This underscores the importance of our research, as it addresses this gap and strives to develop effective approaches capable of handling non-linear (particularly generalized linear) rewards within tensor contextual bandit scenarios.

\subsection{Applications of generalized low-rank tensor contextual bandit}
In this section, we employ precision medicine and online advertising as case studies to exemplify the non-linear formulation of the reward function, underscoring the critical need to study the generalized low-rank tensor contextual bandits.

\subsubsection{Binary rewards in precision medicine} Precision drug recommendation, a specific facet of precision medicine, focuses on the selection of the most suitable drug for an individual patient with a particular medical condition. Existing studies predominantly concentrate on singular interaction relationships, like [\textit{drug, target}] \citep{cobanoglu2013predicting} and [\textit{drug, disease}] \citep{luo2018computational} relationships, covering only a fraction of the interconnections between drugs, targets, and diseases. A drug acts against a certain disease by inactivating the pathogenic organism after acting on the target. Therefore, delving into the drug-target-disease triplet association can reveal the potential molecular mechanism. Nevertheless, in cases where historical behavioral records are unavailable, it becomes necessary to explore new matching relationships between emerging drugs, potential targets, and specific diseases. Such exploration could result in costs and delay patient treatment. The objective is to strike the right balance between exploring new [\textit{drug, target, disease}] matching relationships and leveraging historical information. This exploration-exploitation dilemma can be effectively addressed through bandits algorithms.

\begin{figure}[!ht]
	\centering
	\includegraphics[scale=.095]{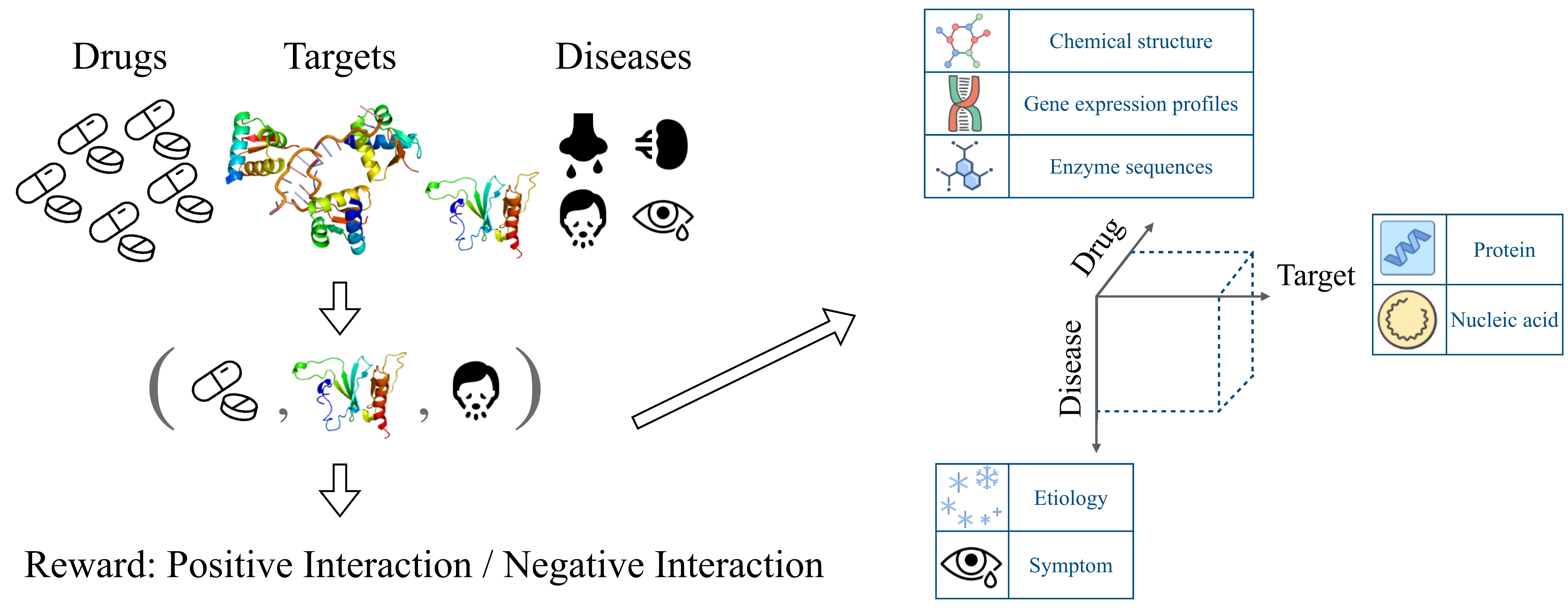}       
 	\caption{Motivating binary rewards example: precise drug recommendation}
  \label{figure:tensor}
\end{figure}

Precise drug recommendation applications involve abundant contextual information. For instance, each individually named disease possesses distinct characteristics, such as etiology, signs and symptoms, and morphological and functional changes. Diseases that share common characteristics are classified together in the disease classification system. Various drugs have diverse attributes, such as chemical structures, gene expression profiles, drug target sequences, and drug-related enzyme sequences. The target is the substance that interacts with a drug and can be categorized into proteins and nucleic acids, among others. Given that treatment response (reward) is influenced by several factors, we can study precise drug recommendation problems using the low-rank tensor contextual bandits framework. Similar low-rank structures have appeared in tensor recovery learning problems related to precision medicine \citep{murali2021cancer}. We model the triplet association by introducing a third-order tensor, as depicted in Fig. \ref{figure:tensor}, where the three dimensions represent drug, target, and disease features, respectively. The drug response (reward) is typically a binary variable indicating positive or negative interaction. This discrete, non-continuous reward calls for introducing the generalized low-rank tensor contextual bandits model.

\subsubsection{Poisson (count-based) rewards in online advertising} In the realm of online advertising, the primary objective is to strategically match relevant advertisements (ads) to individual users, thereby optimizing the click-through rate (CTR) of the displayed ads. Using sponsored search \citep{hillard2010improving} as an example, the interaction context comprises a triplet of key elements: the user, the ad, and the search keywords. The reward in this context is determined by the user's response, typically quantified as the click count on the displayed ad \citep{tang2015personalized}. While previous efforts have largely concentrated on individual interactions, such as user preferences for specific ads \citep{dhillon2021modeling}, they only cover a portion of the intricate web of relationships between users, ads, and search keywords. The challenge in this scenario lies in effectively understanding and leveraging the intricate relationship between the user's attributes, the ad characteristics, the relevance to the search keywords, and the resulting user engagement (measured by click count). Hence, an exploration of the connections between users, ads, and search keywords can unveil potential underlying mechanisms of resonance and appeal. However, there is a need to uncover relationships between new users, ads, and search keywords when historical interaction records are absent. This exploration comes with its costs and may temporarily reduce immediate gains in user engagement. Our goal is to strike an optimal balance between exploring new [\textit{user, ad, search keywords}] affinities and leveraging the insights gained from past interactions. This closely aligns with the fundamental principles of bandits algorithms, providing an elegant solution to the complex dilemma at hand.

\begin{figure}[!ht]
	\centering
	\includegraphics[scale=.095]{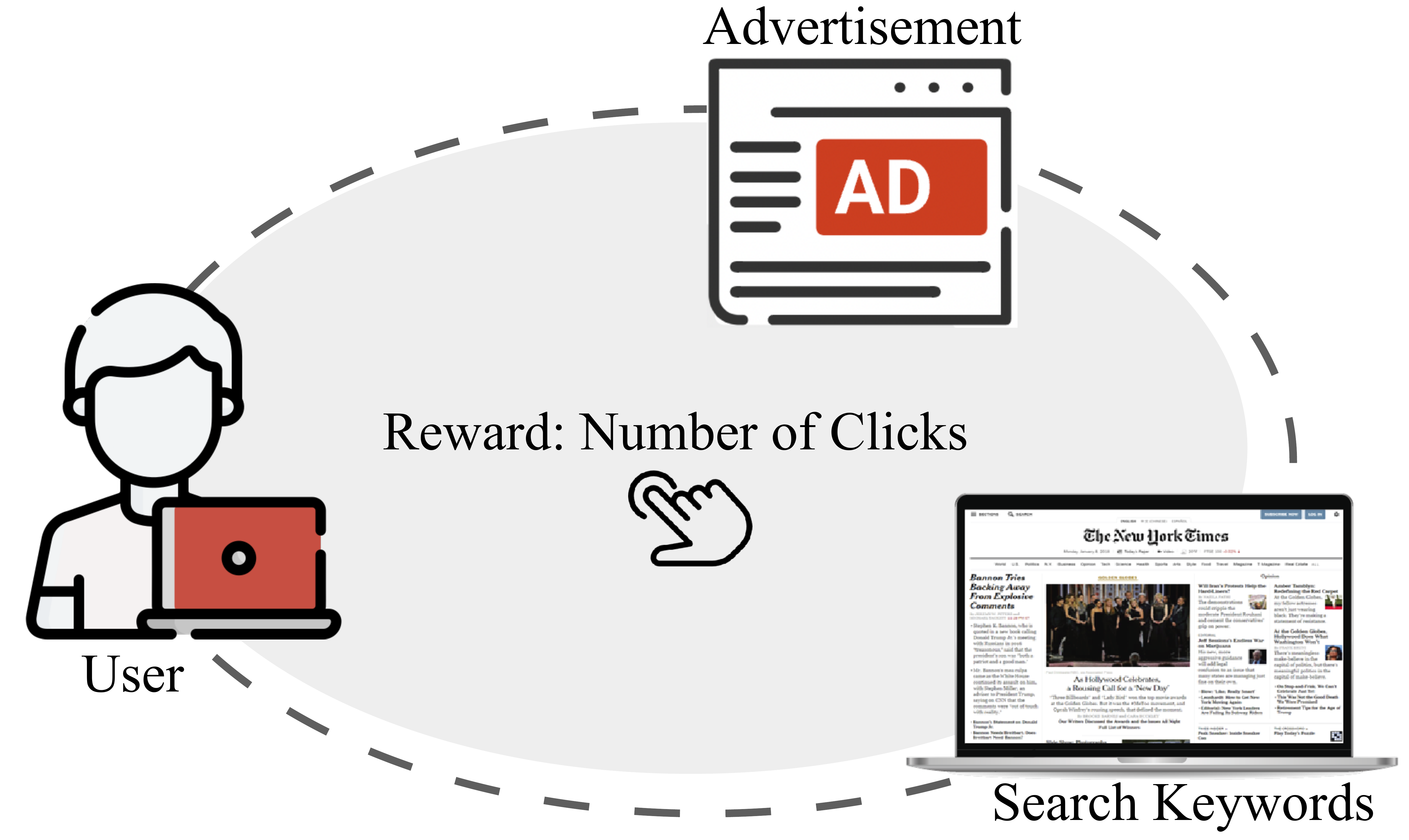}       
 	\caption{Motivating Poisson rewards example: online advertising}
        \label{figure:tensor4}
\end{figure}

Online advertising applications are replete with contextual information. The user's attributes comprise elements such as browsing history, demographics, behavioral patterns, and preferences. On the other hand, ad features encompass visual content, tone, and multimedia elements. Additionally, the chosen search keywords provide crucial contextual information. As shown in Fig. \ref{figure:tensor4}, the user response (reward) typically follows a Poisson (count-based) distribution of click numbers \citep{tang2015personalized}, emphasizing the need for the generalized low-rank tensor contextual bandits model to effectively explore and exploit this feature space. Learning the Poisson reward function is crucial for maximizing user engagement, showcasing the necessity of leveraging the generalized low-rank tensor contextual bandits models.

Beyond the aforementioned two examples, the framework of generalized low-rank tensor contextual bandits is also applicable to various other sequential decision-making scenarios. For the [\textit{top, bottom, shoe}] triplet recommendations on clothing recommendation websites, rewards are often binary variables indicating purchase (click) or no purchase (no click). Similarly, in advertising campaigns, marketers need to comprehensively consider the influence of customers, promotion methods, and marketing channels when designing marketing strategies for new products. The rewards generated by the joint choice of this triplet are also binary variables indicating clicks (purchases) or no clicks (no purchases). The above examples demonstrate that the reward functions in practical application scenarios are often nonlinear (e.g., binary or count-based). Existing linear low-rank tensor contextual bandits are no longer applicable. There is an urgent need to propose a tensor contextual bandits model suitable for nonlinear reward functions, where the multidimensional feature interactions of actions exhibit the low-rank structure. Effective exploration and exploitation algorithms for bandits need to be designed to achieve more accurate decision-making, accompanied by rigorous theoretical analysis of regret bounds.

\subsection{Contributions}
In this paper, we aim to address the following questions:

\begin{itemize}
	\item How to leverage low-rank tensor contextual bandits models to learn non-linear (especially generalized linear) reward functions?
	\item How can an efficient exploration and exploitation algorithm be designed to learn a generalized linear reward function by leveraging the inherent low-rank structure of tensor-formed features?
	\item How can the accuracy of the proposed algorithm be quantified? Specifically, how can we perform the regret bound analysis for the generalized low-rank tensor contextual bandits model?
	\item What are the practical implications of this study? Can the model assist decision-makers in making more informed and accurate choices?
\end{itemize}

The main contributions of this paper are summarized as follows:
\subsubsection{Modeling paradigm} In many real-world sequential decision scenarios, numerous reward functions are binary (e.g., positive interaction or negative interaction) or Poisson (e.g., click count), rather than being strictly continuous. Considering the non-linear relationship between action features and expected rewards, we propose a novel approach known as generalized low-rank tensor contextual bandits based on the efficient transformed t-product framework \citep{kernfeld2015tensor}. This framework is selected for two reasons. Compared to popular approaches based on tensor decomposition (e.g., canonical polyadic [CP], and Tucker), methods utilizing the transformed t-product are both computationally efficient and capable of achieving tight low-rank approximations akin to those in matrix scenarios. Moreover, they offer increased flexibility in managing diverse real-world situations, thus enhancing the learning process through the integration of the invertible linear transformation matrix.

\subsubsection{Algorithmic paradigm} Motivated by the low-rank subspace exploration algorithms proposed by \cite{lu2021low} and  \cite{kang2022efficient}, we introduce a new algorithm named ``Generalized Low-Rank Tensor Exploration Subspace then Refine" (G-LowTESTR). The goal is to achieve an efficient and effective balance between exploration and exploitation when dealing with generalized low-rank tensor contextual bandits problems. Such exploration algorithms within a low-rank tensor subspace can significantly enhance decision accuracy by leveraging the generalized linear relationship between action features and rewards obtained during the exploration stage. When the features are in tensor form, we need to solve the generalized linear model through an optimization problem with tensor nuclear norm regularization. The estimation error bound in this context directly impacts the regret, thus necessitating a rigorous proof. The analysis of the tensor regression error bound under the generalized linear model is intricate, presenting a considerable challenge for this proof. This paper primarily addresses this challenge by analyzing it based on a novel local restricted strong convexity (LRSC) condition.

\subsubsection{Theoretical implications} The regret measures cumulative losses incurred at each time point due to suboptimal action selection and is a common metric for evaluating bandits algorithms. For the $d\times d \times d$ action feature, and the unknown parameter in the generalized linear model defined as $\mathcal{W}^*$, the regret bound of our proposed algorithm, G-LowTESTR, is $\widetilde{O}\left(\frac{d^2\sqrt{\ell T}}{\sqrt{a}(1-\gamma)} \right)$, where $T$ is the horizon, $\gamma$ measures the gap between the first and $\lceil ad\operatorname{rank}_t(\mathcal{W}^*)\rceil$-th singular values of the action feature, $a\in[0,1]$, and $\ell$ is the coefficient of the invertible linear transform (practically taken as $\ell=1$), $\operatorname{rank}_t(\cdot)$ refers to the tensor tubal rank, which will be formally defined in Definition 7. Note that the order of $T$ and $d$ aligns with the order in the regret bounds of existing low-rank contextual bandits algorithms \citep{kang2022efficient,shi2023high}. The parameter `$\frac{1}{\sqrt{a}(1-\gamma)}$' incorporates rank information, and in experiments, this value aligns well with the rank of the unknown parameter. In addition, the research conducted by \cite{kang2022efficient} achieves optimal regret bounds in the matrix case. However, it relies on the impractical assumption that the score function of arm distribution can be accurately learned, making the application of their algorithm challenging in scenarios with extensive datasets. Although the work by \cite{shi2023high} focuses on the tensor bandits using Tucker decomposition, it is computationally demanding in real-world applications and also limited to the scenarios with linear rewards.

\subsubsection{Practical implications} Given that decision-makers must incorporate multi-dimensional feature information to optimize real-time decision-making, and considering the diverse nature of response variables (including continuous, binary, count, and multiclass types), the adoption of the suggested generalized low-rank tensor contextual bandits model and algorithm, as outlined in this paper, is highly suitable. To validate its practical effectiveness, we conduct a comprehensive series of experiments encompassing both synthetic and real-world datasets. Notably, our approach achieves a lower regret bound and computational complexity, resulting in improved accuracy and efficiency in decision-making. This underscores the effectiveness of our proposed G-LowTESTR in providing more informed decisions for complex decision-making challenges.
\subsection{Organization of the paper}
The paper is structured as follows. In section 2, we discuss existing bandits problems, and derive our research contribution. The problem formulation of generalized low-rank tensor contextual bandits is described in section 3. In section 4, we introduce the G-LowTESTR algorithm and analyze its regret bound. In section 5, we evaluate the performance of the algorithm on synthetic datasets and then further verify its ability by applying it to deal with precision medicine and advertising placement applications. Finally, we present our main findings and future research directions in section 6.

\section{Related work}
Although MAB and linear contextual bandits cover various applications in sequential decision-making, they cannot effectively depict the feedback information in more complex decision-making scenarios. Recently, there have been some works proposing rank-one \citep{katariya2017stochastic,katariya2017bernoulli} and the general (generalized) low-rank matrix contextual bandits \citep{gopalan2016low,kveton2017stochastic,jun2019bilinear, lu2021low, jang2021improved, 
	kang2022efficient}.
These types of bandits make full use of the low-rank structure generated by the interaction between entities, leading to a lower regret bound, which has profound theoretical and practical implications \citep{rizk2021best, li2022simple}. Specifically, to efficiently solve the low-rank matrix contextual bandits problem, \cite{lu2021low} reduces it to low-dimensional linear contextual bandits by using the subspace estimate obtained from the matrix nuclear norm optimization. \cite{kang2022efficient} further integrates the Stein method into the generalized low-rank matrix contextual bandits to achieve optimal regret. However, this method is accompanied by an increase in computational complexity.

Despite the widely studied low-rank bandits, research from the tensor perspective is relatively rare. To our best knowledge, \cite{zhou2020stochastic} is the first work to propose the concept of low-rank tensor contextual bandits under Tucker decomposition. In their framework, they model the expected reward as the inner product between the true reward tensor and the action tensor. Notably, in this construct, the first dimensions of the true reward tensor correspond to feature information, while the subsequent dimensions correspond to the arms. It's worth highlighting that their action tensors are restricted to one-hot encoding. More recently, \cite{shi2023high} explores a broader range of scenarios by considering general action tensors, thereby proposing an alternative low-dimensional subspace utilization algorithm. Furthermore, \cite{ide2022targeted} proposed a distinct tensor contextual bandits framework utilizing CP decomposition for online targeted advertising, where the reward is modeled as the inner product of the action's feature tensor and an unknown low-rank parameter tensor. 

Our work is related to \cite{zhou2020stochastic}, \cite{lu2021low}, \cite{kang2022efficient}, \cite{ide2022targeted} and \cite{shi2023high}, but has significantly differences. Unlike the (generalized) low-rank matrix contextual bandits framework proposed by \cite{lu2021low} and \cite{kang2022efficient}, which focus on two-dimensional interactions of action features, our approach considers a tensor contextual bandits framework capable of capturing multi-dimensional action features. Notably, this expansion is far from straightforward due to the entirely distinct nature of the tensor framework. As an innovative approach, we ingeniously employ tensor nuclear norm regularization to characterize the inherent low-rank nature of the parameter tensor. In addition, we introduce novel techniques like the tensor local restricted strong convexity (LRSC) condition for theoretical analysis. Distinct from the research carried out by \cite{zhou2020stochastic}, \cite{shi2023high}, and \cite{ide2022targeted} in the multi-armed or low-rank tensor contextual bandits frameworks, this work introduces a more comprehensive generalized low-rank tensor contextual bandits model. Our model finds application in decision scenarios encompassing diverse response (reward) variable types, such as continuous, binary, count, and multiclass. Moreover, our model is built upon the transformed t-product framework. In contrast to models relying on other tensor decomposition ways,  our model not only enhances computational efficiency but also achieves the same tight low-rank approximation as that in matrix cases.

\section{The generalized low-rank tensor contextual bandits model}
To make the most of the intricate contextual information available to us, we propose a generalized low-rank tensor contextual bandits model. To illustrate the workings of this model, we use the aforementioned advertising placement as an example to explain. At each time $t$, the decision-maker chooses a certain [\textit{user, ad, search keywords}] triple, and observes the features of user, ad, and search keywords that are denoted as $\mathbf{a}_1,\mathbf{a}_2,\ldots,\mathbf{a}_m\in \mathbb{R}^{d_1}$, $\mathbf{b}_1,\mathbf{b}_2,\ldots,\mathbf{b}_m\in \mathbb{R}^{d_2}$, and $\mathbf{c}_1,\mathbf{c}_2,\ldots,\mathbf{c}_m\in \mathbb{R}^{d_3}$ respectively. Therefore, the action can be formulated as the summation of the outer products $\mathcal{X}_t=\sum_{i=1}^m\mathbf{a}_i\circ \mathbf{b}_i \circ \mathbf{c}_i\in \mathbb{R}^{d_1\times d_2 \times d_3}$. Many practical actions can be modeled as such tensors, such as the [\textit{drug, target, disease}] recommendation in precision medicine, the [\textit{top, bottom, shoe}] recommendation in clothing website, the [\textit{custom, promotion mean, marketing channel}] in the advertising campaign, and the [\textit{type of applicant, type of position, level of interview}] in cohort selection. Furthermore, we denote the action/arm set as $\mathbb{X}\in \mathbb{R}^{d_1 \times d_2 \times d_3}$, and the action $\mathcal{X}_t\in \mathbb{X}$. We assume that the incurred reward $y_t$ follow a canonical exponential family such that:
$$
p(y_t\mid\mathcal{X}_t,\mathcal{W}^*)=h(y_t)\exp\left[\frac{y_t\langle \mathcal{X}_t,\mathcal{W}^*\rangle-b(\langle \mathcal{X}_t,\mathcal{W}^*\rangle)}{\phi}\right],
$$
$$
\mathbb{E}(y_t\mid\mathcal{X}_t,\mathcal{W}^*)=b^\prime(\langle \mathcal{X}_t,\mathcal{W}^*\rangle):=\mu(\langle \mathcal{X}_t,\mathcal{W}^*\rangle).
$$
where $\mathcal{W}^*\in \mathbb{R}^{d_1 \times d_2 \times d_3}$ is an unknown parameter  tensor with the tubal rank $r\ll \min\{d_1,d_2\}$, $h(\cdot)$ is a nuisance parameter, $b(\cdot)$ is a known and strictly
convex log-partition function, and $\mu(\cdot)$ is the corresponding real-valued, non-linear function called the (inverse) link function. Typical examples of $\mu(\cdot)$, and the corresponding regression models are summarized in the table below. 
\begin{table}[!ht]
	\caption{Typical examples of the generalized linear model}
	\label{table:generalized linear}
	\centering
	{\def\arraystretch{1.5}  
		\begin{tabular*}{0.85\textwidth}{ccc}
			\hline
			\hline
			Generalized linear model  & Link function $\mu(x)$ & Types of reward\\
			\hline
			Linear &$\mu(x)=x$&continuous\\
			Binary logistic &$\mu(x)=\frac{e^x}{1+e^x}$&binary\\
			Poisson &$\mu(x)=e^x$&count\\
			\hline
			\hline
		\end{tabular*}
	}
\end{table}

Note that the reward can also be represented as
$$
y_t=\mu\left(\langle \mathcal{X}_t,\mathcal{W}^*\rangle\right)+z_t,
$$
where $z_t$ is the independent sub-Gaussian random noise. For the generalized low-rank tensor contextual bandits problem, the goal is to design a sequential decision-making policy (algorithm) to maximize the total reward $\sum_{t=1}^{T}\mu\left(\langle \mathcal{X}_t,\mathcal{W}^*\rangle\right)$. Define the oracle policy as already knowing the parameter $\mathcal{W}^*$ and thus always choosing the best arm $\mathcal{X}^*=\arg\max_{\mathcal{X}\in \mathbb{X}}\mu\left(\langle \mathcal{X},\mathcal{W}^*\rangle\right)$. Then the regret that measures the difference in total reward between the decision-maker's policy and the oracle policy is \citep{audibert2009exploration}:
$$
R_T:=\sum_{t=1}^{T}\left[\mu\left(\langle \mathcal{X}^*,\mathcal{W}^*\rangle\right)-\mu\left(\langle \mathcal{X}_t,\mathcal{W}^*\rangle\right)\right].
$$

We begin by presenting a set of widely used standard assumptions concerning bandits models, the distribution of action features, and generalized linear models. These assumptions form the fundamental basis for our subsequent theoretical analysis. Furthermore, we provide a comprehensive explanation for the reasonability underlying these assumptions.

\begin{assumption}[Boundednes]\label{assump: bounded}
\begin{spacing}{1.5}
	The Frobenius norm of true parameter $\mathcal{W}^*$ and action feature tensors in $\mathbb{X}$ are bounded: $\|\mathcal{W}^*\|_F\le1$ and $\|\mathcal{X}\|_F\le1$;
 \end{spacing}
\end{assumption}
\vspace{-6mm}

This aligns with the standard bounded assumption in contextual bandits literature, as demonstrated in studies such as \citep{chu2011contextual, kirschner2019stochastic, lu2021low, shi2023high}. This can be easily accomplished through normalization.

\begin{assumption}[Sub-Gaussian action features]\label{assump: subgaussian}
\begin{spacing}{1.5}
	The vectorization of the action feature  $\operatorname{vec}\left(\mathcal { X }_t\right)$ follows a $\kappa_0=\frac{1}{\sqrt{d_1d_2d_3}}$-sub-Gaussian distribution;
\end{spacing}
\end{assumption}
\vspace{-6mm}

\begin{assumption}[Hessian]\label{assump:kappa}
\begin{spacing}{1.5}
	 $\widehat{H}(\mathcal{W}):=\frac{1}{n}\sum_{t=1}^{n}b^{\prime\prime}(\langle\mathcal{W},\mathcal{X}_t\rangle)\operatorname{vec}(\mathcal{X}_t)\operatorname{vec}(\mathcal{X}_t)^\top$, and define its expectation as $\mathbb{E}(\widehat{H}(\mathcal{W}))=H(\mathcal{W})$.
	Assume the matrix $H(\mathcal{W}^*)$ satisfies $\lambda_{\min }\left(H\left(\mathcal{W}^*\right)\right) \geq \kappa_l=\frac{1}{d_1d_2d_3}>0$;
 \end{spacing}
\end{assumption}
\vspace{-6mm}

There exists a sampling distribution $D$ over $\mathbb{X}$ such that $\lambda_{\min }(H\left(\mathcal{W}^*\right)) = \frac{1}{d_1 d_2d_3}$, and $\operatorname{vec}(\mathcal{X})$ is sub-Gaussian with parameter $\kappa_0 =\frac{1}{\sqrt{d_1 d_2 d_3}}$. The sampling distribution is easily found in many settings. For example, if we take $D$ and $\mathbb{X}$ as the uniform distribution and 
Euclidean unit ball/sphere in $\mathbb{R}^{d_1 d_2 d_3}$ respectively, we have (i) $\operatorname{vec}(\mathcal{X})$ is sub-Gaussian with parameter $\kappa_0 =\frac{1}{\sqrt{d_1 d_2 d_3}}$; (ii) $\lambda_{\min }(H\left(\mathcal{W}^*\right)) = \frac{1}{d_1 d_2d_3}$. This satisfies Assumptions \ref{assump: subgaussian} and \ref{assump:kappa}.
\begin{assumption}[Link functions]\label{assump: bx}
\begin{spacing}{1.5}
	The link function $\mu(x)$ satisfies $m\le \left|\mu^{\prime}(x)\right| \leq M<\infty$ for any $x \in \mathbb{R}$ and $\left|\mu^{\prime \prime}(x)\right| \leq|x|^{-1}$ for $|x|>1$.
 \end{spacing}
\end{assumption}
\vspace{-6mm}

This is a standard assumption in generalized linear models, as exemplified by works such as \cite{filippi2010parametric}, \cite{li2017provably}, and \cite{zhou2019learning}. It can be demonstrated that for the logistic link function, $\left|\mu^{\prime}(x)\right| \leq 1/4$ and $\left|\mu^{\prime \prime}(x)\right| \leq 1/4$ are satisfied.

\section{Efficient algorithm for the generalized low-rank tensor contextual bandits}
This section elaborates on the G-LowTESTR algorithm for generalized low-rank tensor contextual bandits problems and provides regret guarantees.
\subsection{The G-LowTESTR algorithm}
\begin{figure}[!ht]
	\centering
	\includegraphics[scale=.12]{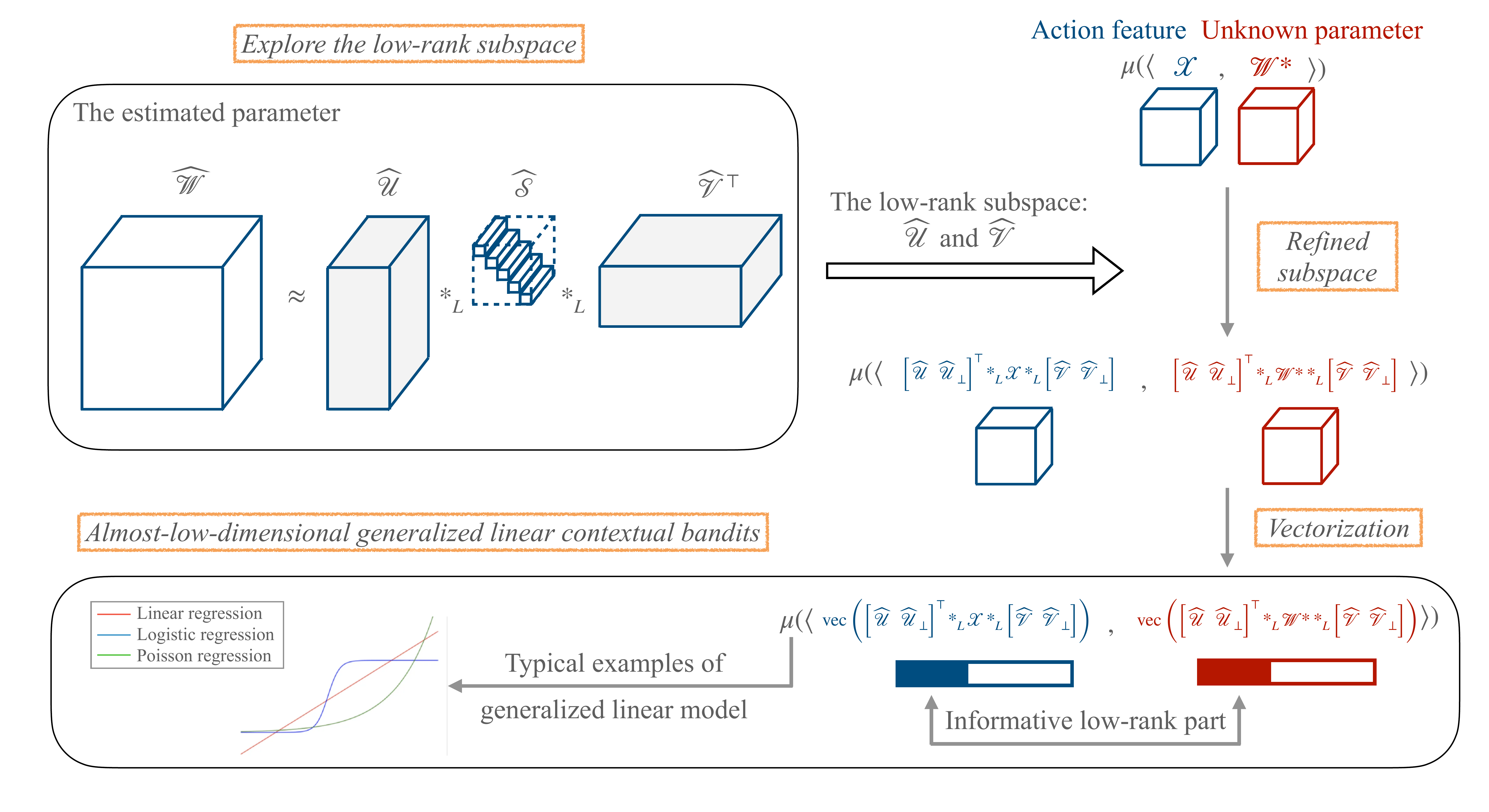}       
 	\caption{Illustration of G-LowTESTR}
        \label{figure:ltb}
\end{figure}

We introduce the ``Generalized Low-Rank Tensor Explore Subspace then Refine'' (G-LowTESTR) algorithm for addressing generalized low-rank tensor contextual bandits problems. Initially, G-LowTESTR estimates $\mathcal{W}^*$ by solving a tensor nuclear norm penalized minimization problem using raw data (action features $\mathcal{X}_{t}$ and reward $y_{t}$). This provides an approximate estimation of the low-rank subspace of $\mathcal{W}^*$. Subsequently, based on this estimated low-rank subspace, G-LowTESTR further transforms the problem into an almost low-dimensional generalized linear contextual bandits scenario, employing the LowGLM-UCB algorithm proposed by \cite{kang2022efficient}. The choice of LowGLM-UCB stems from its ability to achieve lower regret by projecting the arm set onto both the estimated subspace and its complementary counterpart, while adjusting weights in the complementary space. For a clearer comprehension of the G-LowTESTR algorithm, refer to Fig. \ref{figure:ltb}.

Specifically, our algorithm is inspired by \cite{lu2021low} and \cite{kang2022efficient}, employing a framework that explores the low-rank subspace first and then refines it. To accurately estimate this subspace, unlike other naive UCB algorithms, it's imperative to conduct an initial exploration to gather data for its estimation. What deserves emphasis is the thoughtfully designed duration of this exploratory phase (derived from subsequent theory), ensuring an adequately precise estimation of the low-rank subspace—indicating thorough extraction of low-rank structural information. This further implies that additional exploration of the low-rank subspace is superfluous in the subsequent refined stages. In essence, for all bandits algorithms utilizing low-rank structures, this approach of initially exploring the low-rank subspace followed by refinement is meaningful. 

\subsubsection{Explore the low-rank subspace}
We first collect data to find the accurate row and column subspace estimation of $\mathcal{W}^*$. Especially, at each time $t$, the learner selects an action $\mathcal{X}_{t}$ and receives the corresponding reward $y_{t}$. Given the horizon $T_1$, a series of action and reward pairs $\{\mathcal{X}_{t},y_{t}\}_{t=1}^{T_1}$ are accumulated. Considering the low-rank constraint of the unknown parameter $\mathcal{W}^*$, it is natural to estimate $\mathcal{W}^*$ by solving a tensor nuclear norm penalized minimization problem:
\begin{align}
	\widehat{\mathcal{W}}&=\arg\min_{\mathcal{W} \in \mathbb{R}^{d_1 \times d_2\times d_3}} \frac{1}{ T_1} \sum_{t=1}^{T_1}[b(\left\langle \mathcal{X}_{t}, \mathcal{W}\right\rangle)-y_t\left\langle \mathcal{X}_{t}, \mathcal{W}\right\rangle]+\lambda_{T_1}\|\mathcal{W}\|_*\notag\\
	&:= \arg\min_{\mathcal{W} \in \mathbb{R}^{d_1 \times d_2\times d_3}}L(\mathcal{W})+\lambda_{T_1}\|\mathcal{W}\|_*.
	\label{equation: tensor nuclear norm optimization}
\end{align}

To establish an accuracy rate for the M-estimator in high-dimensional data, it becomes necessary to possess restricted strong convexity (RSC) of the loss function \citep{negahban2012unified, raskutti2010restricted}. However, verifying the RSC of the loss function $L(\mathcal{W})$ in Equation (\ref{equation: tensor nuclear norm optimization}) is challenging due to the involvement of the unknown parameter tensor $\mathcal{W}$ in the second partial derivative (Hessian) of the loss function. In order to address this challenge, we extend the concept of local restricted strong convexity (LRSC) introduced by \cite{fan2018lamm} from matrices to tensors. Below, we present the LRSC definition.

\begin{definition}[Local restricted strong convexity (LRSC).]\label{def: LRSC}
\begin{spacing}{1.5}
	Given a constraint set $\mathcal{C} \subset \mathbb{R}^{d_1 \times d_2 \times d_3}$, a local neighborhood $\mathcal{N}$ of $\mathcal{W}^*$, a positive constants $\kappa_{\ell}$, if the following inequality holds for any $\Delta \in \mathcal{C}$ and $\mathcal{W} \in \mathcal{N}:$
	$$
	L(\mathcal{W}+\Delta)-L(\mathcal{W})-\langle\nabla L(\mathcal{W}), \Delta\rangle \geq \kappa_{\ell}\|\Delta\|_F^2 .
	$$
	we call the loss function $L(\cdot)$ is such that $\text{LRSC}\left(\mathcal{C}, \mathcal{N}, \kappa_{\ell}\right)$. 
 \end{spacing}
\end{definition}
\vspace{-6mm}

Note that the LRSC condition is defined by the Taylor-series expansion. Compared to locally strongly convex, which imposes a lower bound on the first-order Taylor error for all $\Delta$ in a neighborhood of the origin, the LRSC condition establishes a lower bound on the first-order Taylor error for $\Delta$ within a restricted set and the parameter $\mathcal{W}$ within the local neighborhood of the true parameter. 

Before analyzing the accuracy of the parameter estimator $\widehat{\mathcal{W}}$, we first prove a general error bound under the LRSC condition.

\begin{theorem}[Converence under LRSC]\label{thm: general bound} 
\begin{spacing}{1.5}
	Define subspaces $\mathcal{K}$ and $\widetilde{\mathcal{K}}^\perp$ as:
 \begin{small}
	\begin{align}
		&\mathcal{K}(\operatorname{lat}\left(\mathcal{U}^r\right),\operatorname{lat}\left(\mathcal{V}^r\right)):=\left\{\mathcal{A} \in \mathbb{R}^{d_1 \times d_2 \times d_3} \mid \operatorname{hor}(\mathcal{A}) \subseteq \operatorname{lat}\left(\mathcal{V}^r\right), \operatorname{lat}(\mathcal{A}) \subseteq \operatorname{lat}\left(\mathcal{U}^r\right)\right\}, \notag\\
		&\widetilde{\mathcal{K}}^{\perp}(\operatorname{lat}\left(\mathcal{U}^r\right),\operatorname{lat}\left(\mathcal{V}^r\right)):=\left\{\mathcal{A} \in \mathbb{R}^{d_1 \times d_2 \times d_3} \mid \operatorname{hor}(\mathcal { A }) \subseteq \operatorname{lat}\left(\mathcal{V}^r\right)^\perp, \operatorname{lat}(\mathcal{A}) \subseteq \operatorname{lat}\left(\mathcal{U}^r\right)^\perp\right\} .\notag
	\end{align}
 \end{small}
\noindent Let $\mathcal{C}(\mathcal{K},\widetilde{\mathcal{K}}^\perp,\mathcal{W}^*):=\left\{\Delta:\left\|\Delta_{\widetilde{\mathcal{K}}^\perp}\right\|_{*} \leq 3\left\|\Delta_{\widetilde{\mathcal{K}}}\right\|_{*}+4\left\|\mathcal{W}^*_{\mathcal{K}^\perp}\right\|_{*}=3\left\|\Delta_{\widetilde{\mathcal{K}}}\right\|_{*}\right\}$ be the restricted set, and
 \begin{small}
 $$\mathcal{N}:=\left\{\mathcal{W}\in\mathbb{R}^{d_1\times d_2 \times d_3}:\|\mathcal{W}-\mathcal{W}^*\|_F^2\le C_1 \frac{rd_3\lambda_n^2}{\ell\kappa_\ell^2},\mathcal{W}-\mathcal{W}^*\in\left(\mathcal{C}(\mathcal{K},\widetilde{\mathcal{K}}^\perp,\mathcal{W}^*), \mathcal{N}, \kappa_{\ell}\right)\right\},$$
 \end{small}
 if $L(\mathcal{W})$ satisfy the $\text{LRSC}\left(\mathcal{C}(\mathcal{K},\widetilde{\mathcal{K}}^\perp,\mathcal{W}^*), \mathcal{N}, \kappa_{\ell}\right)$ condition stated in Definition \ref{def: LRSC}, that is,
 \begin{small}
	$$
	L(\mathcal{W}+\Delta)-L(\mathcal{W})-\langle\nabla L(\mathcal{W}), \Delta\rangle \geq \kappa_{\ell}\|\Delta\|_F^2 , \text{ for any } \Delta \in \mathcal{C}(\mathcal{K},\widetilde{\mathcal{K}}^\perp,\mathcal{W}^*) \text{ and } \mathcal{W} \in \mathcal{N}.
	$$
 \end{small}
	Then condition on the event $G:=\left\{\left\|\frac{1}{n} \sum_{t=1}^n [b^\prime\left(\langle\mathcal{W}^*,\mathcal{X}_t\rangle\right)-y_t]\mathcal{X}_t\right\|_{2} \leq \frac{\lambda_n}{2}\right\}$, any optimal solution $\widehat{\mathcal{W}}$ to Equation (\ref{equation: tensor nuclear norm optimization}) satisfies the bound below:
	$$
	\left\|\widehat{\mathcal{W}}-\mathcal{W}^*\right\|_F^2 \leq C_1 \frac{rd_3\lambda_n^2}{\ell\kappa_\ell^2} ,
	$$
	where  $r=\operatorname{rank}_t\left(\mathcal{W}^*\right)$, and $C_1$ is constant.
 \end{spacing}
\end{theorem}
\vspace{-6mm}

Based on the above general error bound, the following corollary shows that the solution $\widehat{\mathcal{W}}$ of (\ref{equation: tensor nuclear norm optimization}) is guaranteed to converge to $\mathcal{W}^*$ at a fast rate.
\begin{corollary}\label{error bound final}
\begin{spacing}{1.5}
	Based on Assumptions \ref{assump: bounded}-\ref{assump: bx}, any optimal solution of the nuclear norm optimization problem (\ref{equation: tensor nuclear norm optimization}) using $n=\Omega((d_1+d_2)d_3)$ and $\lambda_n^2=C_3\frac{\ell \log d_3}{arnd_3^2\min\{d_1,d_2\}(1-\gamma)^2}$ satisfies:
	$$
	\left\|\widehat{\mathcal{W}}-\mathcal{W}^*\right\|_F^2 \lesssim  \frac{(d_1+d_2)^3 d_3\log d_3}{an(1-\gamma)^2},
	$$
	with probability $1-2\delta$, where $\gamma=\frac{\sigma_1-\sigma_{\lceil ard_3\rceil}}{\sigma_1}\ (a\in[0,1])$ measures the ``gap" between the first and $\lceil ard_3\rceil$-th singular values of $\bar{\mathcal{X}}_t$.
 \end{spacing}
\end{corollary}
\vspace{-6mm}

Equipped with the convergence guarantee of the estimator $\widehat{\mathcal{W}}$, we further run a transformed t-SVD step to obtain the row and column subspace estimators. Denote the skinny transformed t-SVD of $\mathcal{W}^*$ as $\mathcal{W}^*=\mathcal{U}^**_L\mathcal{S}^**_L\mathcal{V}^{*\top}$. If $\widehat{\mathcal{U}}$ ($\widehat{\mathcal{V}}$) and $\mathcal{U}^*$ ($\mathcal{V}^*$) are close, $\widehat{\mathcal{W}}$ is an accurate estimate. The distance is determined by the Frobenius norm of the difference between the two tensors. See the following corollary for details.
\begin{corollary}\label{cor: 17}
\begin{spacing}{1.5}
	By solving the convex problem in Equation (\ref{equation: tensor nuclear norm optimization}),we can compute $\widehat{\mathcal{W}}$ that serves as an estimate of the tensor $\mathcal{W}^*$. After the subspace exploration stage of G-LowTESTR  with $T_1=\Omega((d_1+d_2)d_3)$ and $\lambda_{T_1}^2=\frac{\ell\log d_3}{T_1d_3^2\min\{d_1,d_2\}}$ satisfying the conditions of Corollary \ref{error bound final}, we have, with probability at least $1-2\delta$,
	\begin{align}
		\left\|\widehat{\mathcal{U}}_{\perp}^\top *_L \mathcal{U}^*\right\|_F\left\|\widehat{\mathcal{V}}_{\perp}^\top *_L \mathcal{V}^*\right\|_F \leq \frac{\left\|\mathcal{W}^*-\widehat{\mathcal{W}}\right\|_F^2}{\omega_{min}^2} \lesssim\frac{(d_1+d_2)^3 d_3\log d_3}{aT_1(1-\gamma)^2 \omega_{min}^2}:=\gamma\left(T_1\right),\notag
	\end{align}
	where $\omega_{min}>0$ is the smallest singular value of $\bar{\mathcal{W}}^*$.
 \end{spacing}
\end{corollary}
\vspace{-6mm}

\subsubsection{Generalized linear contextual bandits with refined subspace}
Then we explain how to exploit the estimated low-rank subspaces. The core idea is to transform the generalized low-rank tensor contextual bandits problem with subspace estimation into an almost low-dimensional generalized linear contextual bandits problem, and to achieve a lower regret by imposing penalties on the components complementary to the subspaces $\widehat{\mathcal{U}}$ and $\widehat{\mathcal{V}}$.

\begin{itemize}
	\item Transform the generalized low-rank tensor contextual bandits problem into an almost low-dimensional generalized linear contextual bandits.
\end{itemize}

Denote $\mathcal{M}=\left[\widehat{\mathcal{U}} \ \widehat{\mathcal{U}}_{\perp}\right]^{\top}*_L \mathcal{W}^**_L\left[\widehat{\mathcal{V}}\  \widehat{\mathcal{V}}_{\perp}\right]$ as the rotated version of $\mathcal{W}^*$, then we have
$$
\mathcal{W}^* =\left[\widehat{\mathcal{U}}\ \widehat{\mathcal{U}}_{\perp}\right]*_L \mathcal{M}*_L\left[\widehat{\mathcal{V}}\ \widehat{\mathcal{V}}_{\perp}\right]^{\top}, $$
\begin{small}
$$
\langle \mathcal{X},\mathcal{W}^*\rangle =\left\langle \mathcal{X},\left[\widehat{\mathcal{U}}\ \widehat{\mathcal{U}}_{\perp}\right]*_L \mathcal{M}*_L\left[\widehat{\mathcal{V}}\ \widehat{\mathcal{V}}_{\perp}\right]^{\top} \right\rangle 
=\left\langle\left[\widehat{\mathcal{U}} \ \widehat{\mathcal{U}}_{\perp}\right]^\top*_L \mathcal{X}*_L\left[\widehat{\mathcal{V}} \ \widehat{\mathcal{V}}_{\perp}\right], \mathcal{M}\right\rangle.
$$
\end{small}

Therefore, the generalized low-rank contextual tensor contextual bandits problem with $\mathcal{W}^*$ and arm sets $\mathbb{X}$ can be considered equivalent to the one with $\mathcal{M}$ and arm sets $\mathbb{X}^{\prime}:=\left\{\left[\widehat{\mathcal{U}} \ \widehat{\mathcal{U}}_{\perp}\right]^\top*_L \mathcal{X}*_L\left[\widehat{\mathcal{V}} \ \widehat{\mathcal{V}}_{\perp}\right]: \mathcal{X} \in \mathbb{X}\right\}$, which can be further transformed into the generalized linear contextual bandits with the unknown vector $\theta^*=\operatorname{vec}(\mathcal{M})$ and arm sets
\begin{align}
	\mathbb{X}_{\text {vec }}^{\prime}:&=\{\operatorname{vec}\left(\mathcal{X}_{1: r, 1: r,1:d_3}^{\prime}\right) ; \operatorname{vec}\left(\mathcal{X}_{r+1: d_1, 1: r,1:d_3}^{\prime}\right) ; \notag\\
	&\operatorname{vec}\left(\mathcal{X}_{1: r, r+1: d_2,1:d_3}^{\prime}\right) ; \operatorname{vec}\left(\mathcal{X}_{r+1: d_1, r+1: d_2,1:d_3}^{\prime}\right): \mathcal{X}^{\prime} \in \mathbb{X}^{\prime}\} .\label{armset}
\end{align}

Correspondingly, $\theta^*=\operatorname{vec}(\mathcal{M})$ can also be rearranged as
\begin{align}
	\theta_{1: k}^{*}=\left[\operatorname{vec}\left(\mathcal{M}_{1: r, 1: r,1:d_3}\right) ; \operatorname{vec}\left(\mathcal{M}_{r+1: d_1, 1: r,1:d_3}\right);
	\operatorname{vec}\left(\mathcal{M}_{1: r, r+1: d_2,1:d_3}\right)\right],\notag
\end{align}
\begin{align}
	\theta_{k+1: p}^{*}=\operatorname{vec}\left(\mathcal{M}_{r+1: d_1, r+1: d_2,1:d_3}\right),\label{theta}
\end{align}
where $k=(d_1+d_2)d_3r-d_3r^2$, and the goal of we arrange in this way is to use the results of the estimated subspaces obtained early. 
\begin{itemize}
	\item Apply almost low-dimensional GLM-UCB (LowGLM-UCB) for the refined generalized linear contextual bandits problem.
\end{itemize}

In the almost low-dimensional generalized linear contextual bandits problem of $d_1d_2d_3$-dimension, the player chooses an arm $x_t$ at time $t$ from an arm set $\mathbb{X}_{\operatorname{vec}}^{\prime}  \subseteq \mathbb{R}^{d_1d_2d_3}$ and observes a noisy reward $y_t=\left\langle x_t, \theta^*\right\rangle+z_t$, where $z_t$ is the sub-Gaussian random noise. We assume that $\|x_t\|_2 \leq 1$ for all $x_t \in \mathbb{X}_{\operatorname{vec}}^{\prime} $, $\left\|\theta^*\right\|_2 \leq 1$, and $\left\|\theta_{k+1: p}^*\right\|_2 \leq B_{\perp}$ for the known constant $B_{\perp}$ (ideally $\ll 1$). This means that all-but-$k$ dimensions of $\theta^*$ are close to zero. In our context, the property $\left\|\theta_{k+1: p}^*\right\|_2^2 =\sum_{i>r \wedge j>r} \mathcal{M}_{i jk}^2 =\left\|\widehat{\mathcal{U}}_{\perp}^{\top}*_L\left(\mathcal{U}^{*} *_L\mathcal{S}^{*}*_L \mathcal{V}^{* \top}\right)*_L \widehat{\mathcal{V}}_{\perp}\right\|_F^2 \leq\ell^2\left\|\widehat{\mathcal{U}}_{\perp}^{\top}*_L \mathcal{U}^*\right\|_F^2\left\|\widehat{\mathcal{V}}_{\perp}^{\top}*_L \mathcal{V}^{*}\right\|_F^2 $ implies that $B_{\perp}=\ell\left\|\widehat{\mathcal{U}}_{\perp}^{\top}*_L \mathcal{U}^*\right\|_F\left\|\widehat{\mathcal{V}}_{\perp}^{\top}*_L \mathcal{V}^{*}\right\|_F$. Then we can use the almost-low-dimensional GLM-UCB (LowGLM-UCB) algorithm proposed by \cite{kang2022efficient} to deal with the almost-low-dimensional bandits problem. This algorithm penalizes those components that are complementary to $\widehat{\mathcal{U}}$ and $\widehat{\mathcal{V}}$, thus enjoying a lower regret bound. The processes of G-LowTESTR and LowGLM-UCB are illustrated in Algorithms \ref{alg:LowTESTR} and \ref{alg:LowOFUL}, respectively.
\begin{algorithm}[!htb]
		\small
  \begin{spacing}{1.5}
	\caption{Generalized Low-rank Tensor Explore Subspace Then Refine (G-LowTESTR)}
	\label{alg:LowTESTR}
	\begin{algorithmic}[1]
		\REQUIRE $\mathbb{X}$, $T$, $T_1$, $r$, $\omega_{\min}$, $D$, $B_{\perp},\ \lambda,\ \lambda_{\perp}$.
		\STATE For $t\in [T_1]$, choose action $\mathcal{X}_{t} \in \mathbb{X}$ according to distribution $D$ and receive reward $y_{t}$.
		\STATE Compute $\widehat{\mathcal{W}}$ by minimizing the tensor nuclear norm problem as stated in (\ref{equation: tensor nuclear norm optimization}). 
		\STATE Denote the transformed t-SVD of $\widehat{\mathcal{W}}$ as $\widehat{\mathcal{W}}=\left[\widehat{\mathcal{U}} \ \widehat{\mathcal{U}}_{\perp}\right]*_L \widehat{\mathcal{S}}*_L \left[\widehat{\mathcal{V}}\  \widehat{\mathcal{V}}_{\perp}\right]^{\top}$, where $\widehat{\mathcal{U}}$ and $\widehat{\mathcal{U}}$ are the first $r$ column and row slices respectively.
		\STATE Transform the generalized low-rank tensor contextual bandits into the generalized linear contextual bandits with the unknown vector $\theta^*$ (i.e., (\ref{theta})) and the arm set $\mathbb{X}_{\operatorname{vec}}^{\prime}$ (i.e., (\ref{armset})). 
		\STATE  Execute LowGLM-UCB (Algorithm \ref{alg:LowOFUL}) for $T_2=T-T_1$ rounds, with arm set $\mathbb{X}_{\operatorname{vec}}^{\prime}$, and the low dimension $k=\left(d_1+d_2\right) d_3r-d_3r^2$.
	\end{algorithmic}
 \end{spacing}
\end{algorithm}
\begin{algorithm}[!htb]
		\small
  \begin{spacing}{1.5}
	\caption{LowGLM-UCB \citep{kang2022efficient}}
	\label{alg:LowOFUL}
	\begin{algorithmic}[1]
		\REQUIRE $T_2,\ k$, $B_{\perp},\ \lambda,\ \lambda_{\perp}$, arm set $\mathbb{X}_{\operatorname{vec}}^{\prime} \subset \mathbb{R}^{d_1  d_2  d_3}$ and failure rate $\delta$.
		\STATE $\Lambda=\operatorname{diag}\left(\lambda, \ldots, \lambda, \lambda_{\perp}, \ldots, \lambda_{\perp}\right)$ where $\lambda$ occupies the first $k$ diagonal entries, and initialize $V_1\left(m\right)=\sum_{i=1}^{T_1} \operatorname{vex}(\mathcal{ X }_{s, i})  \operatorname{vex}(\mathcal{ X }_{s,i}) ^{\top}+\Lambda / m$ by the data collected in the subspace exploration.
		\FOR{$t=1, \ldots, T_2$}
		\STATE Estimate $\widehat{\theta}_t$ by 
		$
		\sum_{i=1}^{T_1} \mu\left( \operatorname{vex}(\mathcal{ X }_{s,i}) ^{\top} \widehat{\theta}_t\right)  \operatorname{vex}(\mathcal{ X }_{s,i}) +\sum_{i=1}^{t-1} \mu\left( x_{i} ^{\top} \widehat{\theta}_t\right)  x_i+\Lambda \widehat{\theta}_t
		=\sum_{i=1}^{T_1} y_{s, i} \operatorname{vex}(\mathcal{ X }_{s,i})+\sum_{i=1}^{t-1} y_i x_i.
		$
		\STATE Pull arm $ x_t=\arg \max _{x \in \mathbb{X}_{\operatorname{vec}}^{\prime}}\left\{\mu\left(x^{\top} \widehat{\theta}_t\right)+\alpha_t(\delta/2)\|x\|_{V_t^{-1}\left(m\right)}\right\} $, and receive reward $y_t$, where $\alpha_t(\delta):=\frac{M}{m}\left( \sqrt{k \log \left(1+\frac{m  (t+T_1)}{k \lambda}\right)+\frac{m  (t+T_1)}{\lambda_{\perp}}-\log \left(\delta^2\right)}+\sqrt{m}\left(\sqrt{\lambda} +\sqrt{\lambda_{\perp}} B_{\perp}\right)\right)$, and $B_{\perp}:=\ell\gamma\left(T_1\right)=\Theta(\ell\frac{(d_1+d_2)^3 d_3\log d_3}{aT_1(1-\gamma)^2 \omega_{min}^2})$.
		\STATE Update $V_{t+1}\left(m\right) \longleftarrow V_t\left(m\right)+x_t x_t^{\top}$.
		\ENDFOR
	\end{algorithmic}
 \end{spacing}
\end{algorithm}
\subsection{Regret analysis}
This subsection provides the upper bound of regret for the G-LowTESTR algorithm and compare it with the regret bounds of existing vector or low-rank matrix contextual bandits algorithms.
\begin{theorem}[Regret of G-LowTESTR]\label{thm:regret}
\begin{spacing}{1.5}
	Suppose we run the subspace exploration stage of G-LowTESTR (Algorithm \ref{alg:LowTESTR}) with $T_1=\frac{(d_1+d_2)^{\frac{3}{2}}\sqrt{d_3\ell\log d_3T}}{\omega_{min}\sqrt{a}(1-\gamma)}$  and $\lambda_{T_1}^2=C_3\frac{\ell \log d_3}{arT_1d_3^2\min\{d_1,d_2\}(1-\gamma)^2}$. We invoke LowGLM-UCB (Algorithm \ref{alg:LowOFUL}) with $k=\left(d_1+d_2\right) d_3r-d_3r^2,\ \lambda_{\perp}=\frac{mT}{k \log \left(1+\frac{mT}{k\lambda}\right)}, B=1$, $B_{\perp}=\ell\gamma\left(T_1\right)$, and the rotated arm sets $\mathbb{X}_{\text {vec }}^{\prime}$ defined in Algorithm \ref{alg:LowTESTR}, the overall regret of G-LowTESTR (Algorithm \ref{alg:LowTESTR}) is, with probability at least $1-2\delta$,
	$$
	R_T=\widetilde{O}\left(\frac{(d_1+d_2)^{\frac{3}{2}}\sqrt{d_3\ell T}}{\sqrt{a}(1-\gamma)\omega_{min}} \right) ,
	$$
	where $\omega_{min}$ is the smallest singular value of $\bar{\mathcal{W}}^*$, and $\gamma=\frac{\sigma_1-\sigma_{\lceil ard_3\rceil}}{\sigma_1}\ (a\in[0,1])$ measures the ``gap" between the first and $\lceil ard_3\rceil$-th singular values of $\bar{\mathcal{X}}_t$.
 \end{spacing}
\end{theorem}
\vspace{-6mm}

By Theorem \ref{thm:regret}, we achieve an enhanced regret bound of $
\widetilde{O}\left(\frac{(d_1+d_2)^{\frac{3}{2}}\sqrt{d_3\ell T}}{\sqrt{a}(1-\gamma)} \right)
$ that correlates with the singular value gap $\gamma$ of the action features. This gap is associated with the parameter ``$a$" where $a\in[0,1]$. We define ``$a$" as the proportion of singular values that share the same order as the maximum singular value among all singular values. The subsequent empirical evidence indicates that ``$a$" is generally sizable in many cases. As a result, both the singular value gap and the parameter ``$a$" in our bound tend to be larger, making them comparable to the tubal rank. It's worth noting that, following the approach of \cite{lu2019low} and \cite{song2020robust}, the invertible linear transform matrix is often chosen as Discrete Fourier Transform (DFT), Discrete Cosine Transform (DCT), or Random Orthogonal Matrix (ROM). Notably, DCT and ROM are unitary transform matrices (i.e., $\ell=1$), and for the DFT matrix, it can be made unitary by multiplying it with an appropriate factor. Hence, in practice, $\ell=1$ is commonly set, thereby facilitating the subsequent comparison. 

A straightforward approach for handling tensor data is the vectorization or matricization technique. That is, we can utilize the generalized linear or low-rank matrix contextual bandits to address the problems posed by generalized low-rank tensor contextual bandits. Consequently, to gauge the enhancement in regret bound achieved by directly incorporating the low-rank structure in tensor data, we compare the regret bound of our proposed G-LowTESTR, given by $
\widetilde{O}\left(\frac{(d_1+d_2)^{\frac{3}{2}}\sqrt{d_3\ell T}}{\sqrt{a}(1-\gamma)} \right) 
$, with that of existing methods, namely $
\widetilde{O}\left(d_1d_2d_3\sqrt{T}\right)
$ for GLM-UCB \citep{abbasi2011improved}, $
\widetilde{O}\left((d_2d_3+d_1)^{3/2}\sqrt{\operatorname{rank}(\mathcal{W}^*_{(1)}) T}  \right)
$  for LowESTR \citep{lu2021low}, and $
\widetilde{O}\left((d_2d_3+d_1)\operatorname{rank}(\mathcal{W}^*_{(1)}) \sqrt{T}  \right)
$ for G-ESTT \citep{kang2022efficient}. First, G-LowTESTR exhibits a substantial advantage over GLM-UCB and LowESTR in terms of regret bounds from the perspective of data feature dimensions. Second, G-LowTESTR uses weaker assumptions and can attain comparable regret bounds with G-ESTT in many cases. Our algorithm's superior performance can be attributed to the fact that directly reducing the order of the tensor may result in the loss of its intrinsic structure, including the complex interaction relationships among its slices.

Currently, there is limited theoretical analysis on tensor contextual bandits problems, with only the study by \cite{shi2023high}. They focus on a specific case of the generalized linear model, namely the linear case, and propose the TOFU algorithm. For this linear tensor contextual bandits model, our regret bound is comparable to the $
\widetilde{O}\left((d_2d_3+d_1)\operatorname{rank}(\mathcal{W}^*_{(1)}) \sqrt{T} \right)
$ for TOFU. Furthermore, in comparison to their model and algorithm based on the Tucker decomposition, our transformed t-product framework exhibits higher computational efficiency and superior data compression capabilities.

\section{Experiments}
In this section, we conduct a series of experiments to showcase the performance of our developed algorithm, i.e., G-LowTESTR. We focus on three regression models: linear tensor regression, binary Logistic tensor regression, and Poisson tensor regression, as examples. We start by showcasing the optimal regret achieved by our algorithm using synthetic datasets. Following that, we demonstrate the algorithm's superior performance in real-world applications. All experimental evaluations are executed using Matlab 2022(a) on a computer equipped with an Intel(R) Xeon(R) Gold 5120 CPU @ 2.20GHz. A comprehensive list of the algorithms used for comparison is provided below for reference.
\begin{itemize}
	\item G-LowTESTR (this paper): the generalized low-rank tensor contextual bandits algorithm.
	\item GLM-UCB \cite{abbasi2011improved}: the generalized linear contextual bandits algorithm.
	\item LowESTR \citep{lu2021low}: the generalized low-rank matrix contextual bandits algorithm.
	\item G-ESTT \citep{kang2022efficient}: the generalized low-rank matrix contextual bandits algorithm based on Stein method.
	\item TOFU \citep{shi2023high}: the low-rank tensor contextual bandits algorithm based on the Tucker decomposition (the linear case).
\end{itemize}
\begin{remark}
\begin{spacing}{1.5}
In our context, the ablative experiments involve the generalized linear contextual bandits algorithm GLM-UCB, which doesn't utilize tensor low-rank structures, and the low-rank tensor contextual bandits algorithm TOFU, which doesn't consider generalized linear models.
\end{spacing}
\end{remark}
\vspace{-6mm}

\subsection{Synthetic data experiments}\label{section:syn}
We generate the true parameter tensor $\mathcal{W}^* =\mathcal{P} *_L \mathcal{Q}$, where the entries of $\mathcal{P} \in \mathbb{R}^{d_1 \times r \times d_3}$ and $\mathcal{Q} \in \mathbb{R}^{r \times d_2 \times d_3}$ are independently sampled from the Gaussian distribution $N(0,1)$. For the arms, we draw 100 vectors from $N\left(0, I_{d_1 d_2 d_3}\right)$ and standardize them, and then reshape all standardized $d_1 d_2 d_3$-dimensional vectors to $d_1 \times d_2 \times d_3$ tensors. We use these tensors to build up the arm set $\mathbb{X}$. For each arm $\mathcal{X}_t \in \mathbb{X}$, the reward is generated by $y_t=\mu\left(\left\langle \mathcal{X}_t, \mathcal{W}^*\right\rangle\right)+\eta_t$. We adopt the common generalized linear models, and the process of generating rewards for each model is as follows.
\begin{itemize}
	\item [(1)] Linear case: $y_t=\left\langle \mathcal{X}_t, \mathcal{W}^*\right\rangle+\eta_t$, where $\eta_t \sim N\left(0,0.01^2\right)$. 
	\item [(2)] Binary logistic case: $
	y_t \sim \operatorname{Logistic}\left(p_t\right) \text { with } p_t=\frac{1}{1+e^{-\left\langle\mathcal{X}_{t}, \mathcal{W}^*\right\rangle}}
	$.
     \item [(3)] Poisson case: $y_t \sim \operatorname{Poisson}\left(\lambda_t\right)$ with $\lambda_t=e^{\left\langle\mathcal{X}_{t}, \mathcal{W}^*\right\rangle}$.
\end{itemize}

\begin{figure}[h]
	\centering
	\subfigure[Linear case]
	{
		\begin{minipage}[b]{0.308\linewidth}
			\includegraphics[width=1.1\linewidth]{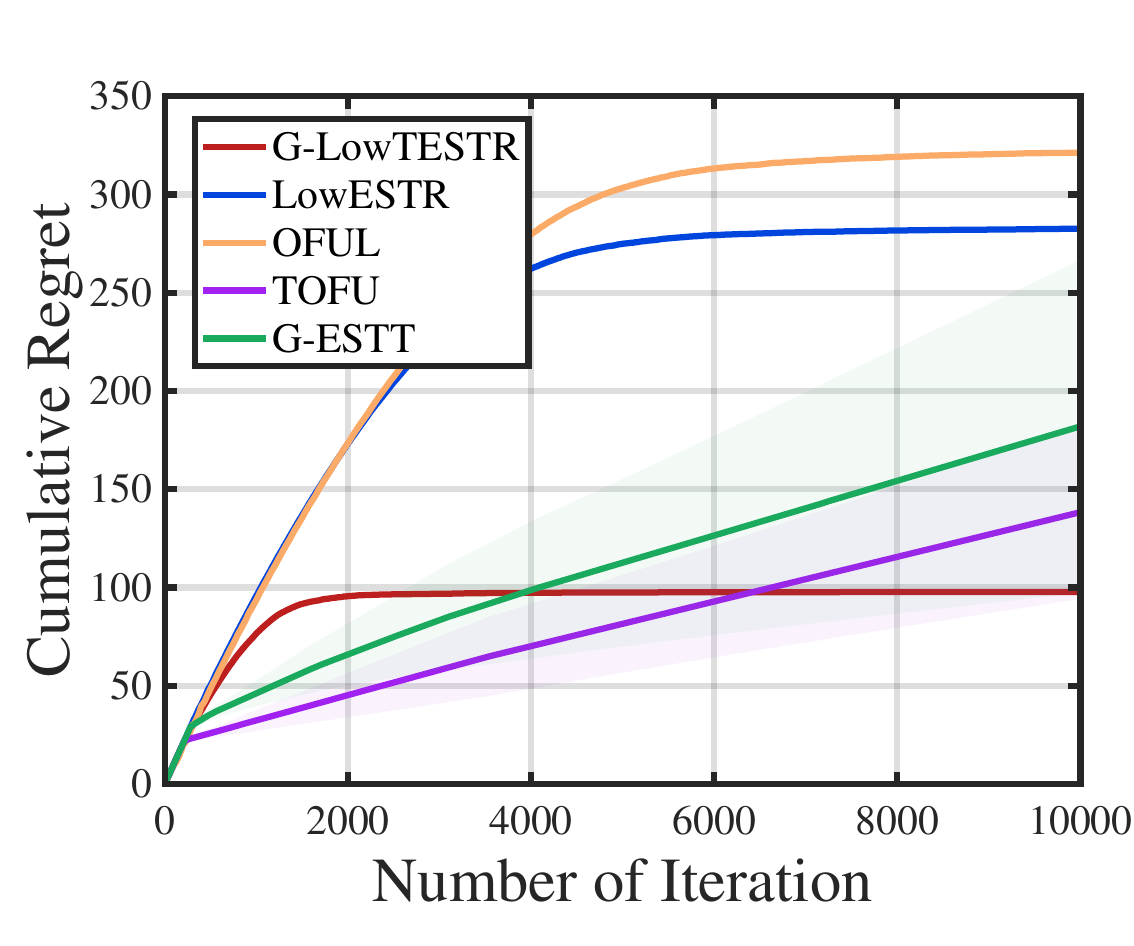}
		\end{minipage}
	}
  \hfill
	\subfigure[Binary logistic case]
{
	\begin{minipage}[b]{0.307\linewidth}
		\includegraphics[width=1.1\linewidth]{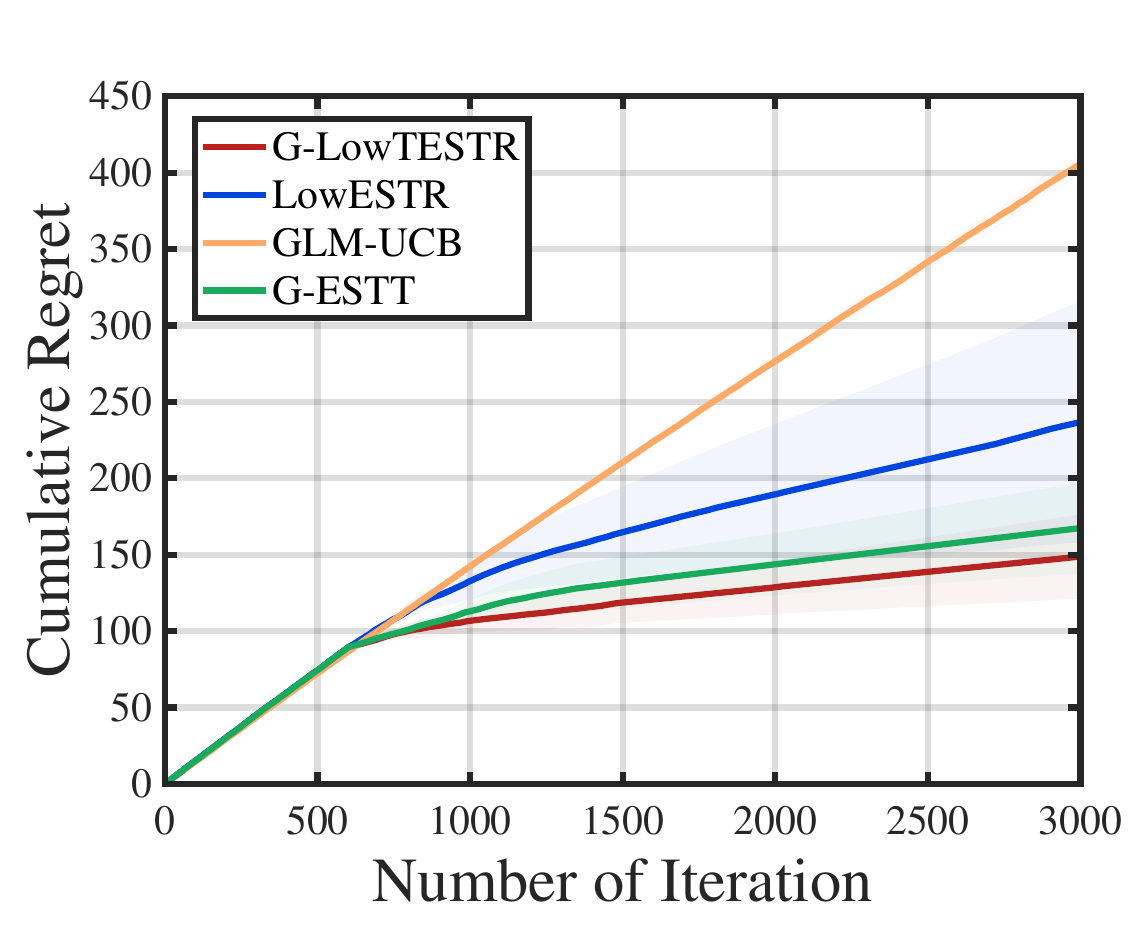}
	\end{minipage}
}
 \hfill
\subfigure[Poisson case]
{
	\begin{minipage}[b]{0.307\linewidth}
		\includegraphics[width=1.1\linewidth]{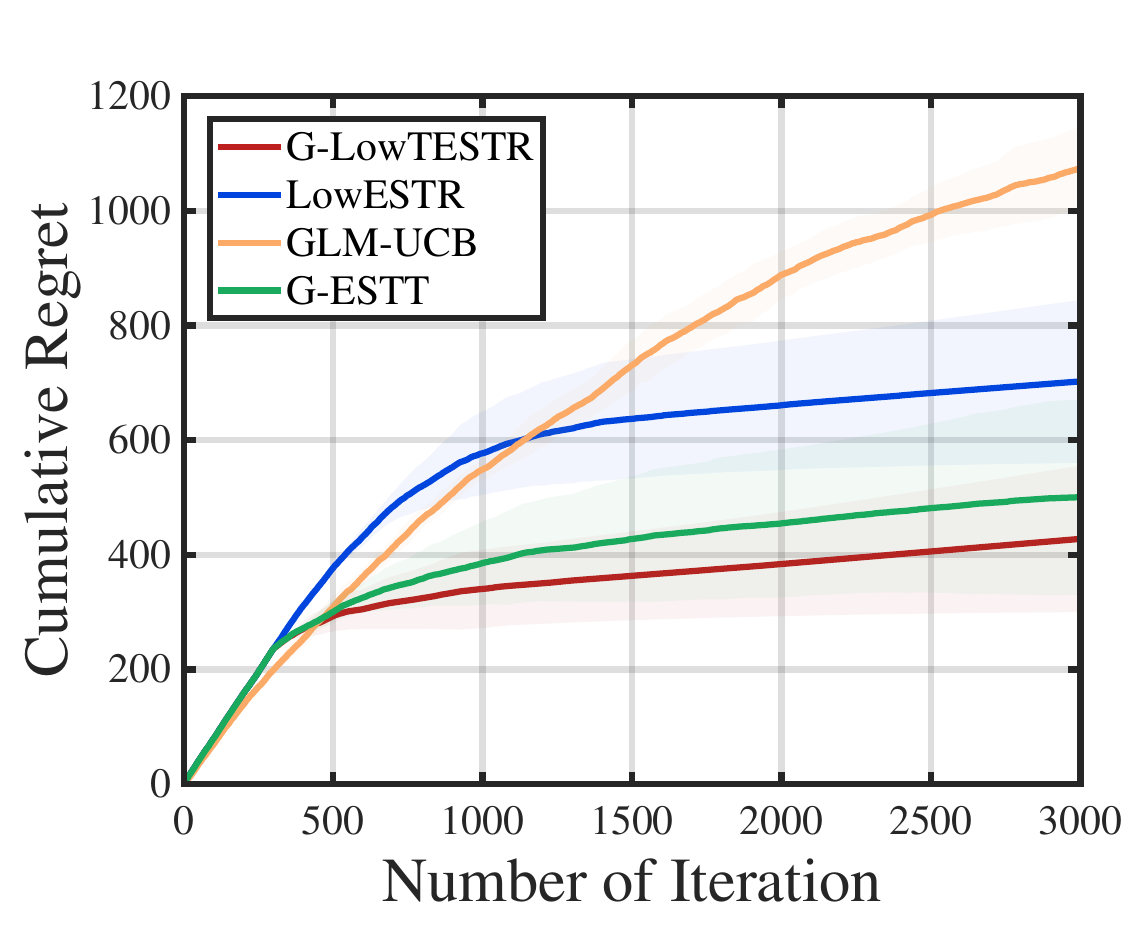}
	\end{minipage}
}
\caption{Comparison analysis with other algorithms in synthetic data} \label{fig:syn2}
\end{figure}

We conducted our simulation with $d_1 = d_2 = 10$, $d_3 = 3$, $r = 1$, and $T = 10000,\ 3000$, respectively. For each simulation setup, we repeated the process 10 times to compute the average regrets and their 1-standard deviation confidence intervals at each time step. The results are presented in Fig. \ref{fig:syn2}. It is evident that the proposed G-LowTESTR outperforms other algorithms in terms of cumulative regret. Additionally, we observe that in the initial iterations, G-LowTESTR achieves a steeper decline in the curve within a shorter period, indicating the exploration stage's enhanced effectiveness. In essence, our G-LowTESTR leverages the low-rank structure more adeptly to precisely estimate the parameter $\mathcal{W}^*$ during the exploration stage, thus offering improved guidance for subsequent decision-making. This advantage arises from the fact that reducing the order of tensor data can compromise its inherent low-rank structure. Hence, when using matrix-related algorithms, the absence of a valid low-rank structure hinders the ability to obtain accurate parameter estimators during the exploration phase.

\subsection{Real-world applications}\label{section:real}
This subsection primarily aims to demonstrate the superiority of algorithms through three practical applications: precision medicine, movie recommendation, and topic modeling on multiway publications. We'll start by providing background information on each application and details about the datasets used. Following that, we will analyze the implications of the experimental results.
\begin{itemize}
    \item [(1)] Linear case: Precision medicine
\end{itemize}

As a real-world application explored in this paper, we consider the problem of choosing the best treatment for patients with cancer. When aiming to choose the best [cell line, drug, dose] triplet pairing, it is necessary to compute the inner product of the feature tensors of all candidate triplet pairings with the estimated parameter at each time. Subsequently, the triplet pair exhibiting the highest inner product would be considered the current optimal choice among the available options. Specifically, whenever a new patient is admitted or a new drug is developed, we collect the relevant features of the patient and the drug at various doses. Subsequently, our model is utilized to calculate the expected rewards associated with administering the drug to the patient at different dosage levels. This enables us to determine the [cell line, drug, dose] combination that yields the most effective treatment. In essence, for a specific patient, we can provide a recommendation for the optimal [drug, dose] pairing, thereby facilitating informed and responsible decision-making in their treatment.

To verify the effectiveness of our method, we use the popular dataset from large-scale anti-cancer drug screens the Cancer Cell Line Encyclopedia (CCLE) dataset\footnote{CCLE dataset, https://depmap.org/portal/download/}. In the experiment, we first complete missing values with the mean of the columns and then normalize them. We consider the [\textit{cell line, target, dose}] triple as an arm and the drug sensitivity data as a reward. We model the reward as the tensor $\mathcal{M}\in \mathbb{R}^{n_1 \times n_2 \times n_3}$. Then, let $\mathcal{W}^* = \mathcal{P} *_L \mathcal{Q}^\top$, where $\mathcal{P}$ and $\mathcal{Q}$ are independently sampled from standard normal distribution. In order to generate the arm set $\mathbb{X}$, we first set the feature numbers of cells and drugs to 8 respectively, i.e., $d_1=d_2=8$, and build cell set $\mathcal{U}\in \mathbb{R}^{n_1 \times d_1 \times n_3}$ and drug set $\mathcal{V}\in \mathbb{R}^{n_2 \times d_2 \times n_3} $ by $\mathcal{B} = \mathcal{U} *_L \mathcal{S} *_L \mathcal{P}^{\dagger}$ and $\mathcal{D} = \mathcal{V}^\top *_L \mathcal{Q}^{\dagger}$, where $\mathcal{M}=\mathcal{U} *_L \mathcal{S} *_L \mathcal{V}^\top$ be the transformed t-SVD of the drug sensitivity data tensor $\mathcal{M}$. As a consequence, each arm $\mathcal{X} \in \mathbb{R}^{d_1 \times d_2 \times n_3}$ is equal to $\mathcal{U}^{\prime \top} *_L \mathcal{V}^\prime$, where $\mathcal{U}^\prime\in \mathbb{R}^{1 \times d_1 \times n_3}$, $\mathcal{V}^\prime\in \mathbb{R}^{1 \times d_2 \times n_3}$ are chosen from $\mathcal{U}$, $\mathcal{V}$. Therefore, we process 8568 arms, and compare the cumulative regret of our algorithm with other matrix-based and tensor-based algorithms.

\begin{itemize}
    \item [(2)] Binary logistic case: Movie recommendation
\end{itemize}

We consider the movie recommendation problem on the IMDB dataset\footnote{IMDB dataset, http://komarix.org/ac/ds/}. IMDB is a collection of links gathered from the Internet Movie Database and it contains four objective types: users, movies, directors and actors. Therefore, for each user, the problem we need to address is the ternary matching of [\textit{movie}, \textit{director}, \textit{actor}], where feature $\mathcal{X}$ represents the interaction feature of this triplet, and the reward $y$ denotes the link relationship between the triplets (1 indicates a link, otherwise 0). Specifically, we grouped 443 movies into 21 categories based on movie genres, clustered 17761 actors into 5 categories using k-nearest neighbors (KNN), and selected the densest 100 directors. In feature engineering, we adopt a similar approach as in the linear case, resulting in a total of $2100$ arms, each of which has $500$ features. Note that there are currently no other tensor-based generalized linear algorithms, hence we compare the cumulative regret of our algorithm with that of other matrix-based generalized linear algorithms.

\begin{itemize}
    \item [(3)] Poisson case: Topic modeling on multiway publications
\end{itemize}

We further explore topic modeling in multiway publications using a scientific database comprising abstracts sourced from papers authored by a diverse group of researchers at Duke University\footnote{Publications dataset, obtained from https://scholars.duke.edu/}. This dataset encapsulates the [\textit{author}, \textit{word}, \textit{venue}] triplet, with the reward $y$ denoting the count number of publications (i.e., a count-based variable). Specifically, we extracted a dense sub-dataset of size 10$\times$10$\times$3 from the original 2425$\times$9088$\times$4068 dataset, from which we derived 100 arms. Employing a methodology similar to the two previously mentioned applications, we constructed the feature tensor $\mathcal{X}$ to represent the interaction feature of the [\textit{author}, \textit{word}, \textit{venue}] triplet. Our comparative analysis involves evaluating our algorithm against other matrix-based generalized linear algorithms.

\begin{figure}[h]
	\centering
	\subfigure[Precision medicine]
	{
		\begin{minipage}[b]{0.307\linewidth}
			\includegraphics[width=1.1\linewidth]{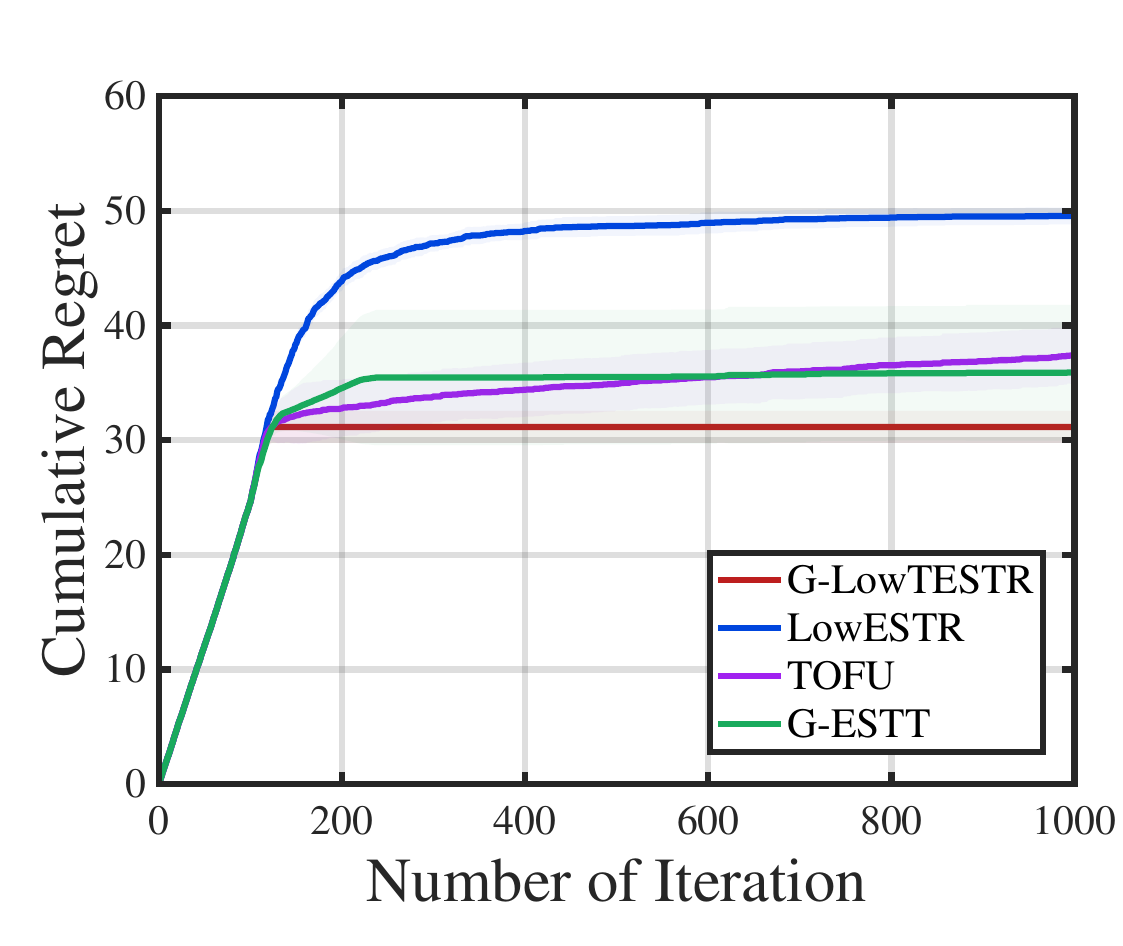}
		\end{minipage}
	}
        \hfill
	\subfigure[Movie recommendation]
{
	\begin{minipage}[b]{0.307\linewidth}
		\includegraphics[width=1.1\linewidth]{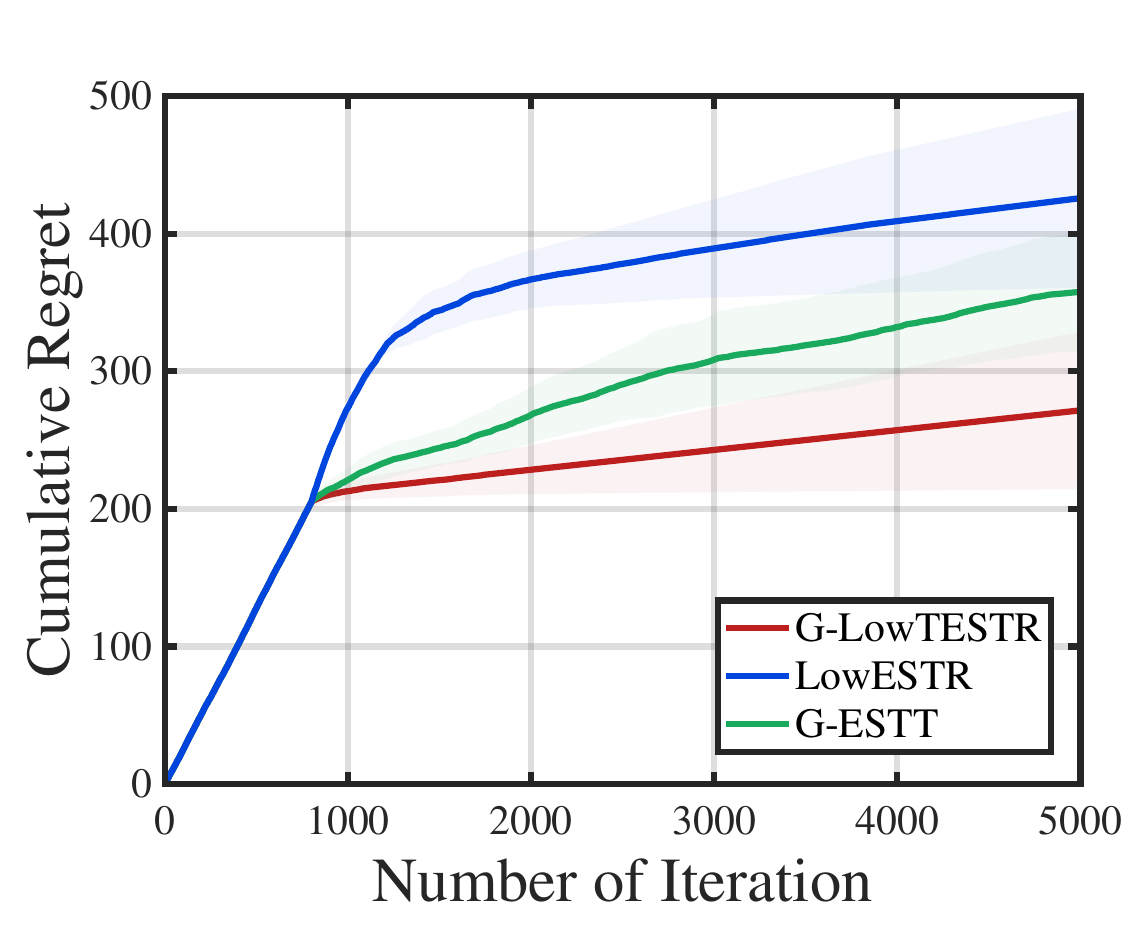}
	\end{minipage}
}
         \hfill
\subfigure[Topic modeling]
{
	\begin{minipage}[b]{0.307\linewidth}
		\includegraphics[width=1.08\linewidth]{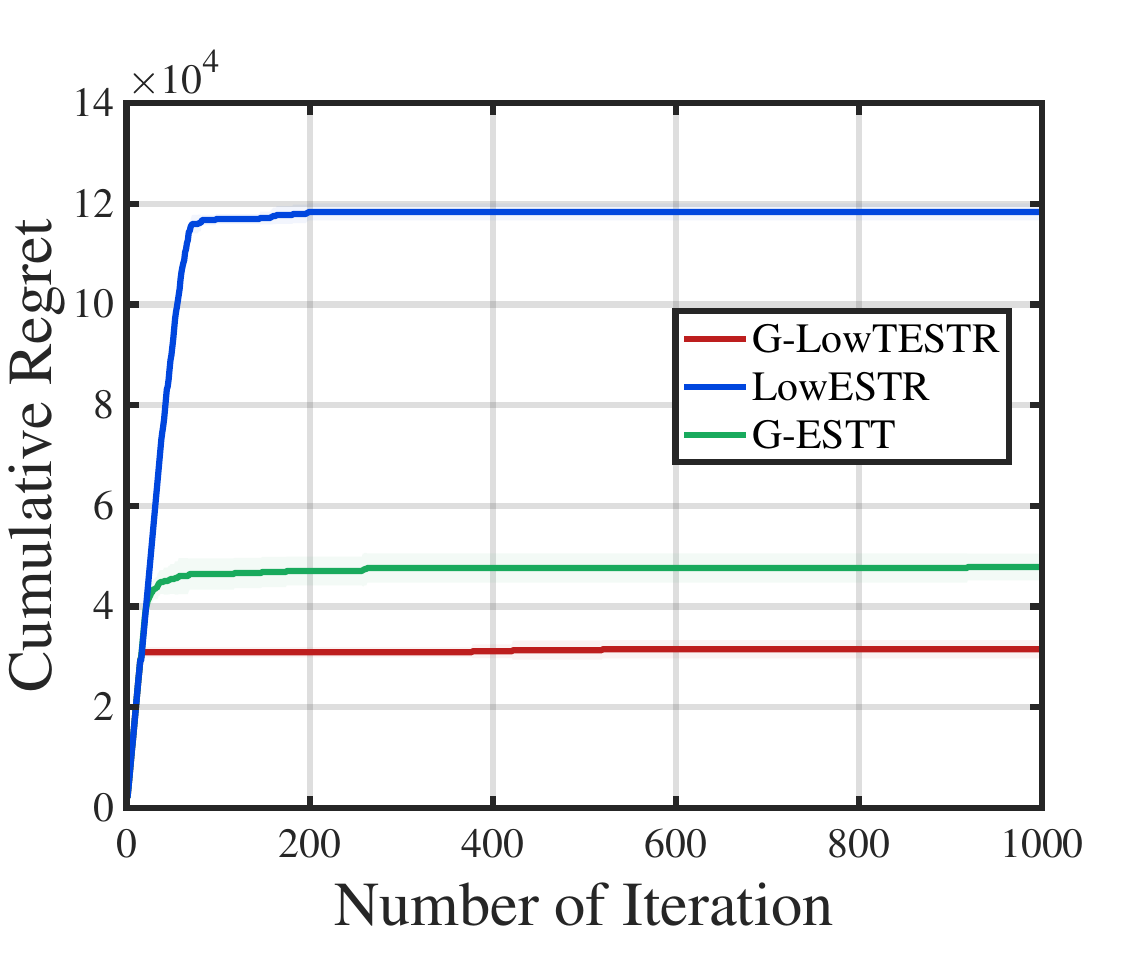}
	\end{minipage}
}
\caption{Comparison analysis with other algorithms in real data} \label{fig:real}
\end{figure}
The regret of the algorithm measures the difference in rewards between the current decision and the optimal decision. That is, the smaller the regret, the greater the reward for the current decision. It can be seen from Fig. \ref{fig:real} that after exploration, the proposed G-LowTESTR can get a smaller slope in a short time. That is, G-LowTESTR can more effectively estimate the parameter $\mathcal{W}^*$, and achieve a smaller regret. This shows that compared with other algorithms, our algorithm performs better by utilizing the low-rank structure of multi-dimensional data.

\section{Conclusion and future work}
This paper introduces the generalized low-rank tensor contextual bandits model, which not only leverages distinct feature information from various data sources but also accommodates the inherent non-linear nature of rewards. Additionally, an efficient G-LowTESTR algorithm is designed based on a tensor algebra framework, making it suitable for handling large-scale datasets in highly complex decision-making scenarios. Furthermore, theoretical evidence showcases that the incorporation of low-rank tensor structures yields improved regret bounds. Comprehensive experiments utilizing synthetic data and real-world applications validate the superiority of the proposed method over vectorization and matricization approaches, further reinforcing the significance of our theoretical contributions. We observe that many significant large-scale applications, including recommender systems, precision medicine, and anomaly detection, are undergoing a shift towards decentralization, where raw data is spread across multiple clients. In response to this trend, federated learning techniques have emerged as a means to handle local datasets privately and collaboratively. In the future, we would like to introduce a novel federated low-rank bandits algorithm designed to simultaneously explore low-rank subspaces across multiple clients. 



\bibliographystyle{apalike} 
\bibliography{ref}

\begin{thebibliography}{}

\bibitem[Abbasi-Yadkori et~al., 2011]{abbasi2011improved}
Abbasi-Yadkori, Y., P{\'a}l, D., and Szepesv{\'a}ri, C. (2011).
\newblock Improved algorithms for linear stochastic bandits.
\newblock {\em Advances in Neural Information Processing Systems}, 24.

\bibitem[Agrawal et~al., 2023]{agrawal2023tractable}
Agrawal, P., Tulabandhula, T., and Avadhanula, V. (2023).
\newblock A tractable online learning algorithm for the multinomial logit
  contextual bandit.
\newblock {\em European Journal of Operational Research}, 310(2):737--750.

\bibitem[Aramayo et~al., 2023]{aramayo2023multiarmed}
Aramayo, N., Schiappacasse, M., and Goic, M. (2023).
\newblock A multiarmed bandit approach for house ads recommendations.
\newblock {\em Marketing Science}, 42(2):271--292.

\bibitem[Audibert et~al., 2009]{audibert2009exploration}
Audibert, J.-Y., Munos, R., and Szepesv{\'a}ri, C. (2009).
\newblock Exploration--exploitation tradeoff using variance estimates in
  multi-armed bandits.
\newblock {\em Theoretical Computer Science}, 410(19):1876--1902.

\bibitem[Auer et~al., 2002]{auer2002finite}
Auer, P., Cesa-Bianchi, N., and Fischer, P. (2002).
\newblock Finite-time analysis of the multiarmed bandit problem.
\newblock {\em Machine Learning}, 47(2):235--256.

\bibitem[Bastani and Bayati, 2020]{bastani2020online}
Bastani, H. and Bayati, M. (2020).
\newblock Online decision making with high-dimensional covariates.
\newblock {\em Operations Research}, 68(1):276--294.

\bibitem[Bastani et~al., 2022]{bastani2022learning}
Bastani, H., Harsha, P., Perakis, G., and Singhvi, D. (2022).
\newblock Learning personalized product recommendations with customer
  disengagement.
\newblock {\em Manufacturing \& Service Operations Management},
  24(4):2010--2028.

\bibitem[Bubeck et~al., 2012]{bubeck2012regret}
Bubeck, S., Cesa-Bianchi, N., et~al. (2012).
\newblock Regret analysis of stochastic and nonstochastic multi-armed bandit
  problems.
\newblock {\em Foundations and Trends{\textregistered} in Machine Learning},
  5(1):1--122.

\bibitem[Chu et~al., 2011]{chu2011contextual}
Chu, W., Li, L., Reyzin, L., and Schapire, R. (2011).
\newblock Contextual bandits with linear payoff functions.
\newblock In {\em Proceedings of the Fourteenth International Conference on
  Artificial Intelligence and Statistics}, pages 208--214. JMLR Workshop and
  Conference Proceedings.

\bibitem[Cobanoglu et~al., 2013]{cobanoglu2013predicting}
Cobanoglu, M.~C., Liu, C., Hu, F., Oltvai, Z.~N., and Bahar, I. (2013).
\newblock Predicting drug--target interactions using probabilistic matrix
  factorization.
\newblock {\em Journal of Chemical Information and Modeling},
  53(12):3399--3409.

\bibitem[Dhillon and Aral, 2021]{dhillon2021modeling}
Dhillon, P.~S. and Aral, S. (2021).
\newblock Modeling dynamic user interests: A neural matrix factorization
  approach.
\newblock {\em Marketing Science}, 40(6):1059--1080.

\bibitem[Fan et~al., 2018]{fan2018lamm}
Fan, J., Liu, H., Sun, Q., and Zhang, T. (2018).
\newblock I-lamm for sparse learning: Simultaneous control of algorithmic
  complexity and statistical error.
\newblock {\em Annals of Statistics}, 46(2):814.

\bibitem[Filippi et~al., 2010]{filippi2010parametric}
Filippi, S., Cappe, O., Garivier, A., and Szepesv{\'a}ri, C. (2010).
\newblock Parametric bandits: The generalized linear case.
\newblock {\em Advances in Neural Information Processing Systems}, 23.

\bibitem[Gopalan et~al., 2016]{gopalan2016low}
Gopalan, A., Maillard, O.-A., and Zaki, M. (2016).
\newblock Low-rank bandits with latent mixtures.
\newblock {\em arXiv preprint arXiv:1609.01508}.

\bibitem[Grant et~al., 2020]{grant2020adaptive}
Grant, J.~A., Leslie, D.~S., Glazebrook, K., Szechtman, R., and Letchford,
  A.~N. (2020).
\newblock Adaptive policies for perimeter surveillance problems.
\newblock {\em European Journal of Operational Research}, 283(1):265--278.

\bibitem[Grant and Szechtman, 2021]{grant2021filtered}
Grant, J.~A. and Szechtman, R. (2021).
\newblock Filtered poisson process bandit on a continuum.
\newblock {\em European Journal of Operational Research}, 295(2):575--586.

\bibitem[Gur and Momeni, 2022]{gur2022adaptive}
Gur, Y. and Momeni, A. (2022).
\newblock Adaptive sequential experiments with unknown information arrival
  processes.
\newblock {\em Manufacturing \& Service Operations Management},
  24(5):2666--2684.

\bibitem[Hillard et~al., 2010]{hillard2010improving}
Hillard, D., Schroedl, S., Manavoglu, E., Raghavan, H., and Leggetter, C.
  (2010).
\newblock Improving ad relevance in sponsored search.
\newblock In {\em Proceedings of the Third ACM International Conference on Web
  Search and Data Mining}, pages 361--370.

\bibitem[Id{\'e} et~al., 2022]{ide2022targeted}
Id{\'e}, T., Murugesan, K., Bouneffouf, D., and Abe, N. (2022).
\newblock Targeted advertising on social networks using online variational
  tensor regression.
\newblock In {\em arXiv preprint arXiv:2208.10627}.

\bibitem[Jang et~al., 2021]{jang2021improved}
Jang, K., Jun, K.-S., Yun, S.-Y., and Kang, W. (2021).
\newblock Improved regret bounds of bilinear bandits using action space
  analysis.
\newblock In {\em International Conference on Machine Learning}, pages
  4744--4754. PMLR.

\bibitem[Jun et~al., 2019]{jun2019bilinear}
Jun, K.-S., Willett, R., Wright, S., and Nowak, R. (2019).
\newblock Bilinear bandits with low-rank structure.
\newblock In {\em International Conference on Machine Learning}, pages
  3163--3172. PMLR.

\bibitem[Kang et~al., 2022]{kang2022efficient}
Kang, Y., Hsieh, C.-J., and Lee, T. C.~M. (2022).
\newblock Efficient frameworks for generalized low-rank matrix bandit problems.
\newblock {\em Advances in Neural Information Processing Systems},
  35:19971--19983.

\bibitem[Katariya et~al., 2017a]{katariya2017bernoulli}
Katariya, S., Kveton, B., Szepesv{\'a}ri, C., Vernade, C., and Wen, Z. (2017a).
\newblock Bernoulli rank-$1 $ bandits for click feedback.
\newblock In {\em arXiv preprint arXiv:1703.06513}.

\bibitem[Katariya et~al., 2017b]{katariya2017stochastic}
Katariya, S., Kveton, B., Szepesvari, C., Vernade, C., and Wen, Z. (2017b).
\newblock Stochastic rank-1 bandits.
\newblock In {\em Artificial Intelligence and Statistics}, pages 392--401.
  PMLR.

\bibitem[Kernfeld et~al., 2015]{kernfeld2015tensor}
Kernfeld, E., Kilmer, M., and Aeron, S. (2015).
\newblock Tensor--tensor products with invertible linear transforms.
\newblock {\em Linear Algebra and its Applications}, 100(485):545--570.

\bibitem[Kilmer and Martin, 2011]{kilmer2011factorization}
Kilmer, M.~E. and Martin, C.~D. (2011).
\newblock Factorization strategies for third-order tensors.
\newblock {\em Linear Algebra and its Applications}, 435(3):641--658.

\bibitem[Kirschner and Krause, 2019]{kirschner2019stochastic}
Kirschner, J. and Krause, A. (2019).
\newblock Stochastic bandits with context distributions.
\newblock {\em Advances in Neural Information Processing Systems}, 32.

\bibitem[Kveton et~al., 2017]{kveton2017stochastic}
Kveton, B., Szepesv{\'a}ri, C., Rao, A., Wen, Z., Abbasi-Yadkori, Y., and
  Muthukrishnan, S. (2017).
\newblock Stochastic low-rank bandits.
\newblock In {\em arXiv preprint arXiv:1712.04644}.

\bibitem[Lai et~al., 1985]{lai1985asymptotically}
Lai, T.~L., Robbins, H., et~al. (1985).
\newblock Asymptotically efficient adaptive allocation rules.
\newblock {\em Advances in Applied Mathematics}, 6(1):4--22.

\bibitem[Li et~al., 2012]{li2012unbiased}
Li, L., Chu, W., Langford, J., Moon, T., and Wang, X. (2012).
\newblock An unbiased offline evaluation of contextual bandit algorithms with
  generalized linear models.
\newblock In {\em Proceedings of the Workshop on On-line Trading of Exploration
  and Exploitation 2}, pages 19--36. JMLR Workshop and Conference Proceedings.

\bibitem[Li et~al., 2010]{li2010contextual}
Li, L., Chu, W., Langford, J., and Schapire, R.~E. (2010).
\newblock A contextual-bandit approach to personalized news article
  recommendation.
\newblock In {\em Proceedings of the 19th International Conference on World
  Wide Web}, pages 661--670.

\bibitem[Li et~al., 2017]{li2017provably}
Li, L., Lu, Y., and Zhou, D. (2017).
\newblock Provably optimal algorithms for generalized linear contextual
  bandits.
\newblock In {\em International Conference on Machine Learning}, pages
  2071--2080. PMLR.

\bibitem[Li et~al., 2022]{li2022simple}
Li, W., Barik, A., and Honorio, J. (2022).
\newblock A simple unified framework for high dimensional bandit problems.
\newblock In {\em International Conference on Machine Learning}, pages
  12619--12655. PMLR.

\bibitem[Lu et~al., 2019]{lu2019low}
Lu, C., Peng, X., and Wei, Y. (2019).
\newblock Low-rank tensor completion with a new tensor nuclear norm induced by
  invertible linear transforms.
\newblock In {\em Proceedings of the IEEE/CVF Conference on Computer Vision and
  Pattern Recognition}, pages 5996--6004.

\bibitem[Lu et~al., 2021a]{lu2021low}
Lu, Y., Meisami, A., and Tewari, A. (2021a).
\newblock Low-rank generalized linear bandit problems.
\newblock In {\em International Conference on Artificial Intelligence and
  Statistics}, pages 460--468. PMLR.

\bibitem[Lu et~al., 2021b]{lu2021bandit}
Lu, Y., Xu, Z., and Tewari, A. (2021b).
\newblock Bandit algorithms for precision medicine.
\newblock In {\em arXiv preprint arXiv:2108.04782}.

\bibitem[Luo et~al., 2018]{luo2018computational}
Luo, H., Li, M., Wang, S., Liu, Q., Li, Y., and Wang, J. (2018).
\newblock Computational drug repositioning using low-rank matrix approximation
  and randomized algorithms.
\newblock {\em Bioinformatics}, 34(11):1904--1912.

\bibitem[Murali et~al., 2021]{murali2021cancer}
Murali, V.~S., {\c{C}}obano{\u{g}}lu, D.~A., Hsieh, M., Zinn, M., Malladi,
  V.~S., Gesell, J., Williams, N.~S., Welf, E.~S., Raj, G.~V., and
  {\c{C}}obano{\u{g}}lu, M.~C. (2021).
\newblock Cancer drug discovery as a low rank tensor completion problem.
\newblock In {\em bioRxiv}. Cold Spring Harbor Laboratory.

\bibitem[Negahban et~al., 2012]{negahban2012unified}
Negahban, S.~N., Ravikumar, P., Wainwright, M.~J., and Yu, B. (2012).
\newblock A unified framework for high-dimensional analysis of m-estimators
  with decomposable regularizers.

\bibitem[Raskutti et~al., 2010]{raskutti2010restricted}
Raskutti, G., Wainwright, M.~J., and Yu, B. (2010).
\newblock Restricted eigenvalue properties for correlated gaussian designs.
\newblock {\em The Journal of Machine Learning Research}, 11:2241--2259.

\bibitem[Recht et~al., 2010]{recht2010guaranteed}
Recht, B., Fazel, M., and Parrilo, P.~A. (2010).
\newblock Guaranteed minimum-rank solutions of linear matrix equations via
  nuclear norm minimization.
\newblock {\em SIAM review}, 52(3):471--501.

\bibitem[Rizk et~al., 2021]{rizk2021best}
Rizk, G., Thomas, A., Colin, I., Laraki, R., and Chevaleyre, Y. (2021).
\newblock Best arm identification in graphical bilinear bandits.
\newblock In {\em International Conference on Machine Learning}, pages
  9010--9019. PMLR.

\bibitem[Schwartz et~al., 2017]{schwartz2017customer}
Schwartz, E.~M., Bradlow, E.~T., and Fader, P.~S. (2017).
\newblock Customer acquisition via display advertising using multi-armed bandit
  experiments.
\newblock {\em Marketing Science}, 36(4):500--522.

\bibitem[Shi et~al., 2023]{shi2023high}
Shi, C., Shen, C., and Sidiropoulos, N.~D. (2023).
\newblock On high-dimensional and low-rank tensor bandits.
\newblock {\em IEEE International Symposium on Information Theory}.

\bibitem[Song et~al., 2020]{song2020robust}
Song, G., Ng, M.~K., and Zhang, X. (2020).
\newblock Robust tensor completion using transformed tensor singular value
  decomposition.
\newblock {\em Numerical Linear Algebra with Applications}, 27(3):e2299.

\bibitem[Tang et~al., 2015]{tang2015personalized}
Tang, L., Jiang, Y., Li, L., Zeng, C., and Li, T. (2015).
\newblock Personalized recommendation via parameter-free contextual bandits.
\newblock In {\em Proceedings of the 38th international ACM SIGIR conference on
  research and development in information retrieval}, pages 323--332.

\bibitem[Thompson, 1933]{thompson1933likelihood}
Thompson, W.~R. (1933).
\newblock On the likelihood that one unknown probability exceeds another in
  view of the evidence of two samples.
\newblock {\em Biometrika}, 25(3-4):285--294.

\bibitem[Vershynin, 2010]{vershynin2010introduction}
Vershynin, R. (2010).
\newblock Introduction to the non-asymptotic analysis of random matrices.
\newblock {\em arXiv preprint arXiv:1011.3027}.

\bibitem[Wainwright, 2019]{wainwright2019high}
Wainwright, M.~J. (2019).
\newblock {\em High-dimensional statistics: A non-asymptotic viewpoint},
  volume~48.
\newblock Cambridge University Press.

\bibitem[Zhou et~al., 2020]{zhou2020stochastic}
Zhou, J., Hao, B., Wen, Z., Zhang, J., and Sun, W.~W. (2020).
\newblock Stochastic low-rank tensor bandits for multi-dimensional online
  decision making.
\newblock In {\em arXiv e-prints}.

\bibitem[Zhou et~al., 2019]{zhou2019learning}
Zhou, Z., Xu, R., and Blanchet, J. (2019).
\newblock Learning in generalized linear contextual bandits with stochastic
  delays.
\newblock {\em Advances in Neural Information Processing Systems}, 32.

\end{thebibliography}
\appendix
\section{Notations and preliminaries}
We briefly review basic notations and definitions that will be used throughout the paper. The symbols $a$, $\mathbf{a}$, $A$, and $\mathbf{\mathcal{A}}$ denote scalars, vectors, matrices,
and tensors, respectively. For the third-order tensor $\mathbf{\mathcal{A}} \in \mathbb{R}^{n_{1} \times n_{2} \times n_{3}}$, the $(i_1,i_2,i_3)$ th entry is denoted by $\mathbf{\mathcal{A}}_{i_1i_2i_3}$ and the $k$th frontal slice is denoted by $\mathcal{A}^{(k)}:=\mathcal{A}_{::k}$.
The Frobenius norm is defined by $\|\mathbf{\mathcal{A}}\|_{F}=\sqrt{\sum_{i_1,i_2,i_3}\left|\mathbf{\mathcal{A}}_{i_1i_2 i_3}\right|^{2}}$. The inner product between $A$ and $B$ in $\mathbb{R}^{n_{1} \times n_{2}}$ is defined as $\langle A,  B \rangle=\mathtt{trace}\left(A^{\top} B\right)$, where $A^{\top}$ denotes the transpose of $A$ and $\mathtt{trace}(\cdot)$ denotes the matrix trace. The inner product between $\mathcal{A}$ and $\mathcal{B}$ in $\mathbb{R}^{n_{1} \times n_{2}  \times n_{3}}$ is defined as $\langle\mathcal{A}, \mathcal{B}\rangle=\sum_{i=1}^{n_3}\left\langle \mathcal{A}^{(i)},  \mathcal{B}^{(i)}\right\rangle$.

Let $L$ be the  invertible linear
transform matrix with the property of
\begin{align}
	LL^{\top}=L^{\top} L=\ell I,\label{Lp}
\end{align}
where $I$ is the identity matrix, we define $\breve{\mathcal{A}}=\mathcal{A}\times_3 L$ and the inverse  $\mathcal{A}=\breve{\mathcal{A}}\times_3 L^{-1}$. $\bar{\mathcal{A}}$ is a block diagonal matrix composed of each frontal slice of $\breve{\mathcal{A}}$.


\begin{definition}[Transformed t-product; \cite{kernfeld2015tensor}] \label{defphiproduct}
\begin{spacing}{1.5}
	Let tensors $\mathcal{A} \in  \mathbb{R}^{n_{1} \times n_{2} \times n_{3}}$ and $\mathcal{B} \in \mathbb{R}^{n_{2} \times a \times n_{3}} .$ Then, the transformed t-product $\mathcal{A} *_L \mathcal{B}$ is defined as the tensor $\mathcal{C} \in \mathbb{R}^{n_{1} \times a \times  n_{3}}$, with
	$
	\mathtt{bdiag}\left(\breve{\mathcal{C}}\right)=\mathtt{bdiag}\left(\breve{\mathcal{A}}\right) \times \mathtt{bdiag}\left(\breve{\mathcal{B}}\right),
	$
	where $\times$ denotes the standard matrix product.
 \end{spacing}
\end{definition}
\vspace{-8mm}

\begin{definition}[Conjugate transpose; \cite{kernfeld2015tensor}]
\begin{spacing}{1.5}
	The conjugate transpose of $\mathcal{A} \in \mathbb{R}^{n_{1} \times n_{2}  \times n_{3}}$ with respect to $L$ is the tensor $\mathcal{A}^{\top} \in \mathbb{C}^{n_{2} \times n_{1} \times  n_{3}}$, which is obtained by
	$
	\mathtt{bdiag}\left(\breve{\mathcal{A}^{\top}}\right)=\mathtt{bdiag}\left(\breve{\mathcal{A}}\right)^{\top}.
	$
 \end{spacing}
\end{definition}
\vspace{-8mm}

\begin{definition}[Identity tensor; \cite{kernfeld2015tensor}]	
\begin{spacing}{1.5}
	If tensor $\breve{\mathcal{I}}\in \mathbb{R}^{m \times m\times n_3 }$ satisfies $\breve{\mathcal{I}}^{(j)}=I_{m\times m}$ for $j=1,\ldots,n_3$, then $\mathcal{I}=\breve{\mathcal{I}}\times_3 L^{-1}$ is the identity tensor.
 \end{spacing}
\end{definition}
\vspace{-8mm}

\begin{definition}[Unitary tensor; \cite{kernfeld2015tensor}]
\begin{spacing}{1.5}
	Tensor $\mathbf{\mathcal{Q}} \in \mathbb{C}^{n \times n \times n_{3} }$ is unitary with respect to the transformed t-product if it satisfies
	$
	\mathcal{Q}^{\top} *_L \mathcal{Q}=\mathcal{Q} *_L \mathcal{Q}^{\top}=\mathcal{I},
	$
	where $\mathcal{I}$ is the identity tensor.
 \end{spacing}
\end{definition}
\vspace{-8mm}

\begin{definition}[F-diagonal tensor; \cite{kilmer2011factorization}]	
\begin{spacing}{1.5}
	Tensor $\mathcal{A}$ is f-diagonal if each of its frontal slices is a diagonal matrix.
 \end{spacing}
\end{definition}
\vspace{-8mm}

\begin{definition}[Transformed t-SVD; \cite{kernfeld2015tensor}]\label{ttSVD}
\begin{spacing}{1.5}
	Let $\mathbf{\mathcal{A}}$ be an $n_{1} \times n_{2} \times n_{3} $ tensor. It can be factored as
	$
	\mathcal{A}=\mathcal{U} *_L \mathcal{S} *_L \mathcal{V}^{\top},
	$
	where $\mathcal{U}\in \mathbb{C}^{n_{1} \times n_{1} \times n_{3}}$ and $\mathcal{V}\in \mathbb{C}^{n_{2} \times n_{2} \times n_{3} }$ are unitary tensors with respect to the transformed $t$-product, and $\mathcal{S}\in \mathbb{C}^{n_{1} \times n_{2} \times n_{3} }$ is an f-diagonal tensor.
 \end{spacing}
\end{definition}
\vspace{-8mm}

\begin{definition}[Tensor tubal rank; \cite{song2020robust}]\label{tubalrank}
\begin{spacing}{1.5}
	For $\mathcal{A} \in$ $\mathbb{R}^{n_{1} \times n_2 \times n_{3}},$ the tensor tubal rank, denoted by $\operatorname{rank}_{t}(\mathcal{A})=\#\{i, \mathcal{S}(i, i,:) \neq 0\}$, is the number of nonzero tube fibers of $\mathcal{S},$ where $\mathcal{S}$ is from the transformed t-SVD of $\mathcal{A}=\mathcal{U} *_L \mathcal{S} *_L \mathcal{V}^{\top} $.
 \end{spacing}
\end{definition}
\vspace{-8mm}

\begin{definition}[Tensor spectral norm; \cite{lu2019low}]
\begin{spacing}{1.5}
	The tensor spectral norm of $\mathcal{A}\in \mathbb{R}^{n_{1} \times n_{2} \times  n_{3}}$ under $L$ is defined as
	$
	\left\|\mathcal{A}\right\|_2=\left\|\bar{\mathcal{A}}\right\|_2
	$.
 \end{spacing}
\end{definition}
\vspace{-8mm}

\begin{lemma}[Tensor nuclear norm; \cite{lu2019low}]\label{tnnnnn}
\begin{spacing}{1.5}
	For any $\mathcal{B} \in \mathbb{R}^{n_{1} \times n_{2} \times  n_{3}}$, the tensor nuclear norm of $\mathcal{A}\in \mathbb{R}^{n_{1} \times n_{2}  \times n_{3}}$ is $\|\mathcal{A}\|_{*} :=\sup \limits_{\|\mathcal{B}\|_2 \leq 1}\langle\mathcal{A}, \mathcal{B}\rangle=\frac{1}{\ell}\|\bar{\mathcal{A}}\|_{*}$.
 \end{spacing}
\end{lemma}
\vspace{-8mm}

\section{Proofs}
\subsection{Proof of of Theorem 1}
In this section, we shall prove Theorem 1. Due to our proof is based on the local restricted strongly convexity (LRSC) condition, we first determine the decomposability of the tensor nuclear norm and the form of the restricted set. Before that, we now introduce some related definitions. For any tensor $\mathcal{A}\in \mathbb{R}^{d_1 \times d_2 \times d_3}$, we denote the subspaces spanned by its horizontal slices and lateral slices as $\operatorname{hor}(\mathcal{A})$ and $\operatorname{lat}(\mathcal{A})$, respectively. Specifically,
$$
\begin{aligned}
	\operatorname{hor}(\mathcal{A}) & :=\left\{\mathcal{B} \in \mathbb{R}^{d_2 \times 1 \times d_3} \mid \mathcal{B}=\mathcal{A}^* *_L \mathcal{X}, \mathcal { X } \in \mathbb{R}^{d_1 \times 1 \times d_3}\right\}, \\
	\operatorname{lat}(\mathcal{A}) & :=\left\{\mathcal{B} \in \mathbb{R}^{d_1 \times 1 \times d_3} \mid \mathcal{B}=\mathcal{A} *_L \mathcal{X}, \mathcal{X} \in \mathbb{R}^{d_2 \times 1 \times d_3}\right\}.
\end{aligned}
$$
And for the given positive integer $r\le d^\prime:=\min\{d_1,d_2\}$, we define $\mathcal{U}\in \mathbb{R}^{d_1 \times d^\prime \times d_3}$ and $\mathcal{V}\in \mathbb{R}^{d_2 \times d^\prime \times d_3}$ as orthonormal tensors, and denote their first $r$ lateral slices as $\mathcal{U}^r$ and $\mathcal{V}^r$, respectively. Then we define the following two subspaces
\begin{small}
\begin{align}
	&\mathcal{K}(\operatorname{lat}\left(\mathcal{U}^r\right),\operatorname{lat}\left(\mathcal{V}^r\right)):=\left\{\mathcal{A} \in \mathbb{R}^{d_1 \times d_2 \times d_3} \mid \operatorname{hor}(\mathcal{A}) \subseteq \operatorname{lat}\left(\mathcal{V}^r\right), \operatorname{lat}(\mathcal{A}) \subseteq \operatorname{lat}\left(\mathcal{U}^r\right)\right\}, \notag\\
	&\widetilde{\mathcal{K}}^{\perp}(\operatorname{lat}\left(\mathcal{U}^r\right),\operatorname{lat}\left(\mathcal{V}^r\right)):=\left\{\mathcal{A} \in \mathbb{R}^{d_1 \times d_2 \times d_3} \mid \operatorname{hor}(\mathcal { A }) \subseteq \operatorname{lat}\left(\mathcal{V}^r\right)^\perp, \operatorname{lat}(\mathcal{A}) \subseteq \operatorname{lat}\left(\mathcal{U}^r\right)^\perp\right\} .\label{two subspaces}
\end{align}
\end{small}

\begin{lemma}[The decomposability of tensor nuclear norm]
\begin{spacing}{1.5}
	The tensor nuclear norm is decomposable, i.e., for any tensor $\mathcal{ A }\in	\mathcal{K}(\operatorname{lat}\left(\mathcal{U}^r\right),\operatorname{lat}\left(\mathcal{V}^r\right))$ and tensor $\mathcal{B}\in	\widetilde{\mathcal{K}}^{\perp}(\operatorname{lat}\left(\mathcal{U}^r\right),\operatorname{lat}\left(\mathcal{V}^r\right))$, we have
		$$
	\left\|\mathcal{A}+\mathcal{B}\right\|_{*}=\left\|\mathcal{A}\right\|_{*}+\left\|\mathcal{B}\right\|_{*} .
	$$
 \end{spacing}
\end{lemma}
\vspace{-6mm}

\begin{proof} 
Due to $\mathcal{ A }\in	\mathcal{K}(\operatorname{lat}\left(\mathcal{U}^r\right),\operatorname{lat}\left(\mathcal{V}^r\right))$ and $\mathcal{B}\in	\widetilde{\mathcal{K}}^{\perp}(\operatorname{lat}\left(\mathcal{U}^r\right),\operatorname{lat}\left(\mathcal{V}^r\right))$, denote $\mathcal{T}_{11}\in\mathbb{R}^{r\times r \times d_3}$ and $\mathcal{T}_{22}\in\mathbb{R}^{(d^\prime-r) \times  (d^\prime-r)  \times d_3}$ as arbitrary tensors, then we can quivalently express $\mathcal{ A }$ and $\mathcal{ B }$ as
$$
\mathcal{A}=\mathcal{U}*_L\left[\begin{array}{ll}
	\mathcal{T}_{11} & 0_{r\times (d^\prime-r) \times d_3} \\
	0_{ (d^\prime-r) \times r \times d_3} & 0_{ (d^\prime-r) \times  (d^\prime-r)  \times d_3}
\end{array}\right] *_L\mathcal{V}^{\top}, $$ and$$ \mathcal{B}=\mathcal{U}*_L\left[\begin{array}{ll}
0_{r\times r \times d_3} & 0_{r\times (d^\prime-r) \times d_3} \\
0_{ (d^\prime-r) \times r \times d_3} & 	\mathcal{T}_{22}
\end{array}\right] *_L\mathcal{V}^{\top}.
$$
Due to the unitary invariance of the matrix nuclear norm, i.e., for any matrix $A$ and two orthogonal matrices $U$ and $V$, $\|UAV^\top\|_*=\|A\|_*$, we can obtain the  unitary invariance of the tensor nuclear norm: $\|\mathcal{U}*_L\mathcal{A}*_L\mathcal{V}^\top\|_*
=\frac{1}{\ell}\|\bar{\mathcal{U}}\bar{\mathcal{A}}\bar{\mathcal{V}}^\top\|_*
=\frac{1}{\ell}\|\bar{\mathcal{A}}\|_*
=\|\mathcal{A}\|_*$, where $\mathcal{A}$ is an arbitrary tensor and $\mathcal{U}$, $\mathcal{V}$ are the orthogonal tensors. Then according to the unitary invariance of the tensor nuclear norm, we have

\begin{align}
&\left\|\mathcal{A}+\mathcal{B}\right\|_{*}\notag\\
	=&	\left\|\mathcal{U}*_L\left[\begin{array}{ll}
		\mathcal{T}_{11} & 0_{r\times (d^\prime-r) \times d_3} \\
		0_{ (d^\prime-r) \times r \times d_3} & 0_{ (d^\prime-r) \times  (d^\prime-r)  \times d_3}
	\end{array}\right] *_L\mathcal{V}^{\top}\right.\notag\\
 &\left.+\mathcal{U}*_L\left[\begin{array}{ll}
	0_{r\times r \times d_3} & 0_{r\times (d^\prime-r) \times d_3} \\
	0_{ (d^\prime-r) \times r \times d_3} & 	\mathcal{T}_{22}
\end{array}\right] *_L\mathcal{V}^{\top}\right\|_{*} \notag\\
	=&	\left\|\left[\begin{array}{ll}
	\mathcal{T}_{11} & 0_{r\times (d^\prime-r) \times d_3} \\
	0_{ (d^\prime-r) \times r \times d_3} & 0_{ (d^\prime-r) \times  (d^\prime-r)  \times d_3}
\end{array}\right] +\left[\begin{array}{ll}
	0_{r\times r \times d_3} & 0_{r\times (d^\prime-r) \times d_3} \\
	0_{ (d^\prime-r) \times r \times d_3} & 	\mathcal{T}_{22}
\end{array}\right]\right\|_{*} .\label{additivity tensor}
\end{align}

The additivity of matrix nuclear norm \citep{recht2010guaranteed} implies that the additivity of tensor nuclear norm, that is to say, given $\mathcal{ A },\mathcal{ B }\in\mathbb{R}^{d_1\times d_2 \times d_3}$, if $\mathcal{ A }*_L\mathcal{ B }^\top=0$ (i.e., $\bar{\mathcal{ A }}\bar{\mathcal{ B }}^\top=0$) and $\mathcal{ A }^\top*_L\mathcal{ B }=0$ (i.e., $\bar{\mathcal{ A }}^\top\bar{\mathcal{ B }}=0$), then we have $\|\mathcal{ A }+\mathcal{ B }\|_*=\frac{1}{\ell}\|\bar{\mathcal{ A }}+\bar{\mathcal{ B }}\|_*=\frac{1}{\ell}(\|\bar{\mathcal{ A }}\|_*+\|\bar{\mathcal{ B }}\|_*)=\|\mathcal{ A }\|_*+\|\mathcal{ B }\|_*$, where the second equality holds because of the additivity of matrix nuclear norm. Note that $$\left[\begin{array}{ll}
	\mathcal{T}_{11} & 0_{r\times (d^\prime-r) \times d_3} \\
	0_{ (d^\prime-r) \times r \times d_3} & 0_{ (d^\prime-r) \times  (d^\prime-r)  \times d_3}
\end{array}\right]*_L \left[\begin{array}{ll}
0_{r\times r \times d_3} & 0_{r\times (d^\prime-r) \times d_3} \\
0_{ (d^\prime-r) \times r \times d_3} & 	\mathcal{T}_{22}
\end{array}\right]^{\top}=0$$ 
and $$\left[\begin{array}{ll}
\mathcal{T}_{11} & 0_{r\times (d^\prime-r) \times d_3} \\
0_{ (d^\prime-r) \times r \times d_3} & 0_{ (d^\prime-r) \times  (d^\prime-r)  \times d_3}
\end{array}\right]^\top*_L \left[\begin{array}{ll}
0_{r\times r \times d_3} & 0_{r\times (d^\prime-r) \times d_3} \\
0_{ (d^\prime-r) \times r \times d_3} & 	\mathcal{T}_{22}
\end{array}\right]=0,$$ 
then the additivity and unitary invariance of the tensor nuclear norm can further express (\ref{additivity tensor}) as

\begin{align}
	&\left\|\mathcal{A}+\mathcal{B}\right\|_{*}\notag\\
	=&	\left\|\left[\begin{array}{ll}
		\mathcal{T}_{11} & 0_{r\times (d^\prime-r) \times d_3} \\
		0_{ (d^\prime-r) \times r \times d_3} & 0_{ (d^\prime-r) \times  (d^\prime-r)  \times d_3}
	\end{array}\right] +\left[\begin{array}{ll}
		0_{r\times r \times d_3} & 0_{r\times (d^\prime-r) \times d_3} \\
		0_{ (d^\prime-r) \times r \times d_3} & 	\mathcal{T}_{22}
	\end{array}\right]\right\|_{*} \notag\\
=&	\left\|\left[\begin{array}{ll}
	\mathcal{T}_{11} & 0_{r\times (d^\prime-r) \times d_3} \\
	0_{ (d^\prime-r) \times r \times d_3} & 0_{ (d^\prime-r) \times  (d^\prime-r)  \times d_3}
\end{array}\right] \right\|_*+\left\|\left[\begin{array}{ll}
	0_{r\times r \times d_3} & 0_{r\times (d^\prime-r) \times d_3} \\
	0_{ (d^\prime-r) \times r \times d_3} & 	\mathcal{T}_{22}
\end{array}\right]\right\|_{*}\notag\\
=&	\left\|\mathcal{U}*_L\left[\begin{array}{ll}
	\mathcal{T}_{11} & 0_{r\times (d^\prime-r) \times d_3} \\
	0_{ (d^\prime-r) \times r \times d_3} & 0_{ (d^\prime-r) \times  (d^\prime-r)  \times d_3}
\end{array}\right]*_L\mathcal{V}^\top \right\|_*\notag\\
&+\left\|\mathcal{U}*_L\left[\begin{array}{ll}
	0_{r\times r \times d_3} & 0_{r\times (d^\prime-r) \times d_3} \\
	0_{ (d^\prime-r) \times r \times d_3} & 	\mathcal{T}_{22}
\end{array}\right]*_L\mathcal{V}^\top\right\|_{*}\notag\\
=&	\left\|\mathcal{A}\right\|_*+\left\|\mathcal{B}\right\|_{*}.\notag
\end{align}

By this, the decomposability of tensor nuclear norm is proved.
\end{proof}
\begin{lemma}\label{restricted set}
\begin{spacing}{1.5}
	Denote $\widehat{\Delta}=\widehat{\mathcal{W}}-\mathcal{W}^*$, then if $$\left\|\frac{1}{n} \sum_{t=1}^n [b^\prime\left(\langle\mathcal{W}^*,\mathcal{X}_t\rangle\right)-y_t]\mathcal{X}_t\right\|_{2} \leq \frac{\lambda_n}{2},$$ we can obtain that
	\begin{itemize}
		\item[(a)] $\operatorname{rank}_{t}\left(\widehat{\Delta}_{\widetilde{\mathcal{K}}}\right) \leq 2 r$, and
		\item[(b)]$\left\|\widehat{\Delta}_{\widetilde{\mathcal{K}}^\perp}\right\|_{*} \leq 3\left\|\widehat{\Delta}_{\widetilde{\mathcal{K}}}\right\|_{*}+4\left\|\mathcal{W}^*_{\mathcal{K}^\perp}\right\|_{*}=3\left\|\widehat{\Delta}_{\widetilde{\mathcal{K}}}\right\|_{*}$, which is called as the restricted set $\mathcal{C}(\mathcal{K},\widetilde{\mathcal{K}}^\perp,\mathcal{W}^*)$.
	\end{itemize}
 \end{spacing}
\end{lemma}
\vspace{-6mm}
\begin{proof}
For (a), given any tensor $\widetilde{\mathcal{A}}\in	\widetilde{\mathcal{K}}(\operatorname{lat}\left(\mathcal{U}^r\right),\operatorname{lat}\left(\mathcal{V}^r\right))$, it can be expressed as
$$
\widetilde{\mathcal{A}}=\mathcal{U}*_L\left[\begin{array}{ll}
	\widetilde{\mathcal{T}}_{11} & 	\widetilde{\mathcal{T}}_{12}  \\
		\widetilde{\mathcal{T}}_{21} & 0_{ (d^\prime-r) \times  (d^\prime-r)  \times d_3}
\end{array}\right] *_L\mathcal{V}^{\top},
$$
where $	\widetilde{\mathcal{T}}_{11}\in\mathbb{R}^{r\times r\times d_3},	\widetilde{\mathcal{T}}_{12}\in \mathbb{R}^{r\times (d^\prime-r) \times d_3}, 	\widetilde{\mathcal{T}}_{21}\in\mathbb{R}^{ (d^\prime-r) \times r \times d_3}$ are arbitrary tensors. The representation shows that the tensor $\widetilde{\mathcal{A}}\in	\widetilde{\mathcal{K}}(\operatorname{lat}\left(\mathcal{U}^r\right),\operatorname{lat}\left(\mathcal{V}^r\right))$ has the tubal-rank at most $2r$. Thus $\operatorname{rank}_{t}\left(\hat{\Delta}_{\widetilde{\mathcal{K}}}\right) \leq 2 r$.

For (b), due to $$
\widehat{\mathcal{W}}=\arg\min_{\mathcal{W} \in \mathbb{R}^{d_1 \times d_2\times d_3}} \frac{1}{ n} \sum_{t=1}^{n}[b(\left\langle \mathcal{X}_{t}, \mathcal{W}\right\rangle)-y_t\left\langle \mathcal{X}_{t}, \mathcal{W}\right\rangle]+\lambda_{n}\|\mathcal{W}\|_* ,
$$ 
we define $ \frac{1}{ n} \sum_{t=1}^{n}[b(\left\langle \mathcal{X}_{t}, \mathcal{W}\right\rangle)-y_t\left\langle \mathcal{X}_{t}, \mathcal{W}\right\rangle]$ as $L(\mathcal{W})$, then we have
\begin{align}
	&L(\widehat{\mathcal{W}})+\lambda_{n}\|\widehat{\mathcal{W}}\|_*-(L(\mathcal{W}^*)+\lambda_{n}\|\mathcal{W}^*\|_*)\notag\\
	=&L(\widehat{\Delta}+\mathcal{W}^*)+\lambda_{n}\|\widehat{\Delta}+\mathcal{W}^*\|_*-(L(\mathcal{W}^*)+\lambda_{n}\|\mathcal{W}^*\|_*)
	\le0.\label{optimal}
\end{align}
Next, we analyze $L(\widehat{\Delta}+\mathcal{W}^*)-L(\mathcal{W}^*)$ and $\|\widehat{\Delta}+\mathcal{W}^*\|_*-\|\mathcal{W}^*\|_*$ in turn. First, combined with the convexity of $L(\cdot)$ and the property that $\left|\left\langle\mathcal{A},\mathcal{B}\right\rangle\right|=\frac{1}{\ell}\left|\left\langle\bar{\mathcal{A}},\bar{\mathcal{B}}\right\rangle\right|\le\frac{1}{\ell}\left\|\bar{\mathcal{A}}\right\|_2\left\|\bar{\mathcal{B}}\right\|_*=\left\|\mathcal{A}\right\|_2\left\|\mathcal{B}\right\|_*$, we obtain
\begin{align}
	L(\widehat{\Delta}+\mathcal{W}^*)-L(\mathcal{W}^*)\ge-\left|\langle\nabla L(\mathcal{W}^*),\widehat{\Delta}\rangle\right|
	\ge-\left\|\nabla L(\mathcal{W}^*)\right\|_2\left\|\widehat{\Delta}\right\|_*.\label{condition}
\end{align}
Combining $\|\nabla L(\mathcal{W}^*)\|_2=\|\frac{1}{n} \sum_{t=1}^n [b^\prime\left(\langle\mathcal{W}^*,\mathcal{X}_t\rangle\right)-y_t]-\mathcal{X}_t\|_2 \leq \frac{\lambda_n}{2}$ with the triangular inequality, we can further lower bound (\ref{condition}) as
\begin{align}
	&L(\widehat{\Delta}+\mathcal{W}^*)-L(\mathcal{W}^*)
	\ge-\left\|\nabla L(\mathcal{W}^*)\right\|_2\left\|\widehat{\Delta}\right\|_*
	\ge-\frac{\lambda_n}{2}\left\|\widehat{\Delta}\right\|_*\notag\\
	=&-\frac{\lambda_n}{2}\left\|\widehat{\Delta}_{\widetilde{\mathcal{K}}^\perp}+\widehat{\Delta}_{\widetilde{\mathcal{K}}}\right\|_*
	\ge-\frac{\lambda_n}{2}\left(\left\|\widehat{\Delta}_{\widetilde{\mathcal{K}}^\perp}\right\|_*+\left\|\widehat{\Delta}_{\widetilde{\mathcal{K}}}\right\|_*\right).\label{first}
\end{align}
Second, according to the triangular inequality and the decomposability of tensor nuclear norm, we have
\begin{align}
&	\|\widehat{\Delta}+\mathcal{W}^*\|_*-\|\mathcal{W}^*\|_*
	=	\|\widehat{\Delta}_{\widetilde{\mathcal{K}}^\perp}+\widehat{\Delta}_{\widetilde{\mathcal{K}}}+\mathcal{W}^*_{\mathcal{K}^\perp}+\mathcal{W}^*_{\mathcal{K}}\|_*-\|\mathcal{W}^*_{\mathcal{K}^\perp}+\mathcal{W}^*_{\mathcal{K}}\|_*\notag\\
	\ge&\|\mathcal{W}^*_{\mathcal{K}}+\widehat{\Delta}_{\widetilde{\mathcal{K}}^\perp}\|_*-(\|\widehat{\Delta}_{\widetilde{\mathcal{K}}}\|_*+\|\mathcal{W}^*_{\mathcal{K}^\perp}\|_*)-(\|\mathcal{W}^*_{\mathcal{K}^\perp}\|_*+\|\mathcal{W}^*_{\mathcal{K}}\|_*)\notag\\	=&\|\mathcal{W}^*_{\mathcal{K}}\|_*+\|\widehat{\Delta}_{\widetilde{\mathcal{K}}^\perp}\|_*-(\|\widehat{\Delta}_{\widetilde{\mathcal{K}}}\|_*+\|\mathcal{W}^*_{\mathcal{K}^\perp}\|_*)-(\|\mathcal{W}^*_{\mathcal{K}^\perp}\|_*+\|\mathcal{W}^*_{\mathcal{K}}\|_*)\notag\\
	=&\|\widehat{\Delta}_{\widetilde{\mathcal{K}}^\perp}\|_*-\|\widehat{\Delta}_{\widetilde{\mathcal{K}}}\|_*-2\|\mathcal{W}^*_{\mathcal{K}^\perp}\|_*.\label{second}
\end{align}
Substituting (\ref{first}))and (\ref{second}) into (\ref{optimal}) yields that $\left\|\widehat{\Delta}_{\widetilde{\mathcal{K}}^\perp}\right\|_{*} \leq 3\left\|\widehat{\Delta}_{\widetilde{\mathcal{K}}}\right\|_{*}+4\left\|\mathcal{W}^*_{\mathcal{K}^\perp}\right\|_{*}$. And because the tubal rank of $\mathcal{W}^*$ is $r$, we have $\left\|\mathcal{W}^*_{\mathcal{K}^\perp}\right\|_{*}=0$. This completes the proof.
\end{proof}

Lemma \ref{restricted set} states that the error tensor must lie in a special set. We now prove a general error bound under the LRSC condition stated in Definition 1.

\begin{proof}[Proof of Theorem 1.]
To prove the error bound, we first define a middle point $\widehat{\mathcal{W}}_\eta=\mathcal{W}^*+\eta\left(\widehat{\mathcal { W }}-\mathcal{W}^*\right)$, where $\eta=1$ if $\| \widehat{\mathcal{W}}-$ $\mathcal{W}^* \|_F \leq l$, and $\eta=l /\left\|\widehat{\mathcal{W}}-\mathcal{W}^*\right\|_F$ if $\left\|\widehat{\mathcal{W}}-\mathcal{W}^*\right\|_F>l$. And let $\widehat{\Delta}=\widehat{\mathcal{W}}-\mathcal{W}^*, \widehat{\Delta}_\eta=\widehat{\mathcal{W}}_\eta-\mathcal{W}^*$. Note that $\widehat{\Delta} \in \mathcal{C}(\mathcal{K},\widetilde{\mathcal{K}}^\perp,\mathcal{W}^*)$, and $\widehat{\Delta}_\eta$ also belongs to this set because $\widehat{\Delta}_\eta$ is parallel to $\widehat{\Delta}$. The LRSC condition of $L(\mathcal{W})$ indicates that
\begin{align}
L(\mathcal{W}^*+\widehat{\Delta}_\eta)-L(\mathcal{W}^*)-\langle\nabla L(\mathcal{W}^*), \widehat{\Delta}_\eta\rangle \geq \kappa_{\ell}\|\widehat{\Delta}_\eta\|_F^2.\label{use set}
\end{align}
Combining with the property that $L(\mathcal{W}^*+\widehat{\Delta}_\eta)-L(\mathcal{W}^*)\le\langle\nabla L(\mathcal{W}^*+\widehat{\Delta}_\eta), \widehat{\Delta}_\eta\rangle=\langle\nabla L(\widehat{\mathcal{W}}_\eta), \widehat{\Delta}_\eta\rangle$ due to the convexity of $L(\cdot)$, (\ref{use set}) can be further expressed as 
\begin{align}
	\kappa_{\ell}\|\widehat{\Delta}_\eta\|_F^2
	\le\langle\nabla L(\widehat{\mathcal{W}}_\eta)-\nabla L(\mathcal{W}^*), \widehat{\Delta}_\eta\rangle:= D_L^s\left(\widehat{\mathcal{W}}_\eta, \mathcal{W}^*\right), \label{sym}
\end{align}
where $D_L^s\left(\mathcal{W}_1, \mathcal{W}_2\right)=$ $D_L\left(\mathcal{W}_1, \mathcal{W}_2\right)+D_L\left(\mathcal{W}_2, \mathcal{W}_1\right)$ is the symmetric Bregman divergence, and the $D_L\left(\mathcal{W}_1, \mathcal{W}_2\right)=L\left(\mathcal{W}_1\right)-L\left(\mathcal{W}_2\right)-$ $\left\langle\nabla L\left(\mathcal{W}_2\right), \mathcal{W}_1-\mathcal{W}_2\right\rangle$ involved is the Bregman divergence and is  convex in its first argument. Obviously, $D_L\left(\mathcal{W}_1, \mathcal{W}_2\right)\ge 0$ with help of the convexity of the function $L(\cdot)$. Define $Q(\eta)=D_L\left(\widehat{\mathcal{W}}_\eta, \mathcal{W}^*\right)$, note that $Q(\eta)$ is convex and $Q^{\prime}(\eta)=\left\langle\nabla L\left(\widehat{\mathcal{W}}_\eta\right)-\nabla L\left(\mathcal { W }^*\right), \widehat{\Delta}\right\rangle=$ $\frac{1}{\eta} D_L^s\left(\widehat{\mathcal{W}}_\eta, \mathcal{W}^*\right), 0<\eta \leq 1$, which yields that
\begin{align}
D_L^s\left(\widehat{\mathcal{W}}_\eta, \mathcal{W}^*\right)=\eta Q^{\prime}(\eta) \leq \eta Q^{\prime}(1)=\eta D_L^s\left(\widehat{\mathcal{W}}, \mathcal{W}^*\right).\label{convex sym}
\end{align}
Combining (\ref{sym}) with (\ref{convex sym}) concludes that
\begin{align}
	\kappa_{\ell}\|\widehat{\Delta}_\eta\|_F^2& \leq D_L^s\left(\widehat{\mathcal{W}}_\eta, \mathcal{W}^*\right) 
	 \leq \eta D_L^s\left(\widehat{\mathcal { W}}, \mathcal{W}^*\right) 
	=\left\langle\nabla L(\widehat{\mathcal { W }})-\nabla L\left(\mathcal{W}^*\right), \widehat{\Delta}_\eta\right\rangle . \label{sub}
\end{align}
As $\widehat{\mathcal{W}}$ is the solution of optimization problem in (1), we can easily refer that $\nabla L(\widehat{\mathcal{W}})+\lambda_n \mathcal{E}=0$, where $\mathcal{E} \in\partial\|\widehat{\mathcal{W}}\|_*$ represents the subgradient of the $\|\cdot\|_*$ at $\widehat{\mathcal{W}}$. Denote the skinny transformed t-SVD of $\widehat{\mathcal{W}}$ as $\widehat{\mathcal{W}}=\mathcal{U} *_L \mathcal{S} *_L \mathcal{V}^\top$, then from Lemma 10 of \citep{song2020robust}, we have $\partial\|\widehat{\mathcal{W}}\|_*=\left\{\mathcal{U} *_L \mathcal{V}^\top+\mathcal{F} \mid \mathcal{U}^\top *_L \mathcal{F}=\right.$ $\left.0, \mathcal{F} *_L \mathcal{V}=0,\|\mathcal{F}\|_2\le1\right\}$. This combines with $\widehat{\Delta}_\eta \in \mathcal{C}(\mathcal{K},\widetilde{\mathcal{K}}^\perp,\mathcal{W}^*)$ can further upper bound (\ref{sub}) by
\begin{align}
	&\kappa_{\ell}\left\|\widehat{\Delta}_\eta\right\|_F^2  \leq-\left\langle\nabla L\left(\mathcal{W}^*\right)+\lambda_n \mathcal{E}, \widehat{\Delta}_\eta\right\rangle 
	 \leq\left\|\nabla L\left(\mathcal{W}^*\right)+\lambda_n \mathcal{E}\right\|_{2}\left\|\widehat{\Delta}_\eta\right\|_*\notag\\
	 \leq&\left(\left\|\nabla L\left(\mathcal{W}^*\right)\right\|_{2}+\lambda_n\left\| \mathcal{E}\right\|_{2}\right)\left\|\widehat{\Delta}_\eta\right\|_*  
	 \leq\left[\frac{\lambda_n}{2}+\lambda_n\left(\left\|\mathcal{U} *_L \mathcal{V}^\top\right\|_{2}+\|\mathcal{F}\|_{2}\right)\right]\left\|\widehat{\Delta}_\eta\right\|_* \notag\\
	 \leq &2.5 \lambda_n\left\|\widehat{\Delta}_\eta\right\|_*
	 \le2.5\lambda_n\left(\left\|(\widehat{\Delta}_\eta)_{\widetilde{\mathcal{K}}^\perp}\right\|_*+\left\|(\widehat{\Delta}_\eta)_{\widetilde{\mathcal{K}}}\right\|_*\right)
	 \le10\lambda_n \left\|(\widehat{\Delta}_\eta)_{\widetilde{\mathcal{K}}}\right\|_*
	 . \label{sub rsc}
\end{align}
By Lemma \ref{restricted set}, $\operatorname{rank}_{t}\left((\widehat{\Delta}_\eta)_{\widetilde{\mathcal{K}}}\right) \leq 2 r$, the relationship between matrix nuclear and Frobenius norms reveals that
\begin{align}
	\left\|(\widehat{\Delta}_\eta)_{\widetilde{\mathcal{K}}}\right\|_*=\frac{1}{\ell}\left\|(\bar{\widehat{\Delta}}_\eta)_{\widetilde{\mathcal{K}}}\right\|_*
	\le\frac{1}{\ell}\sqrt{2rd_3}\left\|(\bar{\widehat{\Delta}}_\eta)_{\widetilde{\mathcal{K}}}\right\|_F=\sqrt{\frac{2rd_3}{\ell}}\left\|(\widehat{\Delta}_\eta)_{\widetilde{\mathcal{K}}}\right\|_F. \label{relation}
\end{align}
According to the non-expansiveness of the projection and (\ref{relation}), (\ref{sub rsc}) is further upper bouned by
\begin{align}
	\kappa_{\ell}\left\|\widehat{\Delta}_\eta\right\|_F^2 \le 10\lambda_n\sqrt{\frac{2rd_3}{\ell}} \left\|\widehat{\Delta}_\eta\right\|_F.\notag
\end{align}
So, $
\left\|\widehat{\Delta}_\eta\right\|_F^2 \leq C_1 \frac{rd_3\lambda_n^2}{\ell\kappa_\ell^2} 
$. If we choose $l>\sqrt{C_1 \frac{rd_3\lambda_n^2}{\ell\kappa_\ell^2} }$ in advance, we have $\widehat{\Delta}_\eta=\widehat{\Delta}$, which completes the proof.
\end{proof}
\subsection{Proof of of Theorem 2}
To upper bound the total regret, we first introduce the following lemma to analyze the regret of LowGLM-UCB, and then combine it with the regret of the exploration stage. 
\begin{lemma}[Theorem C.1 in \citep{kang2022efficient}]\label{lemma: 18}
\begin{spacing}{1.5}
	The regret of LowGLM-UCB with $\lambda_{\perp}=\frac{mT}{k \log \left(1+\frac{mT}{k\lambda}\right)}$ is, with probability at least $1-\delta$,
	$\widetilde{O}\left(k\sqrt{T}+T B_{\perp}\right)$.
 \end{spacing}
\end{lemma}
\vspace{-6mm}

It remains to determine $B_{\perp}:=\ell\left\|\widehat{\mathcal{U}}_{\perp}^{\top}*_L \mathcal{U}^*\right\|_F\left\|\widehat{\mathcal{V}}_{\perp}^{\top}*_L \mathcal{V}^{*}\right\|_F$. Before this, we first prove the error bound of the estimated parameter obtained from the exploration stage. By Theorem 1, we know that to bound $\|\widehat{\Delta}\|_F^2$, we should determine the $\lambda_n$, $\kappa_\ell$ and the $L(\mathcal{W})$ satisfy the $\text{LRSC}\left(\mathcal{C}(\mathcal{K},\widetilde{\mathcal{K}}^\perp,\mathcal{W}^*), \mathcal{N}, \kappa_{\ell}, \tau_{\ell}\right)$ condition with high probability. To this end, we state the following auxiliary lemmas. 

\begin{lemma}[Determine the $\lambda_{n}$]\label{lemma: lambda}
\begin{spacing}{1.5}
	Under the Assumptions 1 to 4, we have 
	\begin{align}
		\mathbb{P}\left(\left\|\frac{1}{n} \sum_{t=1}^n [b^\prime\left(\langle\mathcal{W}^*,\mathcal{X}_t\rangle\right)-y_t]\mathcal{X}_t\right\|_{2}\ge\sqrt{C_3\frac{\ell \log d_3}{arnd_3^2\min\{d_1,d_2\}(1-\gamma)^2}}\right)\le \delta,
		\notag
	\end{align} 
where $C_3$ is the constant related to the $\phi$ and $M$, and $\gamma=\frac{\sigma_1-\sigma_{\lceil ard_3\rceil}}{\sigma_1}\ (a\in[0,1])$ measure the ``gap" between the first and $\lceil ard_3\rceil$-th singular values of $\bar{\mathcal{X}}_t$.
\end{spacing}
\end{lemma}
\vspace{-6mm}

\begin{proof}
Introduce the shorthand $\mathcal{M}=\frac{1}{n} \sum_{t=1}^n z_t \mathcal{X}_t$, where $z_t=b^\prime\left(\langle\mathcal{W}^*,\mathcal{X}_t\rangle\right)-y_t$. Note that $$\left\|\mathcal{M}\right\|_2=\left\|\bar{\mathcal{M}}\right\|_2=\max_{i\in[d_3]}\left\|\breve{\mathcal{M}}^{(i)}\right\|_2,$$ 
we should analyze $\left\|\breve{\mathcal{M}}^{(i)}\right\|_2$. Denote $\left\{u^1, \ldots, u^M\right\}$ and $\left\{v^1, \ldots, v^N\right\}$ as the $1 / 4$-covers on $\mathbb{S}^{d_1-1}$ and $\mathbb{S}^{d_2-1}$, respectively, in which the $\mathbb{S}^{n-1}$ denotes the Euclidean unit sphere in $\mathbb{R}^n$. Then according to Lemma 5.7 in \cite{wainwright2019high}, we have $M \leq \left(\frac{2}{1/4}+1\right)^{d_1}=9^{d_1}$ and $N \leq 9^{d_2}$. For any $v \in \mathbb{S}^{d_2-1}$, there exists $v^k\in \left\{v^1, \ldots, v^N\right\}$ such that $\left\|v-v^{k}\right\|_2\le1/4$, and hence
\begin{align}
	&\left\|\breve{\mathcal{M}}^{(i)}\right\|_2
	=\sup _{v \in \mathbb{S}^{d_2-1}}\|\breve{\mathcal{M}}^{(i)} v\|_2 
	=\sup _{v \in \mathbb{S}^{d_2-1}}\|\breve{\mathcal{M}}^{(i)} (v-v^k+v^k)\|_2 \notag\\
	\leq& \frac{1}{4}\|\breve{\mathcal{M}}^{(i)}\|_2+\max _{k\in[ N]}\left\|\breve{\mathcal{M}}^{(i)} v^{k}\right\|_2 .\label{cover1}
\end{align}
Furthermore, for any $u \in \mathbb{S}^{d_1-1}$, there exists $u^j\in \left\{u^1, \ldots, v^M\right\}$ such that $\left\|u-u^{j}\right\|_2\le1/4$, so
\begin{align}
	&\left\|\breve{\mathcal{M}}^{(i)} v^{k}\right\|_2 
	=\sup _{u \in \mathbb{S}^{d_1-1}}\left\langle u,\breve{\mathcal{M}}^{(i)} v^{k}\right\rangle
	=\sup _{u \in \mathbb{S}^{d_1-1}}\left\langle u-u^j+u^j,\breve{\mathcal{M}}^{(i)} v^{k}\right\rangle\notag\\
	\leq& \frac{1}{4}\|\breve{\mathcal{M}}^{(i)}\|_2+\max _{j\in [M]}\left\langle u^j, \breve{\mathcal{M}}^{(i)} v^{k}\right\rangle.\label{cover2}
\end{align}
Combining (\ref{cover1}) with (\ref{cover2}) yields that
\begin{align}
	\|\breve{\mathcal{M}}^{(i)}\|_2 \leq 2 \max _{j\in[ M]} \max _{k\in[ N]}\left|{u^j}^{\top} \breve{\mathcal{M}}^{(i)} v^{k}\right| .\label{connection}
\end{align}
Due to ${u^j}^{\top} \breve{\mathcal{M}}^{(i)} v^{k}=\frac{1}{n}\sum_{t=1}^nz_t{u^j}^{\top}\breve{\mathcal{ X }}_t^{(i)}v^{k}$ and let $\gamma=\frac{\sigma_1-\sigma_{\lceil ard_3\rceil}}{\sigma_1}\ (a\in[0,1])$ measure the ``gap" between the first and $\lceil ard_3\rceil$-th singular values of $\bar{\mathcal{X}}_t$, we further analyze the property of ${u^j}^{\top}\breve{\mathcal{ X }}_t^{(i)}v^{k}$ as follows.
\begin{align}
	&\left|{u^j}^{\top} \breve{\mathcal{ X }}_t^{(i)} v^{k}\right|
	\le\left\|\breve{\mathcal{ X }}_t^{(i)}\right\|_2\le\left\|\bar{\mathcal{X}}_t\right\|_2
	\le\frac{1}{(1-\gamma)\sqrt{ard_3}}\left\|\bar{\mathcal{X}}_t\right\|_F
	=\frac{1}{1-\gamma}\sqrt{\frac{\ell}{ard_3}}\left\|\mathcal{X}_t\right\|_F\notag\\
	=&\frac{1}{1-\gamma}\sqrt{\frac{\ell}{ard_3}}\left\|\operatorname{vec}(\mathcal{X}_t)\right\|_2
	=\frac{1}{1-\gamma}\sqrt{\frac{\ell}{ard_3}}\sup_{\left\| \operatorname{vec}\left(\mathcal{A}\right)\right\|_2\le1}\left\langle \operatorname{vec}\left(\mathcal{X}_t\right), \operatorname{vec}\left(\mathcal{A}\right)\right\rangle.\notag
\end{align}
Since $ \operatorname{vec}(\mathcal{ X }_t)$ is $\frac{1}{\sqrt{d_1d_2d_3}}$-sub-Gaussian, we can obtain that $\left|{u^j}^{\top} \breve{\mathcal{ X }}_t^{(i)} v^{k}\right|$ is $\frac{1}{1-\gamma}\sqrt{\frac{\ell}{ard_1d_2d_3^2}}$-sub-Gaussian. Denote $\langle\mathcal{X}_t,\mathcal{W}^*\rangle=a_t$, it can be proved that
\begin{align}
	&\mathbb{E}[\exp(fz_t)\mid \mathcal{X}_t]\notag\\=&\int_yh(y)\exp\left[\frac{ya_t-b(a_t)}{\phi}\right]\exp(f(b^\prime(a_t)-y))dy\notag\\
	=&\int_yh(y)\exp\frac{1}{\phi}[y(a_t-\phi f)-b(a_t-\phi f)+b(a_t-\phi f)-b(a_t)+\phi fb^\prime(a_t)]dy\notag\\
	=&\exp\frac{1}{\phi}(b(a_t-\phi f)-b(a_t)+\phi fb^\prime(a_t)),\notag
\end{align}
and considering the convexity of $b(\cdot)$, we can further get that
\begin{align}
	&\mathbb{E}[\exp(fz_t)\mid \mathcal{X}_t]=\exp\frac{1}{\phi}(b(a_t-\phi f)-b(a_t)+\phi fb^\prime(a_t))\notag\\
 \le&\exp(-fb^\prime(a_t-\phi f)+fb^\prime(a_t))\le\exp(\phi f^2b^{\prime\prime}(a_t))\le\exp(f^2\phi M).
\end{align}
This yields that $z_t$ is $\sqrt{\phi M}$-sub-Gaussian, and thus 
$$
\left\|z_t\left|{u^j}^{\top} \breve{\mathcal{ X }}_t^{(i)} v^{k}\right|\right\|_{\psi_1}
\le\left\|z_t\right\|_{\psi_2}\left\|\left|{u^j}^{\top} \breve{\mathcal{ X }}_t^{(i)} v^{k}\right|\right\|_{\psi_2}\le\frac{1}{1-\gamma}\sqrt{\frac{\ell\phi M}{ard_1d_2d_3^2}},
$$
that is, $z_t\left|{u^j}^{\top} \breve{\mathcal{ X }}_t^{(i)} v^{k}\right|$ is $\frac{1}{1-\gamma}\sqrt{\frac{\ell\phi M}{ard_1d_2d_3^2}}$-sub-Exponential. Note that
$$
\mathbb{E}\left(z_t\left|{u^j}^{\top} \breve{\mathcal{ X }}_t^{(i)} v^{k}\right|\right)
=\mathbb{E}\left(\left|{u^j}^{\top} \breve{\mathcal{ X }}_t^{(i)} v^{k}\right|\mathbb{E}\left(z_t\mid\breve{\mathcal{ X }}_t^{(i)}\right)\right)=0,
$$
then according to (\ref{connection}) and Bernstein's inequality, we have
\begin{align}
	&\mathbb{P}\left(\left\|\breve{\mathcal{M}}^{(i)}\right\|_2\ge\frac{\lambda_n}{2}\right) 
	\le \sum_{j=1}^{M}\sum_{k=1}^{N}	\mathbb{P}\left(\frac{1}{n}\sum_{t=1}^nz_t\left|{u^j}^{\top}\breve{\mathcal{ X }}_t^{(i)}v^{k}\right|\ge\frac{\lambda_n}{2}\right) \notag\\
	\le& 2\cdot9^{d_1+d_2}\exp\left[-cn\min\left\{\frac{ar\lambda_{n}^2d_1d_2d_3^2(1-\gamma)^2}{4\ell\phi M},\sqrt{\frac{ar\lambda_{n}^2d_1d_2d_3^2(1-\gamma)^2}{4\ell \phi M}}\right\}\right].
	\notag
\end{align}
Thus it holds by union bound that \begin{align}&\mathbb{P}\left(\left\|\mathcal{M}\right\|_2\ge\frac{\lambda_n}{2}\right)\le\sum_{i=1}^{d_3} \mathbb{P}\left(\left\|\breve{\mathcal{M}}^{(i)}\right\|_2\ge\frac{\lambda_n}{2}\right)\notag\\
	\le &2d_3\cdot9^{d_1+d_2}\exp\left[-cn\min\left\{\frac{ar\lambda_{n}^2d_1d_2d_3^2(1-\gamma)^2}{4\ell\phi M},\sqrt{\frac{ar\lambda_{n}^2d_1d_2d_3^2(1-\gamma)^2}{4\ell \phi M}}\right\}\right].	\label{solve lambda}
\end{align}
Set the right-hand side of (\ref{solve lambda}) to equal $\delta$, assume that $n=\Omega(d_1+d_2+\log d_3)$, then 
\begin{align}
	\lambda_n=&c\sqrt{\frac{\ell\phi M}{ard_1d_2d_3^2(1-\gamma)^2}}\notag\\&\cdot\max\left\{\sqrt{\frac{(d_1+d_2)\log9+\log\frac{2}{\delta}+\log d_3}{n}},\frac{(d_1+d_2)\log9+\log\frac{2}{\delta}+\log d_3}{n}\right\}\notag\\
	=&c\sqrt{\frac{\ell\phi M}{ard_1d_2d_3^2(1-\gamma)^2}}\sqrt{\frac{(d_1+d_2)\log9+\log\frac{2}{\delta}+\log d_3}{n}}.\notag
\end{align}
That is to say, $\lambda_n^2=C_3\frac{\ell \log d_3}{arnd_3^2\min\{d_1,d_2\}(1-\gamma)^2}$, where $C_3$ is the constant related to $\phi$ and $M$.
\end{proof}
\begin{lemma}[The LRSC condition of $L(\mathcal{W})$]\label{lemma: lrsc condition}
\begin{spacing}{1.5}
		Define $\mathcal{K}$, $\widetilde{\mathcal{K}}^\perp$ as in (\ref{two subspaces}), and \begin{small}$\mathcal{N}:=\left\{\mathcal{W}\in\mathbb{R}^{d_1\times d_2 \times d_3}:\|\mathcal{W}-\mathcal{W}^*\|_F^2\le C_1 \frac{rd_3\lambda_n^2}{\ell\kappa_\ell^2},\mathcal{W}-\mathcal{W}^*\in\left(\mathcal{C}(\mathcal{K},\widetilde{\mathcal{K}}^\perp,\mathcal{W}^*), \mathcal{N}, \kappa_{\ell}\right)\right\}$\end{small}, then $L(\mathcal{W})$ satisfy the $\text{LRSC}\left(\mathcal{C}(\mathcal{K},\widetilde{\mathcal{K}}^\perp,\mathcal{W}^*), \mathcal{N}, \kappa_{\ell}\right)$ condition.
  \end{spacing}
\end{lemma}
\vspace{-6mm}

\begin{proof}
Denote $\mathcal{W}=\mathcal{W}^*+\Delta$, and according to the Taylor's theorem with Lagrange remainder, we have
\begin{align}
	 &L\left(\mathcal{W}^*+\Delta\right)-L\left(\mathcal{W}^*\right)-\left\langle\nabla L\left(\mathcal{W}^*\right), \Delta\right\rangle \notag\\
	= & \frac{1}{n} \sum_{t=1}^n b^{\prime \prime}\left(\left\langle\mathcal{W}_\eta, \mathcal{X}_t\right\rangle\right)\left\langle\mathcal{X}_t, \Delta\right\rangle^2 
	=  \operatorname{vec}(\Delta)^\top \cdot \widehat{H}\left(\mathcal{W}_\eta\right) \cdot \operatorname{vec}(\Delta) .\notag
\end{align}
So in order to prove the LRSC condition, we only need to analyze the lower bound of $ \operatorname{vec}(\Delta)^\top \cdot \widehat{H}\left(\mathcal{W}_\eta\right) \cdot \operatorname{vec}(\Delta) $. For this, we proceed the proof in two steps. Specifically, we first lower bound the $ \operatorname{vec}(\Delta)^\top \cdot \widehat{H}\left(\mathcal{W}^*\right) \cdot \operatorname{vec}(\Delta) $ on the restricted set $\mathcal{C}(\mathcal{K},\widetilde{\mathcal{K}}^\perp,\mathcal{W}^*)$; then we lower bound the $ \operatorname{vec}(\Delta)^\top \cdot \widehat{H}\left(\mathcal{W}_\eta\right) \cdot \operatorname{vec}(\Delta) $   on the restricted set $\mathcal{C}(\mathcal{K},\widetilde{\mathcal{K}}^\perp,\mathcal{W}^*)$, where $\mathcal{W}_\eta$ is the tensor nuclear norm neighborhood of $\mathcal{W}^*$.
\begin{itemize}
	\item Step 1:  Lower bound the $ \operatorname{vec}(\Delta)^\top \cdot \widehat{H}\left(\mathcal{W}^*\right) \cdot \operatorname{vec}(\Delta) $ on the restricted set $\mathcal{C}(\mathcal{K},\widetilde{\mathcal{K}}^\perp,\mathcal{W}^*)$.
\end{itemize}
Denote the transformed t-SVD of $\Delta$ as $\Delta=\mathcal{U}*_L\mathcal{S}*_L\mathcal{V}^\top$. Due to the unitarily invariance of Froebenius norm, we can easily obtain hat $\|\Delta\|_F=\|\operatorname{vec}(\mathcal{S})\|_2$. So that 
\begin{align}
	 &\operatorname{vec}(\Delta)^\top \cdot \widehat{H}\left(\mathcal{W}^*\right) \cdot \operatorname{vec}(\Delta)
	 	=  \frac{1}{n} \sum_{t=1}^n b^{\prime \prime}\left(\left\langle\mathcal{W}^*, \mathcal{X}_t\right\rangle\right)\left\langle\mathcal{X}_t, \Delta\right\rangle^2 \notag\\
	 		= & \frac{1}{n} \sum_{t=1}^n \left[\sqrt{b^{\prime \prime}\left(\left\langle\mathcal{W}^*, \mathcal{X}_t\right\rangle\right)}\left\langle\mathcal{X}_t, \mathcal{U}*_L\mathcal{S}*_L\mathcal{V}^\top\right\rangle\right]^2 \notag\\
	 		=& \frac{1}{n} \sum_{t=1}^n \left[\sqrt{b^{\prime \prime}\left(\left\langle\mathcal{W}^*, \mathcal{X}_t\right\rangle\right)}\left\langle\mathcal{U}^\top*_L\mathcal{X}_t*_L\mathcal{V}, \mathcal{S}\right\rangle\right]^2\notag\\
	 		=& \frac{1}{n} \sum_{t=1}^n\left\langle\tilde{\mathcal{X}}_t, \mathcal{S}\right\rangle^2
	 		=\operatorname{vec}(\mathcal{S})^\top \cdot\widehat{ \Sigma}_{\tilde{\mathcal{X}} \tilde{\mathcal{X}}} \cdot \operatorname{vec}(\mathcal{S}) \notag\\
	 		=&\operatorname{vec}(\mathcal{S})^\top \cdot \Sigma_{\tilde{\mathcal{X}} \tilde{\mathcal{X}}} \cdot \operatorname{vec}(\mathcal{S}) 
	 		 +\operatorname{vec}(\mathcal{S})^\top \cdot\left(\widehat{\Sigma}_{\tilde{\mathcal{X}} \tilde{\mathcal{X}}}-\Sigma_{\tilde{\mathcal{X}} \tilde{\mathcal{X}}}\right) \cdot \operatorname{vec}(\mathcal{S}),\label{covar}
\end{align}
where $\tilde{\mathcal{X}}_t=\sqrt{b^{\prime \prime}\left(\left\langle\mathcal{W}^*, \mathcal{X}_t\right\rangle\right)}\cdot \mathcal{U}^\top*_L\mathcal{X}_t*_L\mathcal{V}$, $\widehat{\Sigma}_{\tilde{\mathcal{X}} \tilde{\mathcal{X}}}=$ $\frac{1}{n} \sum_{t=1}^n \operatorname{vec}\left(\tilde{\mathcal{X}}_t\right) \cdot \operatorname{vec}\left(\tilde{\mathcal{X}}_t\right)^\top$ and $\mathbb{E}(\widehat{\Sigma}_{\tilde{\mathcal{X}} \tilde{\mathcal{X}}})=\Sigma_{\tilde{\mathcal{X}} \tilde{\mathcal{X}}}$. Further due to Cauchy-Schwarz inequality, and the fact that
$\|\operatorname{vec}(\mathcal{S})\operatorname{vec}(\mathcal{S})^\top\|_1=\|\mathcal{S}\|_1^2=\|\mathcal{S}_{(3)}\|_1^2\le(\sum_{j=1}^{n_1n_2}\sqrt{d_3}\|\mathcal{S}_{(3)}^j\|_2)^2=(\sum_{j=1}^{n_1n_2}\sqrt{d_3}\|\frac{L}{\sqrt{\ell}}\mathcal{S}_{(3)}^j\|_2)^2=(\sum_{j=1}^{n_1n_2}\sqrt{\frac{d_3}{\ell}}\|L\mathcal{S}_{(3)}^j\|_2)^2\le(\sum_{j=1}^{n_1n_2}\sqrt{\frac{d_3}{\ell}}\|L\mathcal{S}_{(3)}^j\|_1)^2=\frac{d_3}{\ell}\|\bar{\mathcal{S}}\|_1^2=d_3\ell\|\Delta\|_*^2$, we lower bound (\ref{covar}) by
\begin{align}
&\operatorname{vec}(\Delta)^\top \cdot \widehat{H}\left(\mathcal{W}^*\right) \cdot \operatorname{vec}(\Delta)\notag\\
\ge& \lambda_{\min}(\Sigma_{\tilde{\mathcal{X}} \tilde{\mathcal{X}}} )\|\Delta\|_F^2-\left|\operatorname{vec}(\mathcal{S})^\top \cdot\left(\widehat{\Sigma}_{\tilde{\mathcal{X}} \tilde{\mathcal{X}}}-\Sigma_{\tilde{\mathcal{X}} \tilde{\mathcal{X}}}\right) \cdot \operatorname{vec}(\mathcal{S})\right|\notag\\
\ge& \lambda_{\min}(\Sigma_{\tilde{\mathcal{X}} \tilde{\mathcal{X}}} )\|\Delta\|_F^2-\left\|\widehat{\Sigma}_{\tilde{\mathcal{X}} \tilde{\mathcal{X}}}-\Sigma_{\tilde{\mathcal{X}} \tilde{\mathcal{X}}}\right\|_\infty\|\operatorname{vec}(\mathcal{S})\operatorname{vec}(\mathcal{S})^\top\|_1\notag\\
\ge &\lambda_{\min}(\Sigma_{\tilde{\mathcal{X}} \tilde{\mathcal{X}}} )\|\Delta\|_F^2-d_3\ell\left\|\widehat{\Sigma}_{\tilde{\mathcal{X}} \tilde{\mathcal{X}}}-\Sigma_{\tilde{\mathcal{X}} \tilde{\mathcal{X}}}\right\|_\infty\|\Delta\|_*^2.\label{step 1 total}
\end{align}
It remains to analyze $\lambda_{\min}(\Sigma_{\tilde{\mathcal{X}} \tilde{\mathcal{X}}} )$ and $\left\|\widehat{\Sigma}_{\tilde{\mathcal{X}} \tilde{\mathcal{X}}}-\Sigma_{\tilde{\mathcal{X}} \tilde{\mathcal{X}}}\right\|_\infty$. According to Assumption 3, the following inequality holds.
\begin{align}
	 &\lambda_{\min }\left(\Sigma_{\tilde{\mathcal{X}} \tilde{\mathcal{X}}}\right) \notag\\
	 =&\inf _{\substack{\mathcal{Z}_1, \mathcal{Z}_2 \in \mathbb{R}^{d_1 \times d_2 \times d_3} \\
			\left\|\boldsymbol{Z}_1\right\|_F=\left\|\mathcal{Z}_2\right\|_F=1}} \operatorname{vec}\left(\mathcal{Z}_1\right)^\top \Sigma_{\tilde{\mathcal{X}} \tilde{\mathcal{X}}}\operatorname{vec}\left(\mathcal{Z}_2\right) \notag\\
	 =&\inf_{\substack{\mathcal{Z}_1, \mathcal{Z}_2 \in \mathbb{R}^{d_1 \times d_2 \times d_3} \\
	 		\left\|\boldsymbol{Z}_1\right\|_F=\left\|\mathcal{Z}_2\right\|_F=1}} \mathbb{E}\left[b^{\prime \prime}\left(\left\langle\mathcal{W}^*, \mathcal { X }_t\right\rangle\right)\left\langle\mathcal{U}^\top *_L \mathcal{X}_t *_L \mathcal{V}, \mathcal{Z}_1\right\rangle\left\langle\mathcal{U}^\top *_L \mathcal{X}_t*_L \mathcal{V}, \mathcal{Z}_2\right\rangle\right] \notag\\
	 =&\inf_{\substack{\mathcal{Z}_1, \mathcal{Z}_2 \in \mathbb{R}^{d_1 \times d_2 \times d_3} \\
			\left\|\boldsymbol{Z}_1\right\|_F=\left\|\mathcal{Z}_2\right\|_F=1}} \mathbb{E}\left[b^{\prime \prime}\left(\left\langle\mathcal{W}^*, \mathcal { X }_t\right\rangle\right)\left\langle\mathcal{X}_t, \mathcal{U} *_L \mathcal{Z}_1 *_L \mathcal{V}^\top\right\rangle\left\langle\mathcal{X}_t, \mathcal{U} *_L \mathcal{Z}_2 *_L \mathcal{V}^\top\right\rangle\right] \notag\\
	 =&\inf _{\substack{\mathcal{Z}_1, \mathcal{Z}_2 \in \mathbb{R}^{d_1 \times d_2 \times d_3} \\
			\left\|\boldsymbol{Z}_1\right\|_F=\left\|\mathcal{Z}_2\right\|_F=1}}\operatorname{vec}\left(\mathcal{U} *_L \mathcal{Z}_1 *_L \mathcal{V}^\top\right)^\top \cdot \widehat{H}\left(\mathcal{W}^*\right) \cdot \operatorname{vec}\left(\mathcal{U} *_L \mathcal{Z}_2 *_L \mathcal{V}^\top\right)\notag \\
	 =&\lambda_{\min }\left(\widehat{H}\left(\mathcal{W}^*\right) \right) \geq \kappa_\ell=\frac{1}{d_1d_2d_3} . \label{step 1 1}
\end{align}
We next to use the covering argument to bound $\left\|\widehat{\Sigma}_{\tilde{\mathcal{X}} \tilde{\mathcal{X}}}-\Sigma_{\tilde{\mathcal{X}} \tilde{\mathcal{X}}}\right\|_\infty$.

Let $\mathbb{S}^{d \times 1 \times d_3}=\left\{\mathcal{U} \in \mathbb{R}^{d \times 1 \times d_3}:\|\mathcal{U}\|_F=1\right\}$, and $\mathbb{N}^{d \times 1 \times d_3}$ be the $1 / 8$-covers on $\mathbb{S}^{d \times 1 \times d_3}$. Then according to Lemma 5.7 in \cite{wainwright2019high}, we have $|\mathbb{N}^{d \times 1 \times d_3}| \leq \left(\frac{2}{1/8}+1\right)^{dd_3}=17^{dd_3}$, and for any $\mathcal{U}_1, \mathcal{U}_2 \in \mathbb{S}^{d_1\times1\times d_3}$ and $\mathcal{V}_1, \mathcal{V}_2 \in \mathbb{S}^{d_2\times1\times d_3}$, there exists $\grave{\mathcal{U}}_1, \grave{\mathcal{U}}_2 \in \mathbb{N}^{d_1\times1\times d_3}$ and $\grave{\mathcal{V}}_1, \grave{\mathcal{V}}_2 \in \mathbb{N}^{d_2\times1\times d_3}$ such that $\left\|\mathcal{U}_i-\grave{\mathcal{U}}_i\right\|_F\le1/8$ and $\left\|\mathcal{V}_i-\grave{\mathcal{V}}_i\right\|_F\le1/8$ for $i=1,2$. For any $A \in \mathbb{R}^{d_1d_2 d_3 \times d_1d_2 d_3}$, define
\begin{align}
	& \Phi(A):=\sup _{\substack{\mathcal{U}_1, \mathcal{U}_2 \in \mathbb{S}^{d_1 \times 1 \times d_3} \\
			\mathcal{V}_1, \mathcal{V}_2 \in \mathbb{S}^{d_2 \times 1 \times d_3}}} \operatorname{vec}\left(\mathcal{U}_1 *_L \mathcal{V}_1^\top\right)^\top \cdot A \cdot \operatorname{vec}\left(\mathcal{U}_2 *_L \mathcal{V}_2^\top\right), \notag\\
	& \Phi_{\mathbb{N}}(A):=\sup _{\substack{\mathcal{U}_1, \mathcal{U}_2 \in \mathbb{N}^{d_1 \times 1 \times d_3} \\
			\mathcal{V}_1, \mathcal{V}_2 \in \mathbb{N}^{d_2 \times 1 \times d_3}}} \operatorname{vec}\left(\mathcal{U}_1 *_L \mathcal{V}_1^\top\right)^\top \cdot A \cdot \operatorname{vec}\left(\mathcal{U}_2 *_L \mathcal{V}_2^\top\right) . \notag
\end{align}
Then it follows that 
\begin{small}
\begin{align}
	&\operatorname{vec}\left(\mathcal{U}_1 *_L \mathcal{V}_1^\top\right)^\top \cdot A \cdot \operatorname{vec}\left(\mathcal{U}_2 *_L \mathcal{V}_2^\top\right)\notag\\
	=&\operatorname{vec}\left((\mathcal{U}_1-\grave{\mathcal{U}}_1+\grave{\mathcal{U}}_1) *_L (\mathcal{V}_1-\grave{\mathcal{V}}_1+\grave{\mathcal{V}}_1)^\top\right)^\top \cdot A \cdot \operatorname{vec}\left((\mathcal{U}_2-\grave{\mathcal{U}}_2+\grave{\mathcal{U}}_2) *_L (\mathcal{V}_2-\grave{\mathcal{V}}_2+\grave{\mathcal{V}}_2)^\top\right)\notag\\
	=&\operatorname{vec}\left(\grave{\mathcal{U}}_1*_L\grave{\mathcal{V}}_1^\top+(\mathcal{U}_1-\grave{\mathcal{U}}_1)*_L\grave{\mathcal{V}}_1^\top+\mathcal{U}_1*_L(\mathcal{V}_1-\grave{\mathcal{V}}_1)^\top\right)^\top \cdot A \cdot \notag\\
 &\operatorname{vec}\left(\grave{\mathcal{U}}_2*_L\grave{\mathcal{V}}_2^\top+(\mathcal{U}_2-\grave{\mathcal{U}}_2)*_L\grave{\mathcal{V}}_2^\top+\mathcal{U}_2*_L(\mathcal{V}_2-\grave{\mathcal{V}}_2)^\top\right)\notag\\	
	\le& \Phi_{\mathbb{N}}(A)+\frac{1}{2}\Phi(A)+\frac{1}{16}\Phi(A).\notag
\end{align}
\end{small}
That is to say,
$$
\Phi(A) \leq(16 / 7) \Phi_{\mathbb{N}}(A).
$$
So that, combined with the union bound, we have
\begin{align}
&	\mathbb{P}\left(\left\|\widehat{\Sigma}_{\tilde{\mathcal{X}} \tilde{\mathcal{X}}}-\Sigma_{\tilde{\mathcal{X}} \tilde{\mathcal{X}}}\right\|_\infty\ge t\right)\le\mathbb{P}\left(\Phi(\widehat{\Sigma}_{\tilde{\mathcal{X}} \tilde{\mathcal{X}}}-\Sigma_{\tilde{\mathcal{X}} \tilde{\mathcal{X}}})\ge t\right)\notag\\
	\le& \mathbb{P}\left((16 / 7) \Phi_{\mathbb{N}}(\widehat{\Sigma}_{\tilde{\mathcal{X}} \tilde{\mathcal{X}}}-\Sigma_{\tilde{\mathcal{X}} \tilde{\mathcal{X}}})\ge t\right)\notag\\
	=&\mathbb{P}\left(\sup _{\substack{\mathcal{U}_1, \mathcal{U}_2 \in \mathbb{N}^{d_1 \times 1 \times d_3} \\
			\mathcal{V}_1, \mathcal{V}_2 \in \mathbb{N}^{d_2 \times 1 \times d_3}}} \operatorname{vec}\left(\mathcal{U}_1 *_L \mathcal{V}_1^\top\right)^\top \cdot (\widehat{\Sigma}_{\tilde{\mathcal{X}} \tilde{\mathcal{X}}}-\Sigma_{\tilde{\mathcal{X}} \tilde{\mathcal{X}}}) \cdot \operatorname{vec}\left(\mathcal{U}_2 *_L \mathcal{V}_2^\top\right) 
	\ge(7/16) t\right)\notag\\
	\le&17^{d_1d_3+d_2d_3}\mathbb{P}\left(\left| \operatorname{vec}\left(\mathcal{U}_1 *_L \mathcal{V}_1^\top\right)^\top \cdot (\widehat{\Sigma}_{\tilde{\mathcal{X}} \tilde{\mathcal{X}}}-\Sigma_{\tilde{\mathcal{X}} \tilde{\mathcal{X}}}) \cdot \operatorname{vec}\left(\mathcal{U}_2 *_L \mathcal{V}_2^\top\right) \right|
	\ge(7/16) t\right).\label{infty}
\end{align}
Lemma 5.14 in \cite{vershynin2010introduction} further yields that for any $\mathcal{U}_1, \mathcal{U}_2  \in\mathbb{S}^{d_1 \times 1 \times d_3}, \mathcal{V}_1, \mathcal{V}_2 \in\mathbb{S}^{d_2 \times 1 \times d_3}$, we have
\begin{align}
	& \left\|\left\langle\mathcal{U}_1 *_L \mathcal{V}_1^\top, \tilde{\mathcal{X}}_t\right\rangle\left\langle\mathcal{U}_2 *_L \mathcal{V}_2^\top, \tilde{\mathcal{X}}_t\right\rangle\right\|_{\psi_1} \notag\\
	\leq & \frac{1}{2}\left(\left\|\left\langle\mathcal{U}_1 *_L \mathcal{V}_1^\top, \tilde{\mathcal{X}}_t\right\rangle^2\right\|_{\psi_1}+\left\|\left\langle\mathcal{U}_2 *_L \mathcal{V}_2^\top, \tilde{\mathcal{X}}_t\right\rangle^2\right\|_{\psi_1}\right) \notag\\
	\leq & \left\|\left\langle\mathcal{U}_1 *_L \mathcal{V}_1^\top, \sqrt{b^{\prime \prime}\left(\left\langle\mathcal{W}^*, \mathcal{X}_t\right\rangle\right)} \cdot \mathcal{U}^\top *_L \mathcal{\mathcal { X }}_t *_L \mathcal{V}\right\rangle\right\|_{\psi_2}^2 \notag\\
	 &+\left\|\left\langle\mathcal{U}_2 *_L \mathcal{V}_2^\top, \sqrt{b^{\prime \prime}\left(\left\langle\mathcal{W}^*, \mathcal{X}_t\right\rangle\right)} \cdot \mathcal{U}^\top *_L \mathcal{X}_t *_L \mathcal{V}\right\rangle\right\|_{\psi_2}^2 
	\leq  \frac{2 M}{d_1d_2d_3} .\notag
\end{align}
Considering the Bernstein inequality, we have
\begin{align}
	&\mathbb{P}\left(\left| \operatorname{vec}\left(\mathcal{U}_1 *_L \mathcal{V}_1^\top\right)^\top \cdot (\widehat{\Sigma}_{\tilde{\mathcal{X}} \tilde{\mathcal{X}}}-\Sigma_{\tilde{\mathcal{X}} \tilde{\mathcal{X}}}) \cdot \operatorname{vec}\left(\mathcal{U}_2 *_L \mathcal{V}_2^\top\right) \right|
	\ge\frac{7}{16} t\right)\notag\\
	  \leq& 2 \exp \left[-cn \min \left(\frac{49 t^2d_1^2d_2^2d_3^2}{1024M^2 }, \frac{7 td_1d_2d_3}{32M }\right)\right].\label{infty bernstein}
\end{align}
Substituting (\ref{infty bernstein}) into (\ref{infty}) gives that
\begin{align}
&\mathbb{P}\left(\left\|\widehat{\Sigma}_{\tilde{\mathcal{X}} \tilde{\mathcal{X}}}-\Sigma_{\tilde{\mathcal{X}} \tilde{\mathcal{X}}}\right\|_\infty\ge t\right)\notag\\
\le&17^{d_1d_3+d_2d_3}\cdot2 \exp \left[-cn \min \left(\frac{49 t^2d_1^2d_2^2d_3^2}{1024M^2 }, \frac{7 td_1d_2d_3}{32M }\right)\right]\notag\\
=&2 \exp \left[(d_1+d_2)d_3\log 17-cn \min \left(\frac{49 t^2d_1^2d_2^2d_3^2}{1024M^2 }, \frac{7 td_1d_2d_3}{32M }\right)\right].\label{cov bernstein}
\end{align}
Set the right-hand side of (\ref{cov bernstein}) to equal $\delta$, assume that $n=\Omega((d_1+d_2)d_3)$, then
\begin{align}
t=&C_4\frac{M}{d_1d_2d_3}\max\left\{\sqrt{\frac{(d_1+d_2)d_3\log 17+\log\frac{2}{\delta}}{n}},\frac{(d_1+d_2)d_3\log 17+\log\frac{2}{\delta}}{n}\right\}\notag\\
=&C_4\frac{M}{d_1d_2d_3}\sqrt{\frac{(d_1+d_2)d_3\log 17+\log\frac{2}{\delta}}{n}}.\notag
\end{align}
That is to say, with probability at least $1-\delta$, 
\begin{align}
	\left\|\widehat{\Sigma}_{\tilde{\mathcal{X}} \tilde{\mathcal{X}}}-\Sigma_{\tilde{\mathcal{X}} \tilde{\mathcal{X}}}\right\|_\infty\le\frac{C_4}{\sqrt{nd_3\min\{d_1,d_2\}^3}}.\label{step 1 2}
\end{align}
Substituting (\ref{step 1 1}) and (\ref{step 1 2}) into (\ref{step 1 total}) yields that
\begin{align}
	\operatorname{vec}(\Delta)^\top \cdot \widehat{H}\left(\mathcal{W}^*\right) \cdot \operatorname{vec}(\Delta)
&	\ge \lambda_{\min}(\Sigma_{\tilde{\mathcal{X}} \tilde{\mathcal{X}}} )\|\Delta\|_F^2-d_3\ell\left\|\widehat{\Sigma}_{\tilde{\mathcal{X}} \tilde{\mathcal{X}}}-\Sigma_{\tilde{\mathcal{X}} \tilde{\mathcal{X}}}\right\|_\infty\|\Delta\|_*^2\notag\\
&	\ge\frac{1}{d_1d_2d_3}\|\Delta\|_F^2-\frac{C_4d_3\ell}{\sqrt{nd_3\min\{d_1,d_2\}^3}}\|\Delta\|_*^2.\notag
\end{align}
Considering that $\Delta\in\mathcal{C}(\mathcal{K},\widetilde{\mathcal{K}}^\perp,\mathcal{W}^*)$, Lemma \ref{restricted set} states that
$\left\|\Delta_{\widetilde{\mathcal{K}}^\perp}\right\|_{*} \leq 3\left\|\Delta_{\widetilde{\mathcal{K}}}\right\|_{*}$, thus $\|\Delta\|_*\le\|\Delta_{\widetilde{\mathcal{K}}^\perp}\|_*+\|\Delta_{\widetilde{\mathcal{K}}}\|_*\le4\|\Delta_{\widetilde{\mathcal{K}}}\|_*\le4\sqrt{\frac{2rd_3}{\ell}}\|\Delta_{\widetilde{\mathcal{K}}}\|_F\le4\sqrt{\frac{2rd_3}{\ell}}\|\Delta\|_F$. If $C_4=O(\sqrt{\frac{n}{r^2d_3^5\max\{d_1,d_2\}}})$, we finally obtain that 
$$	\operatorname{vec}(\Delta)^\top \cdot \widehat{H}\left(\mathcal{W}^*\right) \cdot \operatorname{vec}(\Delta)\ge\left(\frac{1}{d_1d_2d_3}-\frac{32C_4rd_3^{\frac{3}{2}}}{\sqrt{n\min\{d_1,d_2\}^3}}\right)\|\Delta\|_F^2\ge\frac{C_5}{d_1d_2d_3}\|\Delta\|_F^2.$$
\begin{itemize}
	\item Step 2: Lower bound the $ \operatorname{vec}(\Delta)^\top \cdot \widehat{H}\left(\mathcal{W}_\eta\right) \cdot \operatorname{vec}(\Delta) $ on the restricted set $\mathcal{C}(\mathcal{K},\widetilde{\mathcal{K}}^\perp,\mathcal{W}^*)$, where $\mathcal{W}_\eta$ is the tensor nuclear norm neighborhood of $\mathcal{W}^*$.
\end{itemize}
\begin{spacing}{1.5}
Define $\widehat{h}(\mathcal{W})=\frac{1}{n} \sum_{t=1}^n b^{\prime \prime}\left(\left\langle\mathcal{W}, \mathcal{X}_t\right\rangle\right) \cdot1_{\left\{\left|\left\langle\mathcal{W}^*, \mathcal{X}_t\right\rangle\right|>\tau\left\|\mathcal{X}_t\right\|_{2}\right\}} \cdot \operatorname{vec}\left(\mathcal { X }_t\right) \operatorname{vec}\left(\mathcal { X }_t\right)^\top$ and $h(\mathcal{W})=$ $\mathbb{E}(\widehat{h}(\mathcal{W}))$ for constant $\tau$. $\widehat{H}(\mathcal{ W})\succeq\widehat{h}(\mathcal{ W})$ indicates thst $\operatorname{vec}(\Delta)^\top\cdot\widehat{H}(\mathcal{ W})\cdot\operatorname{vec}(\Delta)\ge\operatorname{vec}(\Delta)^\top\cdot\widehat{h}(\mathcal{ W})\cdot \operatorname{vec}(\Delta)$. That is to say, for lower bounding the $\operatorname{vec}(\Delta)^\top\cdot\widehat{H}(\mathcal{ W}_\eta)\cdot \operatorname{vec}(\Delta)$, we only need to lower bound $\operatorname{vec}(\Delta)^\top\cdot\widehat{h}(\mathcal{ W}_\eta)\cdot \operatorname{vec}(\Delta)$. Similarly, we first bound the $\operatorname{vec}(\Delta)^\top\cdot\widehat{h}(\mathcal{ W}^*)\cdot \operatorname{vec}(\Delta)$ and then bound the difference between $\operatorname{vec}(\Delta)^\top\cdot\widehat{h}(\mathcal{ W}_\eta)\cdot \operatorname{vec}(\Delta)$ and $\operatorname{vec}(\Delta)^\top\cdot\widehat{h}(\mathcal{ W}^*)\cdot \operatorname{vec}(\Delta)$. We can similarly apply the proof procedure in Step 1, before that, we need to solve $\lambda_{\min }\left(\Sigma_{\tilde{\mathcal{X}} \tilde{\mathcal{X}}}\cdot1_{\left\{\left|\left\langle\mathcal{W}^*, \mathcal{X}_t\right\rangle\right|>\tau\left\|\mathcal{X}_t\right\|_{2}\right\}}\right) $ by  bounding $\lambda_{\min}(h(\mathcal{W}^*))$. Note that for any $v \in \mathbb{R}^{d_1d_2 d_3}$, 
\end{spacing}
\begin{small}
\begin{align}
	& n v^\top h\left(\mathcal{W}^*\right)v \notag\\
	= & \mathbb{E}\left[\sum_{t=1}^n b^{\prime \prime}\left(\left\langle\mathcal{W}^*, \mathcal{X}_t\right\rangle\right)\left(\operatorname{vec}\left(\mathcal{X}_t\right)^\top v\right)^2\right] \notag\\
	 &-\mathbb{E}\left[\sum_{t=1}^n b^{\prime \prime}\left(\left\langle\mathcal{W}^*, \mathcal{X}_t\right\rangle\right) \cdot1_{\left\{\left|\left\langle\mathcal{W}^*, \mathcal{X}_t\right\rangle\right|\le\tau\left\|\mathcal{X}_t\right\|_{2}\right\}} \cdot\left(\operatorname{vec}\left(\mathcal{X}_t\right)^\top v\right)^2\right] \notag\\
	\ge & n v^\top \widehat{H}\left(\mathcal{W}^*\right) v 
	 -\sqrt{\mathbb{E}\left[\sum_{t=1}^n b^{\prime \prime}\left(\left\langle\mathcal{W}^*, \mathcal{X}_t\right\rangle\right)^2\left(\operatorname{vec}\left(\mathcal{X}_t\right)^\top v\right)^4\right]} \cdot \sqrt{\mathbb{E} \sum_{t=1}^n 1_{\left\{\left|\left\langle\mathcal{W}^*, \mathcal{X}_t\right\rangle\right|\le\tau\left\|\mathcal{X}_t\right\|_{2}\right\}}} \notag\\
	\ge & n \frac{1}{d_1d_2d_3}\|v\|_2^2-n M K \sqrt{\mathbb{P}\left\{\left|\left\langle\mathcal{W}^*, \mathcal{X}_t\right\rangle\right|\le\tau\left\|\mathcal{X}_t\right\|_{2}\right\}}\|v\|_2^2, \label{indicator}
\end{align}
\end{small}
where $M$ is the upper bound of $b^{\prime \prime}(\cdot)$ and $K$ is the largest eigenvalue of the fourth moment of $\|\operatorname{vec}(\mathcal{X}_t)\|_2$. According to the Bonferroni inequality, it holds that
\begin{align}
	&\mathbb{P}\left\{\left|\left\langle\mathcal{W}^*, \mathcal{X}_t\right\rangle\right|>\frac{c_1}{c_2}\left\|\mathcal{X}_t\right\|_{2}\right\}\notag\\
	\ge&	\mathbb{P}\left\{\left|\left\langle\mathcal{W}^*, \mathcal{X}_t\right\rangle\right|>c_1\sqrt{\frac{\ell}{d_1d_2d_3}\left[d_1+d_2+\log \frac{d_3}{p_0+1}\right]}\right\}\notag\\
 &+	\mathbb{P}\left\{\left\|\mathcal{X}_t\right\|_{2}\le{c_2\sqrt{\frac{\ell}{d_1d_2d_3}\left[d_1+d_2+\log \frac{d_3}{p_0+1}\right]}}\right\}. \label{bon}
\end{align}
Recall that $\left\|\mathcal{X}_t\right\|_2=\left\|\bar{\mathcal{X}}_t\right\|_2=\max_{i\in[d_3]}\left\|\breve{\mathcal{X}}_t^{(i)}\right\|_2$, and $$	\|\breve{\mathcal{X}}_t^{(i)}\|_2 \leq 2 \max _{j\in[ M]} \max _{k\in[ N]}\left|{u^j}^{\top} \breve{\mathcal{X}}_t^{(i)} v^{k}\right| .$$ Then according to $\left|{u^j}^{\top} \breve{\mathcal{ X }}_t^{(i)} v^{k}\right|$ is $\sqrt{\frac{\ell}{d_1d_2d_3}}$-sub-Gaussian, we have
\begin{align}
	\mathbb{P}(\left|{u^j}^{\top} \breve{\mathcal{ X }}_t^{(i)} v^{k}\right|>t)<2\exp(-\frac{C_6t^2d_1d_2d_3}{\ell}).\notag
\end{align}
Hence, we have
\begin{align}
	\mathbb{P}(\left\|\mathcal{X}_t\right\|_2>t)
	=	\mathbb{P}(\max_{i\in[d_3]}\left\|\breve{\mathcal{X}}_t^{(i)}\right\|_2>t)
	<2d_3\cdot 9^{d_1+d_2}\exp(-\frac{C_6t^2d_1d_2d_3}{\ell}).\label{bon 1}
\end{align}
Setting the right-hand-side of (\ref{bon 1}) as $(p_0+1)/2$, that is, $$t=c_2\sqrt{\frac{\ell}{d_1d_2d_3}\left[d_1+d_2+\log \frac{d_3}{p_0+1}\right]}.$$ $\left\langle\mathcal{W}^*, \mathcal{X}_t\right\rangle$ is a sub-Gaussian variable because it is a linear transformation of a sub-Gaussian vector, and its mean is 0 and sub-Gaussian norm is bounded by $\frac{1}{\sqrt{d_1d_2d_3}}\|\mathcal{W}^*\|_F$. When $\|\mathcal{W}\|^*_F \geq \alpha \sqrt{\ell\left[d_1+d_2+\log \frac{d_3}{p_0+1}\right]}$, taking a sufficiently small $c_1$, we obtain that
\begin{align}
&\mathbb{P}\left(\left|\left\langle\mathcal{W}^*, \mathcal{X}_t\right\rangle\right|>c_1 \sqrt{\frac{\ell}{d_1d_2d_3}\left[d_1+d_2+\log \frac{d_3}{p_0+1}\right]}\right) \notag\\
	\geq  &\mathbb{P}\left(\left|\left\langle\mathcal{W}^*, \mathcal{X}_t\right\rangle\right|>\frac{c_1}{\alpha\sqrt{d_1d_2d_3}}\|\mathcal{W}^*\|_F \right) 
	\geq  \frac{p_0+1}{2} .\label{bon 2}
\end{align}
Therefore, altogether (\ref{bon 1}) and (\ref{bon 2}), (\ref{bon}) can be bounded by $$\mathbb{P}\left\{\left|\left\langle\mathcal{W}^*, \mathcal{X}_t\right\rangle\right|>\frac{c_1}{c_2}\left\|\mathcal{X}_t\right\|_{2}\right\}\ge \left(p_0+1\right) / 2+\left(p_0+1\right) / 2-1=p_0.$$ This yields that (\ref{indicator}) is further bounded by 
$$n v^\top h\left(\mathcal{W}^*\right)v 	\ge  n \frac{1}{d_1d_2d_3}\|v\|_2^2-n M K \sqrt{1-p_0}\|v\|_2^2,$$
Let $1-p_0$ to be small enough that $M K \sqrt{1-p_0} \leq \frac{1}{2d_1d_2d_3}$. Then we can easily prove that $\lambda_{\min }\left(h\left(\mathcal{W}^*\right)\right)>\frac{1}{2d_1d_2d_3}>0$. Then follow the proof procedure in Step 1, 
\begin{align}	\operatorname{vec}(\Delta)^\top \cdot \widehat{h}\left(\mathcal{W}^*\right) \cdot \operatorname{vec}(\Delta)\ge\frac{C_7}{d_1d_2d_3}\|\Delta\|_F^2.\label{h local}
	\end{align}
On the other hand, we have that
\begin{align}
&\left|\operatorname{vec}(\Delta)^\top \cdot \widehat{h}\left(\mathcal{W}^*\right) \cdot \operatorname{vec}(\Delta)-\operatorname{vec}(\Delta)^\top \cdot \widehat{h}(\mathcal{W}_\eta) \cdot \operatorname{vec}(\Delta)\right| \notag\\
	\leq &\operatorname{vec}(\Delta)^\top \cdot \frac{1}{n} \sum_{t=1}^n\left|b^{\prime \prime}\left(\left\langle\mathcal{W}^*, \mathcal{X}_t\right\rangle\right)-b^{\prime \prime}\left(\left\langle\mathcal{W}_\eta, \mathcal{X}_t\right\rangle\right)\right| 
	\cdot 1_{\left\{\left|\left\langle\mathcal{W}^*, \mathcal { X }_t\right\rangle\right|>\tau\left\|\mathcal{X}_t\right\|_{2}\right\}} \notag\\
 &\cdot \operatorname{vec}\left(\mathcal { X }_t\right) \operatorname{vec}\left(\mathcal{X}_t\right)^\top \cdot \operatorname{vec}(\Delta)\notag \\
	=&\operatorname{vec}(\Delta)^\top \cdot \frac{1}{n} \sum_{t=1}^n\left|b^{\prime \prime \prime}\left(\left\langle\tilde{\mathcal{W}}, \mathcal{X}_t\right\rangle\right)\left\langle\mathcal{W}_\eta-\mathcal{W}^*, \mathcal{X}_t\right\rangle\right| 
	\cdot 1_{\left\{\left|\left\langle\mathcal{W}^*, \mathcal { X }_t\right\rangle\right|>\tau\left\|\mathcal{X}_t\right\|_{2}\right\}} \notag\\&\cdot \operatorname{vec}\left(\mathcal{X}_t\right) \operatorname{vec}\left(\mathcal{X}_t\right)^\top \cdot \operatorname{vec}(\Delta),\notag
\end{align}
where $\tilde{\mathcal{W}}$ is a middle point between $\mathcal{W}^*$ and $\mathcal{W}_\eta$. Hence, $\tilde{\mathcal{W}}$, which is centered at $\mathcal{W}^*$ with radius $l$, is also in the nuclear ball. When the indicator function equals to 1 , we have $\left|\left\langle\tilde{\mathcal{W}}, \mathcal{X}_t\right\rangle\right| \geq$ $\left|\left\langle\mathcal{W}^*, \mathcal{X}_t\right\rangle\right|-\left|\left\langle\mathcal{W}^*-\tilde{\mathcal{W}}, \mathcal{X}_t\right\rangle\right| \geq(\tau-l)\left\|\mathcal{X}_t\right\|_{2}$.
If $(\tau-l)\left\|\mathcal{X}_t\right\|_{2}>1$, according to Assumption 4, we have
\begin{align}
	& \left|b^{\prime \prime \prime}\left(\left\langle\tilde{\mathcal{W}}, \mathcal{X}_t\right\rangle\right)\left\langle\mathcal{W}-\mathcal{W}^*, \mathcal{X}_t\right\rangle\right| \cdot1_{\left\{\left|\left\langle\mathcal{W}^*, \mathcal{X}_t\right\rangle\right|>\tau\left\|\mathcal{X}_t\right\|_{2}\right\}} \notag\\
	\leq  &\frac{\left\|\mathcal{X}_t\right\|_{2}\left\|\mathcal{W}-\mathcal{W}^*\right\|_*}{\left|\left\langle\tilde{\mathcal{W}}, \mathcal{X}_t\right\rangle\right|} 
	\leq  \frac{\left\|\mathcal{X}_t\right\|_{2}\left\|\mathcal{W}-\mathcal{W}^*\right\|_*}{(\tau-l)\left\|\mathcal{X}_t\right\|_{2}} 
	\leq  \frac{l}{\tau-l} .\notag
\end{align}
Otherwise, \begin{small}$\left\|\mathcal{X}_t\right\|_{2}\le1 /(\tau-l)$\end{small} and \begin{small}$\left|b^{\prime \prime \prime}\left(\left\langle\tilde{\mathcal{W}}, \mathcal{X}_t\right\rangle\right)\left\langle\mathcal{W}-\mathcal{W}^*, \mathcal{X}_t\right\rangle\right| \cdot 1_{\left\{\left|\left\langle\mathcal{W}^*, \mathcal { X }_t\right\rangle\right|>\tau\left\|\mathcal{X}_t\right\|_{2}\right\}} $ $\leq C \cdot\left\|\mathcal{W}-\mathcal{W}^*\right\|_*\left\|\mathcal{X}_t\right\|_{2} \leq C \cdot \frac{l}{\tau-l}$\end{small}, where $\mathrm{C}$ is the upper bound of $b^{\prime \prime \prime}(x)$ for $|x| \leq 1$. Thus, let $\mathcal{ X }_t^\prime=\mathcal{U}^\top *_L \mathcal{X}_t *_L \mathcal{V}$ and $\widehat{\Sigma}_{\mathcal{X}^\prime \mathcal{X}^\prime}=\frac{1}{n} \sum_{t=1}^n \operatorname{vec}\left(\mathcal{X}_t^\prime\right) \cdot \operatorname{vec}\left(\mathcal{X}_t^\prime\right)^\top$ and $\Sigma_{\mathcal{X}^\prime \mathcal{X}^\prime}=\mathbb{E} \widehat{\Sigma}_{\mathcal{X}^\prime \mathcal{X}^\prime}$. Denote the upper bound of the eigenvalues of $\Sigma_{\mathcal{X}^\prime \mathcal{X}^\prime}$ as $K_1\ (K_1<\infty)$, we have
\begin{align}
&	\left|\operatorname{vec}(\Delta)^\top \cdot \hat{h}\left(\mathcal{W}^*\right) \cdot \operatorname{vec}(\Delta)-\operatorname{vec}(\Delta)^\top \cdot \hat{h}(\mathcal{W}_\eta) \cdot \operatorname{vec}(\Delta)\right| \notag\\
	 \leq& \operatorname{vec}(\Delta)^\top \cdot \frac{C_8 l}{n(\tau-l)} \sum_{t=1}^n \operatorname{vec}\left(\mathcal { X }_t\right) \operatorname{vec}\left(\mathcal { X }_t\right)^\top \cdot \operatorname{vec}(\Delta)\notag\\
	 =& \frac{C_8 l}{\tau-l} \cdot \frac{1}{n} \sum_{t=1}^n\left\langle\mathcal{X}_t, \widehat{\Delta}\right\rangle^2 
	 = \frac{C_8 l}{\tau-l} \cdot \frac{1}{n} \sum_{t=1}^n\left\langle\mathcal{X}_t, \mathcal{U} *_L \mathcal{S} *_L \mathcal{V}^\top\right\rangle^2\notag\\ 
	 = &\frac{C_8 l}{\tau-l} \cdot \frac{1}{n} \sum_{t=1}^n\left\langle\mathcal{U}^\top *_L \mathcal{X}_t *_L \mathcal{V}, \mathcal{S}\right\rangle^2 
	 = \frac{C_8 l}{\tau-l} \cdot \operatorname{vec}(\mathcal{S})^\top \cdot \widehat{\Sigma}_{\mathcal{X} ^\prime\mathcal{X}^\prime} \cdot \operatorname{vec}(\mathcal{S}) \notag\\
	 = &\frac{C_8 l}{\tau-l} \cdot\left[\operatorname{vec}(\mathcal{S})^\top\cdot \Sigma_{\mathcal{X} ^\prime\mathcal{X}^\prime} \cdot \operatorname{vec}(\mathcal{S})+\operatorname{vec}(\mathcal{S})^\top \cdot\left(\widehat{\Sigma}_{\mathcal{X}^\prime \mathcal{X}^\prime}-\Sigma_{\mathcal{X}^\prime \mathcal{X}^\prime}\right) \cdot \operatorname{vec}(\mathcal{S})\right] \notag\\
	  \leq& 
	  \frac{C_8 l}{\tau-l} \cdot\left(K_1+\frac{32C_4rd_3^{\frac{3}{2}}}{\sqrt{n\min\{d_1,d_2\}^3}}\right)\|\Delta\|_F^2
	  \leq \frac{C_8 l}{\tau-l} \cdot \frac{K_1}{d_1d_2d_3}\|\widehat{\Delta}\|_F^2,\label{h local 2}
\end{align}
where $C_8=\max (C, 1)$. In summary, if the constant $l$ is sufficiently small and satisfies $\frac{C_8 l}{\tau-l} \cdot \frac{K_1}{d_1d_2d_3}<\frac{1}{d_1d_2d_3}$, (\ref{h local}) and (\ref{h local 2}) yield that $
	\operatorname{vec}(\Delta)^\top\cdot\widehat{h}(\mathcal{ W}_\eta)\cdot \operatorname{vec}(\Delta)\ge \frac{C_9}{d_1d_2d_3}
$, and thus $
\operatorname{vec}(\Delta)^\top\cdot\widehat{H}(\mathcal{ W}_\eta)\cdot \operatorname{vec}(\Delta)\ge \frac{C_9}{d_1d_2d_3}
$. This completes the proof of $ L\left(\mathcal{W}^*+\Delta\right)-L\left(\mathcal{W}^*\right)-\left\langle\nabla L\left(\mathcal{W}^*\right), \Delta\right\rangle \ge \frac{C_9}{d_1d_2d_3}$.
\end{proof}
Drawing on Theorem 1 and Lemmas \ref{lemma: lambda} to \ref{lemma: lrsc condition}, Corollary 1 is established. This corollary  shows that the solution $\widehat{\mathcal{W}}$ of (1) is guaranteed to converge to $\mathcal{W}^*$ at a fast rate. Equipped with the convergence guarantee of the estimator $\widehat{\mathcal{W}}$, we further prove the accuracy of the row and column subspace estimators.

\begin{proof}[Proof of Corollary 2.]
	For an orthogonal tensor $\mathcal{Q}$ and arbitrary tensors $\mathcal{D},\ \mathcal{E}$, we have $\|\mathcal{Q} *_L \mathcal{D}\|_F=\frac{1}{\sqrt{\ell}}\|\bar{\mathcal{Q}}\bar{\mathcal{D}}\|_F \le\frac{1}{\sqrt{\ell}}\|\bar{\mathcal{Q}}\|_2\|\bar{\mathcal{D}}\|_F=\frac{1}{\sqrt{\ell}}\|\bar{\mathcal{D}}\|_F=\|\mathcal{D}\|_F$ and $\|\mathcal{D} *_L \mathcal{E}\|_F
	=\frac{1}{\sqrt{\ell}}\|\bar{\mathcal{D}}\bar{\mathcal{E}}\|_F
	\ge\frac{1}{\sqrt{\ell}}\|\bar{\mathcal{D}}\|_F \sigma_{\min}(\bar{\mathcal{E}})=\|\mathcal{D}\|_F \sigma_{\min}(\bar{\mathcal{E}})$, where $\sigma_{\min}(\bar{\mathcal{E}})$ is the smallest singular value of $\bar{\mathcal{E}}$. Therefore, we have
	\begin{align}
		&\left\|\widehat{\mathcal{W}}-\mathcal{W}^*\right\|_F
		\geq \left\|\widehat{\mathcal{U}}_{\perp}^\top*_L\left(\widehat{\mathcal{W}}-\mathcal{U}^**_L\mathcal{U}^{*\top}*_L\mathcal{W}^*\right)\right\|_F\notag\\
		=&\left\|\widehat{\mathcal{U}}_{\perp}^\top *_L\mathcal{U}^**_L\mathcal{U}^{*\top}\mathcal{W}^*\right\|_F
		\geq \left\|\widehat{\mathcal{U}}_{\perp}^\top*_L \mathcal{U}^*\right\|_F \omega_{min},\label{inequality:U}
	\end{align}
	where $\omega_{min}$ is the smallest singular value of $\bar{\mathcal{W}}^*$. Similarly, the following inequality holds:
	\begin{align}
		\left\|\widehat{\mathcal{W}}-\mathcal{W}^*\right\|_F
		\geq \left\|\widehat{\mathcal{V}}_{\perp}^\top *_L\mathcal{V}^*\right\|_F \omega_{min}.\label{inequality:V}
	\end{align}
	Combining (\ref{inequality:U}), (\ref{inequality:V}) with Corollary 1 gets the desired result.
\end{proof}

This concludes the proof of $B_{\perp}$. Lastly, we establish the proof for the total regret.

\begin{proof}[Proof of Theorem 2.]
Define $\mathcal{X}^*=\arg\max_{\mathcal{X}\in \mathbb{X}}\langle \mathcal{X},\mathcal{W}^*\rangle$, then according to the assumption $
	\left\|\mathcal{W}^*\right\|_F\le1 , \left\|\mathcal{X}\right\|_F\le1
	$, and the link function $\mu(\cdot)$ is $M$-Lipshitz (according to the Assumption 4), we have
	\begin{align}
		&R_{T_1}=\sum_{t=1}^{T_1}(\mu\langle \mathcal{X}^*,\mathcal{W}^*\rangle-\mu\langle \mathcal{X}_t,\mathcal{W}^*\rangle)\notag\\
		\le&\sum_{t=1}^{T_1}M \langle \mathcal{X}^*-\mathcal{X}_t,\mathcal{W}^*\rangle
		\le T_1 M \cdot \left\| \mathcal{X}^*-\mathcal{X}_t\right\|_F \left\|\mathcal{W}^*\right\|_F
		\le 2T_1M.\label{bound stage1}
	\end{align}
	And Lemma \ref{lemma: 18} gives the regret for LowGLM-UCB, it remains to put  parameters into the bound.
	\begin{align}
		R_{T-T_1}
		=\widetilde{O}\left(k\sqrt{T}+B_{\perp} T\right)
		=\widetilde{O}\left(k\sqrt{T}+ \frac{(d_1+d_2)^3 \ell d_3\log d_3}{aT_1(1-\gamma)^2 \omega_{min}^2} T\right).
		\label{bound stage2}
	\end{align}
	The addition of bounds (\ref{bound stage1}) and (\ref{bound stage2}) results that
	\begin{align}
		R_{T}=R_{T_1}+R_{T-T_1}
		=\widetilde{O}\left(T_1+\frac{(d_1+d_2)^3\ell d_3\log d_3}{aT_1(1-\gamma)^2\omega_{min}^2} T\right).
		\notag
	\end{align} 
	Then by optimizing $T_1$, we can further upper bound the total regret by
	\begin{align}
		R_{T}
		=\widetilde{O}\left(T_1+\frac{(d_1+d_2)^3\ell d_3\log d_3}{aT_1(1-\gamma)^2\omega_{min}^2} T\right)
		=\widetilde{O}\left(\frac{(d_1+d_2)^{\frac{3}{2}}\sqrt{d_3\ell T}}{\sqrt{a}(1-\gamma)\omega_{min}} \right),
		\notag
	\end{align} 
	when $T_1=\frac{(d_1+d_2)^{\frac{3}{2}}\sqrt{d_3\ell\log d_3T}}{\omega_{min}\sqrt{a}(1-\gamma)}$. This completes the proof.
\end{proof}

\end{document}